\documentclass{article} 
\usepackage{style_file,times}

\usepackage{amsmath,amsfonts,bm}

\def\eqref#1{equation~\ref{#1}}

\def\1{\bm{1}}

\DeclareMathAlphabet{\mathsfit}{\encodingdefault}{\sfdefault}{m}{sl}
\SetMathAlphabet{\mathsfit}{bold}{\encodingdefault}{\sfdefault}{bx}{n}

\newcommand{\E}{\mathbb{E}}

\newcommand{\KL}{D_{\mathrm{KL}}}

\usepackage{hyperref}
\usepackage{url}
\usepackage{microtype}
\usepackage{graphicx}
\usepackage{subcaption}
\usepackage{tikz}
\usepackage{amsmath,amssymb,amsthm}
\usepackage{mathtools}
\usepackage{soul}
\usepackage{enumitem}
\usepackage{algorithm}
\usepackage{amsfonts}
\usepackage[noend]{algpseudocode}
\usepackage{booktabs}
\usepackage{multirow}
\usepackage{makecell}
\usepackage{xcolor,pifont}
\usepackage{bm}
\usepackage{authblk}
\newcommand*\colourcheck[1]{%
  \expandafter\newcommand\csname #1check\endcsname{\textcolor{#1}{\ding{52}}}%
}
\colourcheck{blue}
\colourcheck{green}
\colourcheck{red}
\newcommand{\xmark}{\ding{55}}
\usepackage{fontawesome} 

\hypersetup{
    colorlinks=false,
    linkcolor=black,
    urlcolor=black
}

\newtheorem{theorem}{Theorem}[section]

\newtheorem{lemma}[theorem]{Lemma}
\newtheorem{remark}[theorem]{Remark}
\newtheorem{corollary}[theorem]{Corollary}
\newtheorem{assumption}{A\hspace{-2pt}}
\usepackage{scalerel,stackengine}
\stackMath
\newcommand\reallywidehat[1]{%
\savestack{\tmpbox}{\stretchto{%
  \scaleto{%
    \scalerel*[\widthof{\ensuremath{#1}}]{\kern-.6pt\bigwedge\kern-.6pt}%
    {\rule[-\textheight/2]{1ex}{\textheight}}
  }{\textheight}%
}{0.5ex}}%
\stackon[1pt]{#1}{\tmpbox}%
}

\title{\centering \textbf{Certified Self-Consistency:}\\ Statistical Guarantees and Test-Time Training for Reliable Reasoning in LLMs}

\author[ ]{P. Cordero-Encinar}
\author[ ]{A. B. Duncan}
\affil[ ]{Department of Mathematics, Imperial College London, UK.}
\affil[ ]{\textit{\{paula.cordero-encinar22, a.duncan\}@imperial.ac.uk}}

\iclrfinalcopy 
\begin{document}

\maketitle
\vspace{-30pt}
\begin{center}
\faBook\ \href{https://paulaoak.github.io/certified_self_consistency_website/}{\textbf{Website}} \quad
\faGithub\ \href{https://github.com/paulaoak/certified_self_consistency}{\textbf{Code}}
\end{center}

\vspace{15pt}
\begin{abstract}
Recent advances such as self-consistency and test-time reinforcement learning (TTRL) improve the reliability of large language models (LLMs) without additional supervision, yet their underlying mechanisms and statistical guarantees remain poorly understood.
We present a unified framework for certifiable inference in LLMs, showing that majority voting provides a statistical certificate of self-consistency: under mild assumptions, the aggregated answer coincides with the mode of the model’s terminal distribution with high probability. We derive finite-sample and anytime-valid concentration bounds that quantify this confidence, and introduce the Martingale Majority Certificate (MMC), a sequential stopping rule that adaptively determines when sufficient samples have been drawn.
We further prove that label-free post-training methods such as TTRL implicitly sharpen the answer distribution by exponentially tilting it toward its mode, thereby reducing the number of samples required for certification. Building on this insight, we propose new post-training objectives that explicitly optimise this trade-off between sharpness and bias.  Together, these results explain and connect two central test-time scaling strategies, self-consistency and TTRL,  within a single statistical framework for label-free, certifiable reliability in reasoning LLMs.
\end{abstract}

\section{Introduction}\label{sec:introduction}
Large language models (LLMs) have demonstrated striking performance across a range of reasoning tasks, from mathematical problem solving to code generation \citep{brown2020gpt3,openai2023gpt4}. 
A key advance has been \emph{chain-of-thought} (CoT) prompting, which encourages the model to produce explicit intermediate thinking steps before returning a final answer \citep{wei2022cot,kojima2022zeroshot}. 
CoT substantially improves accuracy on problem-solving benchmarks \citep{lewkowycz2022minerva}. 
The quality of LLM outputs is influenced by the underpinning {decoding strategy} adopted at inference time.  Deterministic decoding (e.g.\ greedy or low-temperature sampling), which selects the most probable token at each step,  yields a single trajectory but limits exploration, often causing the model to commit early to a rollout, potentially leading to an incorrect reasoning path in the context of CoT.   On the other hand, stochastic decoding methods such as nucleus or temperature sampling encourage diversity over possible rollouts, revealing alternative chains of thought that may reach the correct solution.  While greedy decoding has been shown to outperform sampling  when comparing single rollouts across various benchmarks \citep{song2024good}, combining multiple samples can yield strong performance.   This is exploited in \emph{test-time scaling} strategies, which seek to improve the reliability and accuracy of model responses at inference time by exploring and aggregating information through ranking or aggregation.   While this requires more compute at test time, such approaches demonstrably improve performance without the need for retraining, particularly for small-footprint models \citep{chan2025lean}.   In the context of LLMs, a wide range of test-time scaling approaches have emerged.   If an external verifier is available (e.g. a proof checker or an external model), various strategies are available, including ranking through Best-of-N \citep{cobbe2021gsm8k, lightman2023let}.
 In the absence of an external verifier, one can resort to aggregation approaches such as self-consistency / majority-voting \citep{wang2022selfconsistency}; trajectory extension approaches which encourage more complete reasoning, such as budget forcing \citep{muennighoffs1} or multi-hop reasoning; or self-evaluation strategies \citep{Kadavath2022LanguageM} where the model explores multiple reasoning branches through search and self-evaluates their quality, such as Tree of Thoughts \citep{yao2023treeofthoughts}, Beam Search \citep{xie2023selfevaluation} and Monte-Carlo Tree Search \citep{coulom_mcts_2007, xie2024monte}.
\\\\
We formalise a LLM rollout as a stochastic decoding process
\[
(Y_t)_{t \ge 0}, \quad Y_t \in \mathcal{V},
\]
where $\mathcal{V}$ is the vocabulary and the process is initialised by a prompt $pr$. 
At each step the model samples
\[
Y_t \sim \pi_\phi(\cdot \mid Y_{<t}, pr),
\]
from a conditional policy parametrised by weights~$\phi$. 
The \emph{thinking phase} consists of the random evolution of this sequence until a termination token is produced, at which point the model emits the response, starting from a random stopping time~$\tau$. 
We denote by
\[
X := g(Y_{\tau:}) \in \mathcal{A}
\]
the canonicalised terminal answer, obtained by applying a deterministic extraction map $g$. 
The induced terminal distribution $\mathbf p = \mathrm{Law}(X)$ over the answer set $\mathcal{A}$ captures the model’s epistemic uncertainty about its own final output. 
In an ideal reasoning model, we would like rollouts to exhibit rich variability in $Y_{1:\tau-1}$ (the reasoning trajectories), yet concentrate mass in the final answer $X$ (the outcome). That is, we seek {diversity over reasoning paths, but consistency over terminal responses}.
\\\\
In the absence of external reward signals, a model must act relative to its own uncertainty. 
Letting $a \in \mathcal{A}$ denote the chosen output and $X \sim \mathbf p$ the stochastic model response, the expected $0$--$1$ loss is $\mathbb{E}[1\{a \neq X\}]$. 
The Bayes-optimal decision minimising this loss is the mode
\[
c^\star = \arg\max_j p_j,
\]
which corresponds to the model’s most probable self-consistent answer. 
Hence, under symmetric loss, recovering the mode is the optimal \emph{model-relative} prediction. 
When a verifier is absent, certifying that a model’s reported answer coincides with this mode provides a natural measure of reliability.

\paragraph{Statistical certificates of self-consistency.}
In practice, the terminal probabilities $\mathbf p$ are unknown and can be estimated only through multiple independent rollouts $X_1,\ldots,X_n$. 
The simplest estimator of the mode is the \emph{majority vote}
\[
\widehat{c}_n := \arg\max_j \hat p_{n,j}, 
\qquad
\hat p_{n,j} = \frac{1}{n}\sum_{i=1}^{n}\mathbf{1}\{X_i=j\}.
\]
This estimator forms the basis of \emph{self-consistency} test-time scaling \citep{wang2022selfconsistency}, which has been shown to stabilise CoT reasoning and improve benchmark accuracy \citep{anil2023palm}.  
From a statistical standpoint, majority voting is the Bayes-optimal estimator of $c^\star$ under 0--1 loss, and an associated upper bound on $\mathbb{P}[\widehat{c}_n \neq c^\star]$ provides a \emph{statistical certificate of self-consistency}: a quantitative guarantee that the aggregated answer coincides with the mode of the terminal law~$\mathbf p$ with high probability. 
Under standard regularity conditions (e.g.\ conditional independence of rollouts and a unique mode of $\mathbf p$), the majority-vote estimator is consistent, satisfying $\Pr[\widehat{c}_n = c^\star] \to 1$ as $n \to \infty$.  
A more practical question concerns the finite-sample regime: how large must $n$ be to guarantee, with confidence $1-\varepsilon$, that $\widehat{c}_n$ already equals $c^\star$?

To address this, we derive a hierarchy of statistical certificates, valid in the finite-sample and asymptotic regimes, leveraging Hoeffding, Bernstein, Chernoff--Markov, and large-deviation concentration bounds for the error probability $\mathbb{P}[\widehat{c}_n \neq c^\star]$. 
Although not tight in the small-sample regime, these bounds clarify how reliability scales with the ensemble size and with the \emph{mode margin} $\delta = p_{c^\star} - p_{j^\star}$, i.e.\ the gap between the top two answer probabilities.
\\\\
If the probabilities $p_j$ were known, one could invert these bounds to determine the number of samples required to achieve a desired confidence $1-\varepsilon$. 
In reality, both $p_j$ and $\delta$ must be estimated on the fly. 
This motivates a \emph{sequential} formulation: as rollouts arrive, can we determine adaptively when the current majority is statistically reliable? 
We introduce the \emph{Martingale Majority Certificate (MMC)}, a sequential procedure based on $e$-values and Ville’s inequality \citep{ville1939collectif,howard2021confidenceseq}, which adaptively tests whether the empirical leader remains significantly ahead of its nearest rival and of all others combined. This guarantees that at the (random) stopping time~$\tau$,
\[
\Pr[\widehat{c}_{n_\tau} \neq c^\star] \le \varepsilon,
\]
thus providing an \emph{anytime-valid certificate} of model self-consistency.

\paragraph{Why test-time training helps.}
Recent work on label-free post-training, such as \emph{test-time reinforcement learning} (TTRL), adapts model parameters online by optimising KL-regularised objectives with respect to its own rollouts \citep{zuo2025ttrl,akyurek2025ttt}. 
These methods empirically improve reliability but their mechanism remains opaque. 
We show that such objectives correspond to an \emph{exponential tilting} of the terminal law $\mathbf p$, yielding a sharpened distribution more concentrated around its mode. 
This transformation increases the mode margin, improving the signal-to-noise ratio of the margin random variable 
\(
\Delta_{j^\star} = \mathbf{1}\{X=c^\star\}-\mathbf{1}\{X=j^\star\},
\)
and thereby reducing the number of samples required for certification.
However, it also introduces a controlled bias relative to the original distribution, governed by the KL regularisation strength. 
Thus, TTRL provides a complementary lever: by reshaping $\mathbf p$ to enlarge $\delta$, it lowers the compute required for reliable self-consistency.  In the presence of a verifier, the deep connection between tilting and Best-of-N scaling for alignment is well understood \citep{beiramitheoretical,yang2024asymptotics,gui2024bonbon} as well as the general uses of tilting in machine learning more broadly \citep{li2023tilted}.   In this work we explore a similar phenomenon that arises purely based on consensus and in the absence of a reward signal.   Related to our approach, recent work has exploited this connection to tilting by introducing an MCMC scheme at inference-time to directly sample this tilted distribution  to boost model capability \citep{karan2025reasoning}.

\paragraph{Emergent calibration in reasoning models.}
Beyond the theoretical and algorithmic results, our experiments reveal a notable empirical regularity: the
\emph{signal-to-noise ratio} (SNR) of the margin variable 
\(\Delta_{j^\star} = \mathbf 1\{X = c^\star\} - \mathbf 1\{X = j^\star\}\),
which quantifies the sharpness of the model’s terminal answer distribution,
correlates strongly with external measures of problem difficulty (Figure~\ref{fig:QWEN-MATH-1.5B-SNR-0.1}).
Across the MATH-500 benchmark, harder problems exhibit systematically lower and more variable SNR values,
while easier problems yield sharply peaked distributions concentrated around a single answer.
\\\\
This behaviour is non-trivial: the model has no access to ground-truth difficulty labels, yet its own 
uncertainty, reflected in the variability of its rollouts, aligns closely with these labels.
This suggests an \emph{emergent form of calibration} in reasoning LLMs:
without explicit supervision or external verification, models appear to ``know when they do not know.''
In statistical terms, the SNR acts as a label-free proxy for epistemic uncertainty and, consequently, for task difficulty.
\\\\
This observation links our theoretical framework to observable model behaviour.
The same margin variable that governs finite-sample concentration and sequential certification (Sections~\ref{sec:theoretical_bounds}--\ref{sec:stopping_rule})
also provides a practical signal for compute-adaptive inference:
when the SNR is low, additional rollouts or verifier checks can be triggered,
whereas high-SNR cases can be certified with fewer samples.
Hence, the SNR not only underpins the theory of certified self-consistency,
but also yields a measurable and actionable indicator of reliability in reasoning models.

\paragraph{Our contributions.}
We develop a framework for \emph{certifiable inference in chain-of-thought LLMs}, viewing majority voting as a statistical certificate for the terminal law $\mathbf p=\mathrm{Law}(X)$. 
Specifically:
\begin{enumerate}
\item \textbf{Finite-sample and asymptotic certificates.}
We derive explicit Hoeffding, Bernstein, Chernoff--Markov, and large-deviation concentration bounds for $\mathbb{P}[\widehat{c}_n \neq c^\star]$, characterising how reliability improves with ensemble size as a function of the mode margin~$\delta$.
\item \textbf{Anytime-valid stopping certificates.}
We propose the \emph{Martingale Majority Certificate (MMC)}, a sequential test that adaptively determines when sufficient rollouts have been drawn, guaranteeing $\Pr[\widehat{c}_n \neq c^\star]\le \varepsilon$ at stopping.
\item \textbf{Explaining test-time reinforcement learning.}
We formalise the connection between KL-regularised TTRL objectives and exponential tilting of $\mathbf p$, explaining why these methods improve reliability by increasing the mode margin and thereby reducing the sample complexity for certification. 
Building on this insight, we introduce alternative post-training objectives optimising this trade-off between sharpness and bias.
\item \textbf{Empirical link between uncertainty and problem difficulty.}
We show that the signal-to-noise ratio (SNR) of the margin variable $\Delta_{j^\star}$, which governs our statistical  certificates, correlates strongly with externally defined difficulty levels,
revealing an emergent form of calibration in reasoning LLMs.
\end{enumerate}

\begin{figure}[t]
    \centering
    \includegraphics[width=\linewidth]{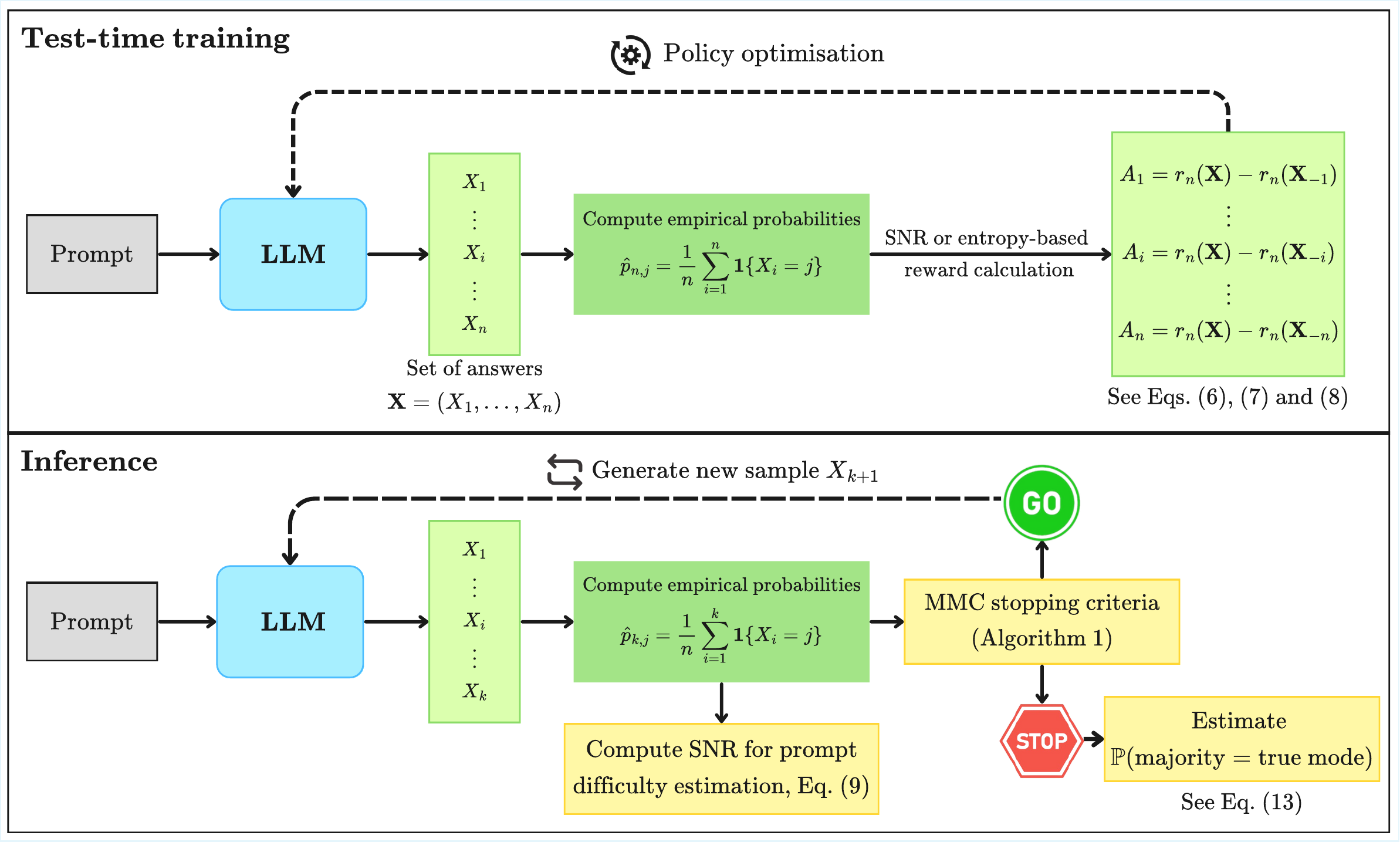}
    \caption{
    \textbf{Overview of the proposed framework.}
    Given a prompt, the model generates multiple reasoning rollouts from the
    reference distribution $\pi_{\mathrm{ref}}(\cdot|{pr})$.
    The resulting terminal answers are aggregated via majority voting, viewed
    as mode estimation under sampling uncertainty.
    The Martingale Majority Certificate (MMC) monitors the empirical margin and
    provides an \emph{anytime-valid} stopping rule for certification.
    Test-time training with SNR or entropy-based adaptation sharpens the
    terminal distribution, thereby increasing the
    signal-to-noise ratio (SNR) and reducing the number of samples required for
    certification. 
    }
    \label{fig:framework}
\end{figure}

Together, these results provide a principled strategy for \emph{certifying that an LLM’s output coincides with its own most probable prediction} through self-consistency.  By linking concentration bounds, martingale stopping rules, and test-time reinforcement learning, we provide a unified statistical framework of when and why self-consistency is reliable for reasoning models, and how test-time adaptation can further reduce the computational cost of this certification. Figure \ref{fig:framework} summarises the components of our framework.

\section{Statistical guarantees for majority voting}\label{sec:theoretical_bounds}

It is well known that majority voting is \emph{consistent}: under i.i.d.\ rollouts and a unique mode
$p_{c^\star}>\max_{j\neq c^\star}p_j$, we have that $\widehat{c}_n\to c^\star$ a.s.\ as $n\to\infty$.  This is a direct extension of Condorcet's original jury theory \citep{condorcet1785essai} to the multi-class setting \citep{list2001epistemic}.    Our goal in this section is to quantify the error, $\mathbb{P}[\widehat{c}_n \neq c^\star]$, i.e. when the majority vote over $n$ i.i.d.\ rollouts
$X_1,\dots,X_n \sim \mathrm{Cat}(\mathbf p)$ fails to return the true mode
$c^\star = \arg\max_j p_j$, and how this error scales with the ensemble size $n$.

\paragraph{Setting and scope.}
We analyse an oracle setting where the terminal answer distribution $\mathbf p=(p_1,\dots,p_k)$ is {known}. 
This isolates what drives certainty under majority aggregation, specifically, how the error $\Pr[\widehat{c}_n\neq c^\star]$ scales with the mode margin $\delta=p_{c^\star}-p_{j^\star}$, the variances $\sigma_j^2$ of the margin random variables $\Delta_j ={ \mathbf 1\{X=\hat c\}-\mathbf 1\{X=j\}}$, and the signal-to-noise ratio of $\Delta_{j^\star}$. 
The resulting finite-sample bounds and asymptotic rates provide {insight} into the determinants of reliability, and form the basis of an operational certificate for inference in Section \ref{sec:stopping_rule}, where we demonstrate how $\mathbf{p}$ can be simultaneously inferred from rollouts.  Throughout we assume i.i.d.\ rollouts (conditional on the prompt) and a unique mode $p_{c^\star}>\max_{j\neq c^\star}p_j$; violations (e.g., strong correlations or ties) weaken guarantees and are handled adaptively by MMC.

Figure~\ref{fig:comparison_bounds_majority_vote} compares the main bounds below with empirical estimates; full proofs are deferred to Appendix \ref{app:proofs_for_theoretical_bounds}.

\subsection{Exact error probability with oracle $\mathbf p$}
\label{subsec:small_sample}

When $\mathbf p$ is known, the error probability admits an exact multinomial expression.

\begin{theorem}[Exact small-sample probability]\label{thm:bounds_small_regime}
For all $n\ge 1$,
\[
\Pr[\widehat{c}_n\neq c^\star]
\;=\;
\sum_{\substack{x\in\mathbb N^k\\ x_1+\cdots+x_k=n\\ x_{c^\star}\le \max_{j\neq c^\star}x_j}}
\frac{n!}{x_1!\cdots x_k!}\,p_1^{x_1}\cdots p_k^{x_k}.
\]
\end{theorem}

This formula provides the ground truth for the oracle setting and is particularly useful for validating bounds.   For small ensembles ($n\lesssim 50$), it is possible to compute this effectively via a dynamic-programming scheme (see Appendix~\ref{app:small_regime_bounds}), but quickly becomes intractable for increasing $n$.    Theorem \ref{thm:bounds_small_regime} is not illuminating about the {drivers} of certainty.   To see these more clearly, we leverage concentration bounds which provide exponentially decaying finite-sample bounds.

\subsection{Finite-sample certificates}
\label{subsec:finite_sample}

Under a unique mode and conditional independence of rollouts, majority voting admits exponentially decaying error bounds which are valid for any finite number of samples. We collect the main instances into a single statement.

\begin{theorem}[Finite-sample certificate]\label{thm:finite_sample_unified}
Assume $p_{c^\star}>\max_{j\neq c^\star}p_j$. Then for all $n\ge 1$,
\begin{align*}
\Pr[\widehat{c}_n\neq c^\star]
\;\le\;
\sum_{j\neq c^\star}
\min\Bigg\{
&\underbrace{\exp\!\Big(-\tfrac{n}{2}(p_{c^\star}-p_j)^2\Big)}_{\text{\emph{Hoeffding}}},\quad
\underbrace{\exp\!\Big(-\,\tfrac{n (p_{c^\star}-p_j)^2}{\,2\sigma_j^2+\frac{2}{3}(p_{c^\star}-p_j)+\frac{2}{3}(p_{c^\star}-p_j)^2}\Big)}_{\text{\emph{Bernstein}}},\\
&\underbrace{\exp\!\Big(n\log\big(1-(\sqrt{p_{c^\star}}-\sqrt{p_j})^2\big)\Big)}_{\text{\emph{Chernoff--Markov}}}
\Bigg\}.
\end{align*}
\end{theorem}

Introducing the probability gap $\delta^2 = \min_{j \neq c^\star}(p_{c^\star} - p_j)^2$, Hoeffding's inequality implies that
\small
$$
\mathbb{P}[\widehat{c}_n \neq c^\star] \leq (k-1)e^{-n\delta^2/2}.
$$
\normalsize
From this we obtain that
$
n \geq -\tfrac{2}{\delta^2}\log\left(\tfrac{\varepsilon}{k-1}\right)
$
samples are sufficient to guarantee that the majority vote is correct with probability at least $1-\varepsilon$.

\noindent
\textit{Interpretation.} We observe that the probability gaps $p_{c^\star} - p_j$ play a major role in these bounds. 
While Hoeffding's rate depends only on the gap,  Bernstein tightens the rate when variances are smaller, offering an advantage when few rivals have non-negligible mass.  
These bounds can be further tightened through the introduction of additional prefactors \citep{bahadur-rao}. A full statement with explicit constants and proofs can be found in Appendices~\ref{app:hoeffding_bound}-\ref{app:chernoff-markov_bound}.   A weighted-majority extension (heterogeneous experts) of Hoeffding's bound is deferred to Appendix~\ref{app:weighted_majority_vote}.

\subsection{Asymptotic consistency and the governing rate}
\label{subsec:asymptotics}

As $n$ grows, the above finite sample bounds yield \emph{exponential} improvement in reliability.   In the asymptotic regime ($n\rightarrow \infty$) we are able to leverage additional strategies which yield different perspectives on the driving factors.  There are two complementary asymptotic lenses:

\emph{(i) Gaussian/CLT regime.}
Viewing the multinomial counts through a multivariate central limit theorem (CLT) yields normal tail
approximations for the pairwise margins $N_{c^\star}-N_j$. These can be further refined through Berry–Esseen corrections, which provide $O(n^{-1/2})$ refinements. 

\emph{(ii) Large-deviations (Sanov/Cramér) regime.}
A large-deviation analysis \citep{dembo2010ldp} characterises the exact first-order exponent:
$\Pr[\widehat{c}_n\neq c^\star]=\exp(-n I^\star(\mathbf p)+o(n))$, where $I^\star(\mathbf p)$ is the
minimal KL divergence to a distribution in which a rival ties the leader. Bahadur–Rao–type refinements provide $\Theta(n^{-1/2})$ prefactors to further tighten these approximations.

The two views agree to second order: for small margins $\delta=p_{c^\star}-p_{j^\star}\ll p_{j^\star}$,
the large-deviation exponent expands as $I^\star(\mathbf p)=\delta^2/(2\sigma_{j^\star}^2)+O(\delta^3)$,
matching the CLT rate (\ref{eq:clt-rate}). Practically, the CLT bound gives a transparent dependence on SNR and is
useful for interpretable sample-complexity proxies, while the Sanov rate is preferable when a sharp
exponent is needed or when inverting for $n$.

The results are detailed in the following theorem, which summarises both the CLT and large-deviations regimes.

\begin{theorem}[Asymptotic consistency]\label{thm:majority_rates}
Assume $p_{c^\star}>\max_{j\neq c^\star}p_j$. Then, as $n\to\infty$,
\begin{align}
\Pr[\widehat{c}_n=c^\star]
&= 1 - \sum_{j\neq c^\star}\Phi\!\Big(-\,\tfrac{(p_{c^\star}-p_j)\sqrt n}{\sigma_j}\Big)\,[1+O(n^{-1/2})]
\nonumber\\[-1mm]
&\ge 1 - \frac{k-1}{2}\exp\!\Big\{-\frac{n}{2}\min_{j\neq c^\star}\Big(\tfrac{p_{c^\star}-p_j}{\sigma_j}\Big)^2\Big\},\label{eq:clt-rate}
\end{align}
where $\Phi$ is the standard normal CDF and $\sigma_j^2=p_{c^\star}+p_j-(p_{c^\star}-p_j)^2$.
Moreover,
\begin{gather*}
\Pr[\widehat{c}_n\neq c^\star] \;=\; \exp\big(-n\,I^\star(\mathbf p)+o(n)\big),\\
\qquad
I^\star(\mathbf p)
= \min_{j\neq c^\star}\inf_{\mathbf q:\,q_{c^\star}=q_j}D_{\mathrm{KL}}(\mathbf q\Vert \mathbf p)
= -\log\!\Big(1-(\sqrt{p_{c^\star}}-\sqrt{p_{j^\star}})^2\Big).
\end{gather*}
\end{theorem}

Motivated by the Gaussian bound we define the \emph{signal-to-noise ratio} (SNR) by
\begin{equation*}
\mbox{SNR}(\Delta_{j^\star}) = \frac{\delta^2}{\,2p_{c^\star}-\delta-\delta^2\,}
\end{equation*}
where $\delta = p_{c^\star} - p_{j^\star}$.  
\noindent
The Gaussian bound reveals that the decay rate is governed by the \emph{worst} signal-to-noise ratio of the margin variables:
\begin{equation}\label{eq:SNR_winner_runner_up}
\min_{j\neq c^\star}\Big(\tfrac{p_{c^\star}-p_j}{\sigma_j}\Big)^2
\;=\;
\Big(\tfrac{p_{c^\star}-p_{j^\star}}{\sigma_{j^\star}}\Big)^2
\;=\;
\mathrm{SNR}(\Delta_{j^\star}).
\end{equation}
We also note that the rate function $I^\star(\mathbf{p})$ recovers the same rate as the Chernoff-Markov bound in Theorem~\ref{thm:finite_sample_unified}.  For small margins $\delta\ll p_{j^\star}$, the large-deviation exponent admits the expansion
$$I^\star(\mathbf p)=\delta^2/\big(2\sigma_{j^\star}^2\big)+O(\delta^3),$$
consistent with the Gaussian rate. Proofs are given in Appendices~\ref{app:majority-clt} and \ref{app:sanov_bound}.

\medskip

From these results, we see that majority voting acts as a statistical amplifier: under a unique mode and conditionally-independent rollouts,
the error probability decays {exponentially} in $n$.
The governing rate is the SNR of the margin $\Delta_{j^\star}$ in (\ref{eq:SNR_winner_runner_up}).
This same quantity controls the Martingale Majority Certificate (Section~\ref{sec:stopping_rule}) and motivates
test-time training objectives that enlarge the mode margin and improve sample efficiency (Section~\ref{sec:test_time_training_loss}).

\section{Martingale Majority Certificate: a practical stopping rule}\label{sec:stopping_rule} 
In this section we introduce the \emph{Martingale Majority Certificate} (MMC), a principled stopping rule that adaptively decides when to stop sampling rollouts while controlling the error of returning the empirical majority. Rather than fixing $n$ in advance, MMC updates after each
new sample and stops once the empirical evidence is sufficient.

We consider the following setting: 
at step $n$, we have samples $X_1, \dots, X_n \sim X$  from the terminal distribution over $\{1, \dots, k\}$, generated from $n$ independent rollouts, where $k$ is possibly unknown.  These are independent and identically distributed, conditioned on the prompt $pr$.
The true but unknown class probabilities are $p_j = \mathbb{P}[X=j\,|\,pr]$, and the empirical frequencies are
$\hat{p}_{n,j}$. 

Our goal is to construct a stopping rule that guarantees, with high confidence, that the majority vote $\widehat{c}_n = \arg\max_j\, \hat{p}_{n,j} $ coincides with the true mode $c^\star = \arg\max_j\, {p}_{j}$. Formally, we seek a strategy such that, at the stopping iteration $n_\tau$, 
$
\mathbb{P}\left[\widehat{c}_{n_\tau}\neq c^\star\right]\leq \varepsilon.
$
 The central challenge in the LLM setting is the potentially large number of possible outcomes. A naive stopping rule would require pairwise comparisons of the empirical probabilities across all classes $i \neq j$, $i,j \in\{1, \dots, k\}$, which becomes computationally prohibitive as $k$ grows.   

To address this, we exploit the observation that the mass of the terminal law is typically concentrated on a few classes $m\ll k$.  Thus, instead of considering all classes individually, we aggregate votes into three categories: $(i)$ the current leader $\widehat{c}_n$, $(ii)$ the top-$(m-1)$ runner-ups, $j_{n,1}^\star, \dots, j_{n,m-1}^\star$, where $j_{n,i}^\star = \arg\max_{j\neq \widehat{c}_n, j_{n,1}^\star,\dots, j_{n,i-1}^\star} \,\hat{p}_{n,j}$, and $(iii)$ all the \emph{others}.
Note that
\small
\begin{align*}
\widehat{c}_n = c^\star 
&\iff \big(\forall \ i\ \in\{1,\dots, m-1\};\ \ p_{\widehat{c}_n}> p_{j_{n,i}^\star}\big)\ \text{AND}\ \big(\forall \ j \in \text{\{\emph{others}\}};\ \  p_{\widehat{c}_n}> p_j\big)\\
&\,\,\Longleftarrow \,\,\,\big(\forall \ i\ \in\{1,\dots, m-1\};\ \ p_{\widehat{c}_n}> p_{j_{n,i}^\star}\big)\ \text{AND}\ \big(  p_{\widehat{c}_n}> {\scriptstyle\sum}_{j\in\, \textit{others}}\, \, p_j\big).
\end{align*}
\normalsize
Accordingly, we perform two tests: leader vs top-$(m-1)$ runner-ups, and leader vs \emph{others}.
We stop only when both conditions are satisfied with high probability, ensuring that $\widehat{c}_n$ coincides with the true mode with high confidence. 
In what follows, we focus on the case $m=2$, a detailed construction of the stopping rule for general $m$ is provided in Appendix \ref{app:subsec_general_stopping_rule}.

\subsection{Anytime-valid $e$-processes}
\label{subsec:mmc_recursive}

At round $n\!\ge\!1$, \emph{before} observing $X_n$, set the predictable top-2 labels as
\[
A_{n-1}:=\widehat c_{\,n-1},\qquad B_{n-1}:=j^\star_{\,n-1},
\]
which are measurable w.r.t.\ $\mathcal F_{n-1}=\sigma(X_1,\dots,X_{n-1})$ (ties broken deterministically).
We maintain the following \emph{recursive, predictable} counts
\[
\begin{aligned}
&\textbf{Leader hits:}&& s_n \;=\; s_{n-1} + \mathbf 1\{X_n = A_{n-1}\}, \quad s_0=0,\\
&\textbf{Runner-up hits (for the A vs B test):}&& f_n \;=\; f_{n-1} + \mathbf 1\{X_n = B_{n-1}\}, \quad f_0=0,\\
&\textbf{Others hits (for the A vs others test):}&& o_n \;=\; o_{n-1} + \mathbf 1\{X_n \notin \{A_{n-1},B_{n-1}\}\}, \quad o_0=0.
\end{aligned}
\]
Thus the sample sizes are
\[
M_n := s_n + f_n,\qquad
T_n := s_n + o_n.
\]
Let $(\pi^{\mathrm{run}}_{n})_{n\ge1}$ and $(\pi^{\mathrm{oth}}_{n})_{n\ge1}$ be  {predictable} priors (each $\pi_n$ is $\mathcal F_{n-1}$-measurable) 
supported on $(1/2,1]$.
Define the two mixture $e$-processes recursively (with optional skipping) by
\begin{align*}
e^{\mathrm{run}}_n
&=\begin{cases}
e^{\mathrm{run}}_{n-1}\cdot 2\!\displaystyle\int \theta\,\pi^{\mathrm{run}}_n(d\theta), & X_n = A_{n-1},\\[1mm]
e^{\mathrm{run}}_{n-1}\cdot 2\!\displaystyle\int (1-\theta)\,\pi^{\mathrm{run}}_n(d\theta), & X_n = B_{n-1},\\[1mm]
e^{\mathrm{run}}_{n-1}, & \text{otherwise,}
\end{cases}\\[2mm]
e^{\mathrm{oth}}_n
&=\begin{cases}
e^{\mathrm{oth}}_{n-1}\cdot 2\!\displaystyle\int \lambda\,\pi^{\mathrm{oth}}_n(d\lambda), & X_n = A_{n-1},\\[1mm]
e^{\mathrm{oth}}_{n-1}\cdot 2\!\displaystyle\int (1-\lambda)\,\pi^{\mathrm{oth}}_n(d\lambda), & X_n \notin \{A_{n-1},B_{n-1}\},\\[1mm]
e^{\mathrm{oth}}_{n-1}, & \text{if } X_n = B_{n-1},
\end{cases}
\end{align*}
with $e^{\mathrm{run}}_0=e^{\mathrm{oth}}_0=1$.

Equivalently, by aggregating the per-round factors,
\begin{align}
e^{\mathrm{run}}_n
&= 2^{M_n}\!\int \prod_{i=1}^n\theta_{i}^{\mathbf{1}\{X_i=A_{i-1}\}}(1-\theta_i)^{\mathbf{1}\{X_i=B_{i-1}\}}\,\Pi^{\mathrm{run}}_n(d\bm\theta),\label{eq:e_process_A_vs_B}\\
e^{\mathrm{oth}}_n
&= 2^{T_n}\!\int \prod_{i=1}^n\lambda_i^{\mathbf{1}\{X_i=A_{i-1}\}}(1-\lambda)_i^{\mathbf{1}\{X_i\notin \{A_{i-1},B_{i-1}\}\}}\,\Pi^{\mathrm{oth}}_n(d\bm\lambda),\label{eq:e_process_A_vs_OTHERS}
\end{align}
where $\Pi^{\mathrm{run}}_n$ (resp.\ $\Pi^{\mathrm{oth}}_n$) denotes a prior on the vector $\bm\theta$ (resp. $\bm\lambda$) and must be predictable, i.e. $\mathcal F_{n-1}$-measurable.
If $\Pi_n$ is a product distribution, we are re-mixing, i.e. not sharing information across steps.   If it is not a product distribution, we have the opportunity to be a bit more efficient.

The following theorem shows that the $e$-processes defined above provide anytime-valid tests.

\begin{theorem}[Anytime validity]\label{thm:mmc_eprocess_recursive}
Let $p_j=\mathbb P[X=j\mid pr]$. For the \emph{A vs B} test (leader vs runner-up), define
$\theta_n = \tfrac{p_{A_{n-1}}}{p_{A_{n-1}}+p_{B_{n-1}}}$ and the one-sided composite null
\[
H^{\mathrm{run}}_0:\quad \theta_n \le \tfrac12 \ \big(\text{equivalently $p_{A_{n-1}}\le p_{B_{n-1}}$}\big) \,\,\text{at every round $n$.}
\]
For the \emph{A vs others} test, define
$\lambda_n = \tfrac{p_{A_{n-1}}}{p_{A_{n-1}}+\sum_{j\notin\{A_{n-1},B_{n-1}\}}p_j} = \tfrac{p_{A_{n-1}}}{1-p_{B_{n-1}}}$
and the composite null
\[
H^{\mathrm{oth}}_0:\quad \lambda_n \le \tfrac12 \ \big(\text{equivalently $p_{A_{n-1}}\le \scriptstyle{\sum_{j\notin\{A_{n-1},B_{n-1}\}}}$$\, p_j$}\big) \,\,\text{at every round $n$.}
\]
Then $\{e^{\mathrm{run}}_n\}_{n\ge0}$ and $\{e^{\mathrm{oth}}_n\}_{n\ge0}$ defined in (\ref{eq:e_process_A_vs_B}), (\ref{eq:e_process_A_vs_OTHERS}) are non-negative test
\emph{supermartingales} w.r.t.\ $\{\mathcal F_n\}$, even with predictable, data-dependent priors and optional skipping.
Under the boundary (simple) nulls ($\theta_n\equiv\tfrac12$ or $\lambda_n\equiv\tfrac12$ on their informative rounds),
they are test \emph{martingales}. Consequently, by Ville’s inequality, for any stopping time,
\[
\sup_{\mathbb P\in H^{\mathrm{run}}_0}\ \mathbb P\Big(\sup_{n\ge0}e^{\mathrm{run}}_n\ge 1/\varepsilon\Big)\le\varepsilon,
\qquad
\sup_{\mathbb P\in H^{\mathrm{oth}}_0}\ \mathbb P\Big(\sup_{n\ge0}e^{\mathrm{oth}}_n\ge 1/\varepsilon\Big)\le\varepsilon.
\]
\end{theorem}
The proof is provided in Appendix \ref{subsec_app:mmc_recursive}.

\begin{corollary}[Union null for stopping]\label{cor:union_recursive}
Let $H_0:=H^{\mathrm{run}}_0\ \cup\ H^{\mathrm{oth}}_0$. Define the MMC stopping time
$N:=\inf\{n:\ e^{\mathrm{run}}_n\ge 1/\varepsilon\ \text{and}\ e^{\mathrm{oth}}_n\ge 1/\varepsilon\}$.
Then $\sup_{\,\mathbb P\in H_0}\mathbb P(N<\infty)\le \varepsilon$.
\end{corollary}

\begin{remark}[Why $o_n$ excludes $B_{n-1}$]
The A vs others null is $p_A \le \sum_{j\notin\{A,B\}}p_j$, which is equivalent to
$\lambda \le 1/2$ when we map successes to $X=A$ and failures to $X\notin\{A,B\}$.
Including $B$ among failures would test $p_A\le 1/2$ (absolute majority), which is unnecessarily strong.
\end{remark}

Pseudocode for implementing the MMC stopping rule is provided in Algorithm \ref{alg:stopping_rule}. If the maximum sample budget is reached, we return an upper bound $\hat{\varepsilon}$ on $\mathbb{P}[\hat{c}_n\neq c^\star]$. Details on how to compute $\hat{\varepsilon}$ are provided in Appendix \ref{app:subsec_estimator_proability}.

\begin{algorithm}[ht]
\small{
\caption{Martingale Majority Certificate stopping rule}
\label{alg:stopping_rule}
\begin{algorithmic}[1]
\Require confidence level $\varepsilon$, budget $N_{\text{budget}}$, prior hyperparameters; deterministic tie-break rule
\State \textbf{Init:} $n\gets 0$; for all $j\in\{1,\dots,k\}$ set label counts $N_j\gets 0$; $s_0=f_0=o_0\gets 0$; $e^{\mathrm{run}}_0=e^{\mathrm{oth}}_0\gets 1$
\While{True}
  \State \textbf{Predictable top-2:} set $A_n\gets\arg\max_j N_j$, $B_n\gets$ second largest (ties broken deterministically)
  \State \textbf{Cache counts (pre-update):} $\tilde s\gets s_n$, $\tilde f\gets f_n$, $\tilde o\gets o_n$
  \State \textbf{Draw a new vote:} sample $X\sim\mathbb P[\,\cdot\,| pr]$; \Comment{the only source of randomness per round}
  \State \textbf{Per-round ratio (A vs B):}
  \[
  \rho_{\mathrm{run}} \;=\;
  \scriptstyle\begin{cases}
 2\!\displaystyle\int \theta\,\pi^{\mathrm{run}}_n(d\theta), & X = A_{n},\\[1mm]
 2\!\displaystyle\int (1-\theta)\,\pi^{\mathrm{run}}_n(d\theta), & X = B_{n},\\[1mm]
1, & \text{otherwise,}
\end{cases}
  \]
  \State \textbf{Per-round ratio (A vs others):}
  \[
  \hspace{26pt}\rho_{\mathrm{oth}} \;=\;
\scriptstyle\begin{cases}
2\!\displaystyle\int \lambda\,\pi^{\mathrm{oth}}_n(d\lambda), & X = A_{n},\\[1mm]
 2\!\displaystyle\int (1-\lambda)\,\pi^{\mathrm{oth}}_n(d\lambda), & X \notin \{A_{n},B_{n}\},\\[1mm]
1, & \text{if } X = B_{n},
\end{cases}
  \]
  \State \textbf{Update $e$-values:} $e^{\mathrm{run}}_{n+1}\gets e^{\mathrm{run}}_{n}\cdot \rho_{\mathrm{run}}$, \ \ $e^{\mathrm{oth}}_{n+1}\gets e^{\mathrm{oth}}_{n}\cdot \rho_{\mathrm{oth}}$
  \State \textbf{Update recursive counts:}
  \[
  (s_{n+1},f_{n+1},o_{n+1})=
  \begin{cases}
    (\tilde s+1,\tilde f,\tilde o), & X=A_n,\\
    (\tilde s,\tilde f+1,\tilde o), & X=B_n,\\
    (\tilde s,\tilde f,\tilde o+1), & \text{otherwise.}
  \end{cases}
  \]
  \State \textbf{Update label counts:} $N_X\gets N_X+1$; \ $n\gets n+1$
  \State \textbf{Check stop:} \textbf{if} $e^{\mathrm{run}}_{n}\ge 1/\varepsilon$ \textbf{and} $e^{\mathrm{oth}}_{n}\ge 1/\varepsilon$ \textbf{then}
     \State \hspace{1.5em} set $\hat c\gets \arg\max_j N_j$; \Return $(\hat c,\ \text{stopped})$
  \State \textbf{Budget:} \textbf{if} $n\ge N_{\text{budget}}$ \textbf{then} \Return $(\arg\max_j N_j,\ \text{abstained})$
\EndWhile
\end{algorithmic}}
\normalsize
\end{algorithm}

\subsection{Two practical priors: truncated \texorpdfstring{$\mathrm{Beta}(a,b)$}{Beta(a,b)} and an updating point prior}
\label{subsec:mmc_priors}
We introduce two priors to compute the $e$-processes. Their performance is evaluated on synthetic data in Appendix \ref{app:subsec_mmc_synthetic_data}.

\paragraph{A. Truncated \texorpdfstring{$\mathrm{Beta}(a,b)$}{Beta(a,b)} prior on $(\tfrac12,1]$.}
For convenience, define the \emph{upper–half Beta mass}
\[
\mathsf{B}_{>1/2}(a,b)\;:=\;\int_{1/2}^1 t^{a-1}(1-t)^{b-1}\,dt\,.
\]
Here we use a \emph{single} latent parameter (shared across informative rounds), that is,
\begin{align*}
\Pi_n^{\mathrm{run}}(d\bm\theta)&\propto \theta^{a-1}(1-\theta)^{b-1}\mathbf 1\{\theta>1/2\}\prod_{i=1}^n\delta_\theta(d\theta_i),\\
\Pi_n^{\mathrm{oth}}(d\bm\lambda)&\propto \lambda^{a-1}(1-\lambda)^{b-1}\mathbf 1\{\lambda>1/2\}\prod_{i=1}^n\delta_\lambda(d\lambda_i).
\end{align*}
The mixture $e$-values admit closed forms in terms of $\mathsf{B}_{>1/2}$:
\[
\,e^{\mathrm{run}}_n
= 2^{M_n}\frac{\mathsf{B}_{>1/2}(a+s_n,\,b+f_n)}{\mathsf{B}_{>1/2}(a,b)}\,,\qquad
{\,e^{\mathrm{oth}}_n
= 2^{T_n}\frac{\mathsf{B}_{>1/2}(a+s_n,\,b+o_n)}{\mathsf{B}_{>1/2}(a,b)}\,}.
\]
These can be updated online by using \emph{ratios}:
\begin{align*}
\frac{e^{\mathrm{run}}_n}{e^{\mathrm{run}}_{n-1}}
&=
\begin{cases}
2\,\dfrac{\mathsf{B}_{>1/2}(a+s_{n-1}+1,\,b+f_{n-1})}{\mathsf{B}_{>1/2}(a+s_{n-1},\,b+f_{n-1})},
& X_n=A_{n-1},\\[2mm]
2\,\dfrac{\mathsf{B}_{>1/2}(a+s_{n-1},\,b+f_{n-1}+1)}{\mathsf{B}_{>1/2}(a+s_{n-1},\,b+f_{n-1})},
& X_n=B_{n-1},\\[2mm]
1,&\text{otherwise,}
\end{cases}\\
\frac{e^{\mathrm{oth}}_n}{e^{\mathrm{oth}}_{n-1}}
&=
\begin{cases}
2\,\dfrac{\mathsf{B}_{>1/2}(a+s_{n-1}+1,\,b+o_{n-1})}{\mathsf{B}_{>1/2}(a+s_{n-1},\,b+o_{n-1})},
& X_n=A_{n-1},\\[2mm]
2\,\dfrac{\mathsf{B}_{>1/2}(a+s_{n-1},\,b+o_{n-1}+1)}{\mathsf{B}_{>1/2}(a+s_{n-1},\,b+o_{n-1})},
& X_n\notin\{A_{n-1},B_{n-1}\},\\[2mm]
1,& X_n=B_{n-1}.
\end{cases}
\end{align*}
\emph{Recommended hyperparameters.} $a=b=\tfrac12$ (Jeffreys) or $a=b=1$ (Laplace) are robust defaults.
Truncation to $(1/2,1]$ ensures support under the one-sided alternative and yields the required
supermartingale property for the composite null via the boundary case $\theta=\tfrac12$ (resp. $\lambda=\tfrac12$).

\paragraph{B. Updating plug-in point prior.}
In this case, we share information across the two tests by maintaining a {single} plug–in estimate of the multinomial parameters for the predictable top–2 and the aggregated others.
\\\\
Fix smoothing hyperparameters $(\alpha_A,\alpha_B,\alpha_O)>0$ and set
\[
\hat p_{A,n} \;:=\; \tfrac{s_{n-1}+\alpha_A}{L_{n-1}+\alpha_A+\alpha_B+\alpha_O},\quad
\hat p_{B,n} \;:=\; \tfrac{f_{n-1}+\alpha_B}{L_{n-1}+\alpha_A+\alpha_B+\alpha_O},\quad
\hat p_{O,n} \;:=\; \tfrac{o_{n-1}+\alpha_O}{L_{n-1}+\alpha_A+\alpha_B+\alpha_O},
\]
where $L_{n-1}:=s_{n-1}+f_{n-1}+o_{n-1}$.
Define the  one–dimensional informative-round parameters
\[
\theta^\star_n \;:=\; \operatorname{clip}\!\left(\frac{\hat p_{A,n}}{\hat p_{A,n}+\hat p_{B,n}},\; \tfrac12+\varepsilon,\; 1-\varepsilon\right),
\qquad
\lambda^\star_n \;:=\; \operatorname{clip}\!\left(\frac{\hat p_{A,n}}{1-\hat p_{B,n}},\; \tfrac12+\varepsilon,\; 1-\varepsilon\right),
\]
where $\varepsilon\in(0,10^{-3}]$ ensures numerical stability.
We consider two different $e$-processes:
\begin{itemize}
    \item[(B.1)] Consider the shared-parameter priors
    \begin{align*}
\hspace{-30pt}\Pi_n^{\mathrm{run}}(d\bm\theta)&=\prod_{i=1}^n\delta_{\theta_n^\star}(d\theta_i),\qquad
\Pi_n^{\mathrm{oth}}(d\bm\lambda)=\prod_{i=1}^n\delta_{\lambda_n^\star}(d\lambda_i).
\end{align*}
    The corresponding mixture $e$-values are given by
\[
\hspace{-30pt}\,e^{\mathrm{run}}_n
= 2^{M_n}(\theta_n^\star)^{s_n}(1-\theta_n^\star)^{f_n},\qquad
{\,e^{\mathrm{oth}}_n
= 2^{T_n}(\lambda_n^\star)^{s_n}(1-\lambda_n^\star)^{o_n}\,}.
\]    
    \item[(B.2)] The second one is defined by its per-round update factors
    \[
\hspace{-30pt}\frac{e^{\mathrm{run}}_n}{e^{\mathrm{run}}_{n-1}}
=
\begin{cases}
2\,\theta^\star_n, & X_n=A_{n-1},\\
2\,(1-\theta^\star_n), & X_n=B_{n-1},\\
1, & \text{otherwise,}
\end{cases}
\qquad
\frac{e^{\mathrm{oth}}_n}{e^{\mathrm{oth}}_{n-1}}
=
\begin{cases}
2\,\lambda^\star_n, & X_n=A_{n-1},\\
2\,(1-\lambda^\star_n), & X_n\notin\{A_{n-1},B_{n-1}\},\\
1, & X_n=B_{n-1}.
\end{cases}
\]
\end{itemize}
By construction, $\theta^\star_n,\lambda^\star_n$ are $\mathcal F_{n-1}$–measurable and lie in $(\tfrac12,1]$ after clipping,
so Theorem~\ref{thm:mmc_eprocess_recursive} applies: $\{e^{\mathrm{run}}_n\}$ and $\{e^{\mathrm{oth}}_n\}$ are
non-negative test supermartingales under their respective composite nulls, and test martingales under the boundary nulls.
Ville’s inequality then yields time–uniform guarantees.  

\smallskip
\textbf{Heuristic sample complexity.}
If the informative–round parameter $\vartheta\in(\tfrac12,1)$ is well tracked by the plug–in estimate,
each $e$–process in (B.1) crosses $1/\varepsilon$ after roughly $\log(1/\varepsilon)/D_{\mathrm{KL}}(\mathrm{Ber}(\vartheta)\|\mathrm{Ber}(\tfrac12))$
informative draws. See Appendix \ref{app:subsec_stopping_time} for details. 

\section{Optimising sample efficiency through test-time training}
\label{sec:test_time_training_loss}
Our ultimate goal is to minimise the number of samples required from the LLM  for the majority vote to return the correct answer with high confidence $1-\varepsilon$. 
From the analysis in Section~\ref{sec:stopping_rule}, the
expected stopping time of the MMC scales approximately as
\begin{equation}\label{eq:expected_number_samples}
N \;\approx\;
\frac{2(p_{\hat c}+p_{j^\star})}{(p_{\hat c}-p_{j^\star})^{2}}
\,\log ({1}/{\varepsilon}),
\end{equation}
so that small mode margins \( \delta = p_{\hat c}-p_{j^\star} \) lead to rapidly increasing sample
requirements (see Appendix \ref{app:subsec_stopping_time} for details).  The key question is whether test-time adaptation can reshape the terminal distribution
to enlarge this margin, thereby improving sample efficiency.

\paragraph{Effect of test-time training.} 
Test-time reinforcement learning (TTRL; \citealp{zuo2025ttrl}) adapts model parameters at inference
time by maximising a KL-regularised objective based on self-generated rewards.  Given a prompt
\(pr\), let $(Y_t)_{t\geq 0}$ be the autoregressive token process from a reference distribution $\pi_{\text{ref}}(\cdot \, |\, pr)$ on trajectories.  Let $X = g(Y_{\tau:})$, where $\tau$ is the time at which the answer is generated, which is a (finite a.s.) stopping time with respect to the canonical filtration.
\\\\
Given $n$ trajectories $Y_1, \ldots, Y_n \sim \pi_{\text{ref}}$, yielding answers $X_1, \ldots, X_n$, let $\widehat{c}_n$ be the associated majority vote.  The reward introduced in \cite{zuo2025ttrl} is $r_n(Y_i) = \mathbf{1}\lbrace X_i = \widehat{c}_n\rbrace$.   The associated KL-regularised optimisation over trajectory laws parametrised by $\pi_{\phi} \ll \pi_{\text{ref}}$ is given by
\[
\max_{\phi}
\;
\mathbb{E}_{Y\sim\pi_\phi(\cdot|pr)}[r_n(Y)\,]
-\beta\,\KL(\pi_\phi\|\pi_{\mathrm{ref}}).
\]
The optimal policy is an \emph{exponentially tilted} distribution
\[
\pi^{\star}(Y|pr)
=\frac{e^{r_n(Y)/\beta}\,\pi_{\mathrm{ref}}(Y|pr)}
       {Z_\beta(pr)},\qquad
Z_\beta(pr)
=1+\pi_{\mathrm{ref}}(\widehat c_n|pr)\bigl(e^{1/\beta}-1\bigr),
\]
where the denominator is the normalising constant
\(Z_\beta=\mathbb{E}_{\pi_{\mathrm{ref}}}[e^{r_n(Y)/\beta}]\).
Writing \(\kappa=1/\beta\), the tilting sharpens the terminal law around the majority mode and
monotonically increases the signal-to-noise ratio (SNR) of the margin variable
\(\Delta_{j^\star_n}=\mathbf 1\{X=\widehat c_n\}-\mathbf 1\{X=j^\star_n\}\):
\[
\tfrac{d}{d\kappa}\mathrm{SNR}\bigl(\Delta_{j^\star_n}\bigr)(\kappa)\ge 0,
\]
with equality only if \(p_{\hat c_n}=1\), i.e. the distribution is a Dirac delta at the majority vote.
Strict monotonicity holds between values of~\(\kappa\) for which the runner-up~\(j^\star_n\) remains
fixed; at swap points the SNR function is continuous but non-differentiable.  Thus, increasing
\(\kappa\) (i.e.\ stronger tilting) consistently improves the margin and reduces the number of samples
required for certification.  See Appendix \ref{app:subsec_analysis_TTRL} for further details.

\paragraph{Two new test-time RL objectives.}
We introduce two label-free group-level rewards designed to optimise the trade-off between sharpness
and bias.  Let $\mathbf{X} = (X_1, \dots, X_n)$ be a set of answers arising from rollouts $\mathbf{Y} =(Y_1, \ldots, Y_n)$ for a given prompt, with $\widehat{c}_n$ denoting the majority vote and $j_n^\star$  the runner-up. Define $N_j= \scriptstyle\sum_i\mathbf{1}\{X_i=j\}$. 

\begin{enumerate}[label=(\roman*)]
\item \textbf{SNR-based reward.}
Directly leveraging the SNR as a driving factor in the efficiency of the MMC scheme we introduce the first reward
\begin{equation}\label{eq:snr_based_reward}
r^{(1)}_n(\mathbf{Y})
=\widehat{\mathrm{SNR}}(\Delta_{j^\star_n})(\mathbf{X})
=\tfrac{(N_{\widehat c_n}-N_{j^\star_n})^{2}}
       {n\left(N_{\widehat c_n}+N_{j^\star_n}\right)
        -(N_{\widehat c_n}-N_{j^\star_n})^{2}}
\;\xrightarrow[n\to\infty]{}\;
\mathrm{SNR}(\Delta_{j^\star_n}).
\end{equation}
This objective aims to directly maximise \(\text{SNR}(\Delta_{j_n^\star})\), which is equivalent to minimising the expected
number of samples required to obtain statistical certificates for the majority vote. 

\item \textbf{Entropy-based reward.}
As we want to encourage a more peaked terminal distribution, another natural option is negative entropy, i.e.
\begin{equation}\label{eq:entropy_based_reward}
r^{(2)}_n(\mathbf{Y})
=\widehat H_n(\mathbf{X})
=\sum_{j:N_j>0}\frac{N_j}{n}\log\frac{N_j}{n}
\;\xrightarrow[n\to\infty]{}\;
\sum_j p_j\log p_j=-H(p).
\end{equation}
Maximising \( \widehat H_n \)  \emph{minimises} the Shannon entropy of the answer
distribution, encouraging a sharper, lower-entropy distribution.
\end{enumerate}

Solving the corresponding KL-regularised variational problems (Appendices~\ref{app:details_new_TTT_loss}, \ref{app:subsec_analysis_TTTEntropy})
yields the respective optimisers.  
As with the TTRL tilt, \(\mathrm{SNR}(\Delta_{j^\star_n})\) is non-decreasing, implying that
sharper distributions require fewer samples for reliable certification.    
It is important to emphasise that our proposed entropy-based reward differs from that of \citep{agarwal2025unreasonableeffectivenessentropyminimization}. 
\\\\
The entropy reward $r_n^{(2)}$ should be understood as penalising entropy of the terminal distribution
of the trajectory distribution, not to the full trajectory law itself.
Formally, let $\pi_{\mathrm{ref}}(Y_{0:\tau})$ denote the reference
distribution over reasoning trajectories with terminal variable
$X=g(Y_{\tau:})$, and write
$p_{\mathrm{ref}}(x)=\pi_{\mathrm{ref}}(X=x)$ for its induced marginal.
Applying the KL chain rule,
\[
\KL(\pi_\phi\|\pi_{\mathrm{ref}})
=\KL(q\|p_{\mathrm{ref}})
+\E_{x\sim q}\!\big[\KL(\pi_\phi(\cdot|X=x)\|
   \pi_{\mathrm{ref}}(\cdot|X=x))\big],
\]
where $q(x)=\pi_\phi(X=x)$ is the terminal marginal of the adapted policy.
Because the entropy reward depends only on $X$, the second term is minimised
when $\pi_\phi(\cdot|X=x)=\pi_{\mathrm{ref}}(\cdot|X=x)$ for all $x$.  Hence, the KL-regularised variational problem over the base measure reduces to one over  the marginal $q$ alone:
\[
\max_{q\in\Delta(\mathcal X)} \;
   \big\{-H(q)-\beta \,\KL(q\|p_{\mathrm{ref}})\big\}.
\]
The unique maximiser of this objective is
$q^\star(x)\propto p_{\mathrm{ref}}(x)^{\kappa}$ with
$\kappa=\beta/(\beta-1)>1$.
Hence the test-time adaptation \emph{tempers the terminal marginal}
$p_{\mathrm{ref}}(x)$, while preserving the reference conditional trajectory
law $\pi_{\mathrm{ref}}(\cdot|X=x)$.  In particular,
\[
\pi_\phi^\star(Y_{0:\tau})
= \pi_{\mathrm{ref}}(Y_{0:\tau}\mid X)\,q^\star(X)
\;\neq\;
\frac{\pi_{\mathrm{ref}}(Y_{0:\tau})^{\kappa}}
     {\int \pi_{\mathrm{ref}}(Y_{0:\tau})^{\kappa}\,dY_{0:\tau}},
\]
except in the degenerate case where
$\pi_{\mathrm{ref}}(\cdot|X=x)$ is uniform for all $x$.
The tempering therefore sharpens only the distribution of final answers,
not the full sequence distribution.  This gives us the best of both worlds:  promoting certainty when providing a final answer, but permitting exploration of diverse pathways during the chain-of-thought reasoning process.  In particular, this should not be confused with \emph{low temperature scaling}, where the conditional next-token distributions of the full trajectory is tempered according to a temperature schedule \cite{wang2020contextual}.  
\\\\
Because the reward functions couple multiple variables, the corresponding gradient estimates can exhibit high variance. To reduce this variance, we adopt a leave-one-out control variate approach \citep{tang2025optimizing}, resulting in the following effective advantage functions for $Y_i$
\small
\begin{equation}\label{eq:effective_advantage_main_text}
A_i^{(1)} = \reallywidehat{\text{SNR}}(\Delta_{j_n^\star})(\mathbf X) - \reallywidehat{\text{SNR}}(\Delta_{j_n^\star})(\mathbf X_{-i}), \qquad A_i^{(2)} = \hat{H}_n(\mathbf X) - \hat{H}_{n-1}(\mathbf X_{-i}).
\end{equation}
\normalsize
This preserves unbiasedness and substantially reduce gradient variance in REINFORCE-style
optimisation.  
\\\\
We post-train our models using the {GRPO} algorithm \citep{shao2024deepseekmathpushinglimitsmathematical} for each of these rewards.  Details can be found in Appendix~\ref{app:details_test_time_training_loss}.   By contrast with the TTRL reward $r_n(Y)=1\{X=\hat{c}_n\}$, a benefit of both SNR- and entropy- based rewards is that these yield smoother signals of consensus.  In practice, this results in significantly faster and more stable convergence of the RL-loss function, consistent with similar observations made in \cite{ma2025general,tao2025hybrid}.

\section{SNR as a label-free estimator of task difficulty} 
\label{sec:snr-difficulty}

The preceding analysis establishes that signal-to-noise ratio  plays a governing role in certifying self-consistency, as well as in the associated  test-time
training objectives.  Given $n$ rollouts $\{Y_{i}\}_{i=1}^n$ from a prompt ${pr}$,
with terminal answers $X_i = g(Y_{i,\tau:})$, let
$\widehat c_n$ and $j^\star_n$ denote the empirical leader and runner-up.
We compute an empirical estimate of the SNR given by,
\begin{equation}\label{eq:snr_for_difficulty_estimation}
\widehat{\mathrm{SNR}}(\Delta_{j^\star_n})(\mathbf{X})
= \frac{(N_{\hat c_n} - N_{j^\star_n})^2}{
n(N_{\hat c_n} + N_{j^\star_n})
 - (N_{\hat c_n} - N_{j^\star_n})^2},
\end{equation}
where $N_j = \sum_i 1\{X_i = j\}$.
This statistic can be computed directly from model rollouts and requires no
access to external signals.
\\\\
In 
Figures~\ref{fig:QWEN-MATH-1.5B-SNR-0.1} and \ref{fig:QWEN-MATH-7B-SNR-0.1} we plot the estimated SNR values, generated over the MATH-500 benchmark against the reported problem level, with 1 being the easiest and 5 being the hardest,  \cite{lightman2023let}.   We observe that $\widehat{\mathrm{SNR}}$ values correlate strongly with
ground-truth difficulty levels:  harder problems exhibit systematically lower
SNR and greater variability.
This emergent calibration occurs without supervision: the model's own
epistemic uncertainty, quantified via SNR, consistently aligns with external difficulty
labels.   As values of $\widehat{\mathrm{SNR}}$ correspond to sharply peaked terminal
marginals in which the model consistently produces the same answer across
rollouts, we observe that ``easy'' prompts yield  high-SNR and thus low epistemic uncertainty.   Conversely, low SNR values arise for diffuse or multi-modal terminal
distributions, suggesting that reasoning models demonstrate uncertainty around harder or more ambiguous questions.   The observations align with previous works which seek to use uncertainty as a proxy for problem difficulty, \cite{wang2025make, wan2025reasoning,fu2025deep}, with the aim of dynamically allocating resources.

\section{Numerical experiments}\label{sec:numerical_experiments}
The goal of this section is threefold: (1) to evaluate the performance of our proposed test-time RL objectives (Section \ref{sec:test_time_training_loss}), (2) to empirically demonstrate that inference-time training strategies reduce the number of samples required by the MMC stopping rule (Algorithm \ref{alg:stopping_rule}) to obtain statistical certificates, compared to pre-trained models, and (3) to show that the SNR serves as a label-free proxy for problem difficulty. Additional experimental details are provided in Appendix~\ref{app:numerical_experiments_details}.
\subsection{Experimental setup}\label{subsec:experimental_setup}

\paragraph{Models and benchmarks.}
We use both base and instruct models of various scales, specifically Qwen2.5-Math-1.5B, Qwen2.5-Math-7B \citep{yang2024qwen25mathtechnicalreportmathematical}, Qwen2.5-7B \citep{qwen2025qwen25technicalreport} and LLaMA-3.1-8B-Instruct \citep{grattafiori2024llama3herdmodels}.
We consider three mathematical reasoning benchmarks: AIME 2024, AMC \citep{li2024numinamath}, and MATH-500 \citep{hendrycks2021measuring}.

\paragraph{Methods and evaluation.}
For test-time training, we use the VERL framework \citep{sheng2024hybridflow} with the GRPO algorithm \citep{shao2024deepseekmathpushinglimitsmathematical} on 8$\times$H100 Nvidia GPUs. We apply our method to each benchmark individually and report both pass@1 and majority-vote accuracy (see Appendix \ref{app:numerical_experiments_details}). We compare the performance of our proposed RL objectives with TTRL \citep{zuo2025ttrl}.

\subsection{Results}\label{subsec:experimental_results}
Table \ref{tab:test-time-training-results} reports the pass@1 performance of various inference-time training strategies across different benchmarks and models. An extended version, comparing the improvements in pass@1 accuracy for both the score and format score, is provided in Table \ref{tab:test-time-training-results-pass1-format-score} (Appendix \ref{app:numerical_experiments_details}).
Overall, these strategies consistently enhance pass@1 performance, with the effect being particularly pronounced for Qwen2.5-Math-1.5B, the smallest model.
This suggest that such test-time methods help reveal the model’s underlying capabilities.

\begin{table}[b!]
\caption{Comparison of pass@1 performance before and after applying test-time training strategies.}
\vspace{-5pt}
\label{tab:test-time-training-results}
\begin{center}
\footnotesize
\begin{minipage}{0.47\linewidth}
\centering
\begin{tabular}{lccc}
\toprule
 & \textbf{AIME} & \textbf{AMC} & \textbf{Math\scriptsize-500\footnotesize} \\
\midrule
\textbf{Qwen2.5-7B} & 9.4 & 31.2 &  59.1 \\
SNR (Ours)& 23.3 &  51.8& 80.3 \\
Entropy (Ours)& 20.0 & 49.2 & 77.6 \\
\citet{zuo2025ttrl} & 24.3 & 53.4 & 79.6  \\
\midrule
\textbf{Llama-3.1-8B} & 4.4 & 21.8 & 48.2  \\
SNR (Ours)& 13.4 & 29.3 & 59.2 \\
Entropy (Ours)& 13.3 & 27.0  &  55.4 \\
\citet{zuo2025ttrl} & 10.0 & 32.3 & 63.7 \\
\bottomrule
\end{tabular}
\end{minipage}
\hfill
\begin{minipage}{0.47\linewidth}
\footnotesize
\centering
\begin{tabular}{lccc}
\toprule
 & \textbf{AIME} & \textbf{AMC} & \textbf{Math\scriptsize-500\footnotesize} \\
\midrule
\textbf{Qwen2.5-Math-7B} & 10.6 & 31.0 & 47.1\\
SNR (Ours) & 36.7 & 65.0 & 84.5 \\
Entropy (Ours) & 38.3 &  65.4 & 82.4 \\
\citet{zuo2025ttrl} & 37.9 & 63.5 &  83.6\\
\midrule
\textbf{Qwen2.5-Math-1.5B} & 7.1 & 28.1 & 31.4 \\
SNR (Ours)& 16.3  & 45.4 & 72.0 \\
Entropy (Ours)& 15.6  & 45.9 & 70.8 \\
\citet{zuo2025ttrl} & 15.8 & 48.4 & 71.9 \\
\bottomrule
\end{tabular}
\end{minipage}
\end{center}
\vspace{-10pt}
\end{table}
\normalsize

Besides, we analyse how test-time training reduces the number of samples required to guarantee, with high confidence, that the majority vote $\widehat{c}_n$  matches the true mode $c^\star$. 
Specifically, Table \ref{tab:stopping-rule-samples-results} reports the majority vote accuracy and the required number of samples under the MMC stopping rule $(N_{\text{adaptive}})$ for two confidence levels, $\varepsilon = 0.1$ and $0.4$, comparing the pre-trained model with the model after test-time training using SNR-based rewards. 
For reference, we also include the majority vote accuracy obtained when using the full sample budget $N_{\text{budget}}$.

We observe that the MMC adaptive sampling scheme substantially reduces the number of samples without causing a noticeable degradation in performance. Moreover, the number of samples required under the MMC stopping rule further decreases after applying test-time training, relative to the pre-trained model. 
This effect is examined in more detail in Table~\ref{tab:ratio_adaptive_sampling_pre_post_trained}, which reports the reduction in the ratio between $N_{\text{adaptive}}$ and $N_{\text{budget}}$ (given their approximately linear relationship). 
The decrease in this ratio after test-time training is most pronounced for the smaller 1.5B model.
Improving sample efficiency is particularly important, as it directly translates to lower inference costs.

\begin{figure}[h!]
  \centering
  \begin{subfigure}{0.49\textwidth}
      \centering
      \includegraphics[width=\textwidth]{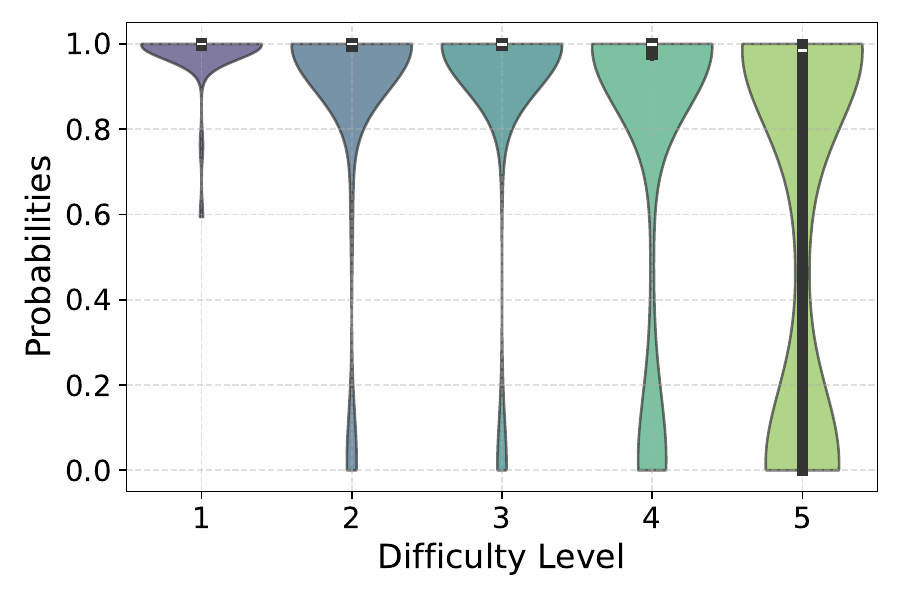}
      \caption{Qwen2.5-Math-1.5B, $\mathbb{P}[\widehat{c}_n = c^\star]$.}
      \label{fig:QWEN-MATH-1.5B-probs-0.1}
  \end{subfigure}
  \hfill
  \begin{subfigure}{0.49\textwidth}
      \centering
      \includegraphics[width=\textwidth]{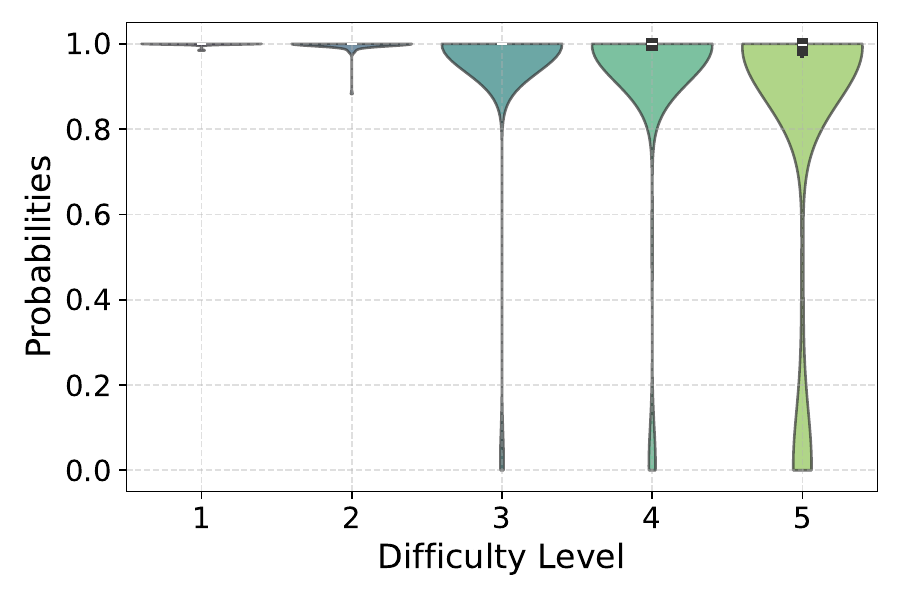}
        \caption{Qwen2.5-Math-7B, $\mathbb{P}[\widehat{c}_n = c^\star]$.}
      \label{fig:QWEN-MATH-7B-probs-0.1}
  \end{subfigure}
  \vfill
  \begin{subfigure}{0.49\textwidth}
      \centering
      \includegraphics[width=\textwidth]{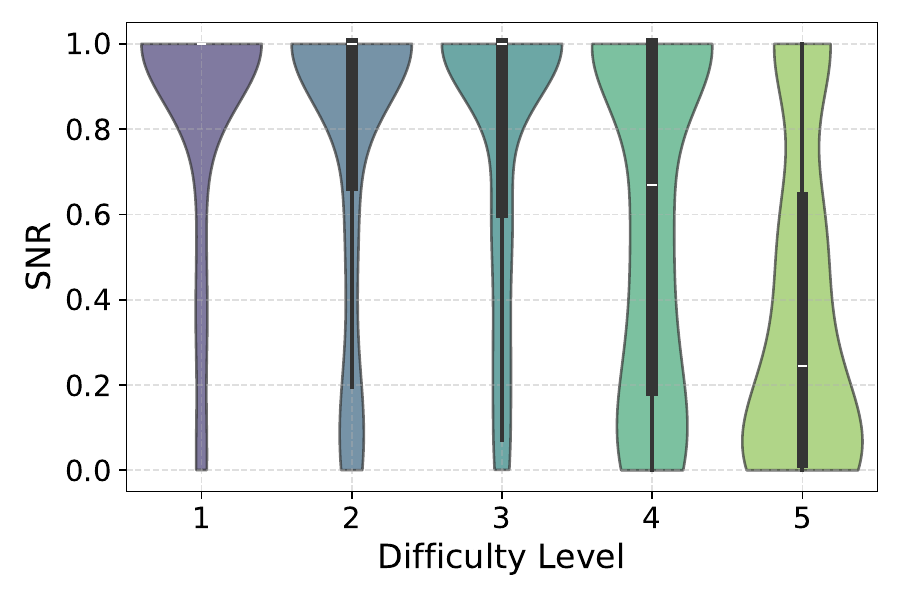}
      \caption{Qwen2.5-Math-1.5B, ${\text{SNR}}(\Delta_{j^\star_n})$.}
      \label{fig:QWEN-MATH-1.5B-SNR-0.1}
  \end{subfigure}
  \hfill
  \begin{subfigure}{0.49\textwidth}
      \centering
      \includegraphics[width=\textwidth]
      {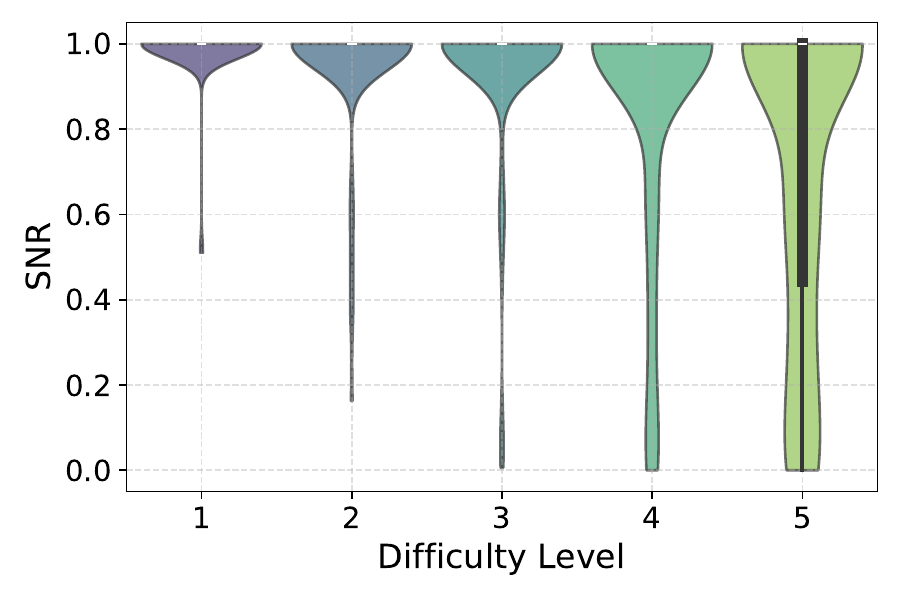}      
        \caption{Qwen2.5-Math-7B, ${\text{SNR}}(\Delta_{j^\star_n})$.}
      \label{fig:QWEN-MATH-7B-SNR-0.1}
  \end{subfigure}
  \caption{Distribution of the estimated lower bound of the probability $\mathbb{P}[\widehat{c}_n = c^\star]$ (computed via Beta approximations) and the signal-to-noise ratio $\text{SNR}(\Delta_{j^\star_n})$ when applying the MMC stopping rule with $\varepsilon = 0.1$ and $N_{\text{budget}}=100$. Results are obtained after test-time training with SNR-based rewards on the MATH-500 dataset.}
  \label{fig:violin_plots_main_text}
\end{figure}

Finally, since the MATH-500 dataset classifies questions into five levels of increasing difficulty, we analyse the distributions of the estimated lower bound of the probability $\mathbb{P}[\widehat{c}_n = c^\star]$, as well as the estimated signal-to-noise ratio $\reallywidehat{\text{SNR}}(\Delta_{j_n^\star})$ across these difficulty levels.
Figure~\ref{fig:violin_plots_main_text} shows that harder questions exhibit greater variability for both $\mathbb{P}[\widehat{c}_n = c^\star]$ and the SNR. 
In addition,  for the smaller 1.5B model, both the  probabilities and SNR distributions are more concentrated near zero for difficult questions compared to the 7B model.
These observations further support the idea that the SNR can serve as a label-free proxy for problem difficulty.

\begin{table}[t!]
\caption{Comparison of majority vote accuracy and required number of samples under the MMC stopping rule $(\mathbf{N}_{\text{adaptive}})$  for $\varepsilon = 0.1$ and $0.4$ between the pre-trained model and after test-time training with SNR-based rewards. Performance is compared to that obtained using the full sample budget ${N_{\text{budget}}}$ (\textcolor{red}{\xmark}). Results are shown for the MATH-500 dataset.}
\label{tab:stopping-rule-samples-results}
\begin{center}
\footnotesize
\begin{tabular}{lccccc}
\toprule
\multirow{3}{*}{$\mathbf{N_{\text{budget}}}$}& \multirow{3}{*}{\makecell{\textbf{Adaptive} \\ \textbf{sampling?}}} & \multicolumn{2}{c}{\textbf{Qwen2.5-Math-7B}}& \multicolumn{2}{c}{\textbf{Qwen2.5-Math-1.5B}}\\
\cmidrule{3-6}
& & \textbf{Pre-trained}& \textbf{Test-time trained} & \textbf{Pre-trained}& \textbf{Test-time trained}\\
& & \textbf{\%} \scriptsize$(\mathbf{N}_{\text{adaptive}})$& \textbf{\%} \scriptsize$(\mathbf{N}_{\text{adaptive}})$ & \textbf{\%} \scriptsize$(\mathbf{N}_{\text{adaptive}})$& \textbf{\%} \scriptsize$(\mathbf{N}_{\text{adaptive}})$\\
\midrule
\multirow{5}{*}{\textbf{10}} &\multirow{1}{*}{\textcolor{red}{\xmark}} &61.6&85.2&36.0 & 78.6\\
&\greencheck& \multirow{2}{*}{61.6 \scriptsize\textbf{(9.7)}}& \multirow{2}{*}{85.2 \scriptsize\textbf{(9.4)}}& \multirow{2}{*}{36.0 \scriptsize\textbf{(9.9)}} & \multirow{2}{*}{78.6 \scriptsize\textbf{(9.4)}}\\
& $\varepsilon = 0.1$&  &  & &\\
&\greencheck& \multirow{2}{*}{61.6 \scriptsize\textbf{(9.2)}}& \multirow{2}{*}{85.2 \scriptsize\textbf{(8.9)}}& \multirow{2}{*}{36.0 \scriptsize\textbf{(9.7)}} & \multirow{2}{*}{78.6 \scriptsize\textbf{(8.6)}}\\
& $\varepsilon = 0.4$& &  &  &\\
\midrule
\multirow{5}{*}{\textbf{50}} &\multirow{1}{*}{\textcolor{red}{\xmark}} & 62.2& 85.6& 37.6& 80.8\\
&\greencheck& \multirow{2}{*}{61.8 \scriptsize\textbf{(39.3)}}& \multirow{2}{*}{85.6 \scriptsize\textbf{(37.6)}}& \multirow{2}{*}{37.6 \scriptsize\textbf{(45.6)}} & \multirow{2}{*}{80.8 \scriptsize\textbf{(34.1)}}\\
& $\varepsilon = 0.1$&  &&  &\\
&\greencheck& \multirow{2}{*}{61.8 \scriptsize\textbf{(38.0)}}& \multirow{2}{*}{85.4 \scriptsize\textbf{(33.4)}}& \multirow{2}{*}{37.4 \scriptsize\textbf{(43.0)}} & \multirow{2}{*}{80.8 \scriptsize\textbf{(31.2)}}\\
& $\varepsilon = 0.4$&& & &\\
\midrule
\multirow{5}{*}{\textbf{100}} &\multirow{1}{*}{\textcolor{red}{\xmark}} & 62.2 & 85.6&\multirow{1}{*}{36.6} & 81.2\\
&\greencheck&  \multirow{2}{*}{62.2 \scriptsize\textbf{(74.9)}}& \multirow{2}{*}{85.6 \scriptsize\textbf{(67.2)}}& \multirow{2}{*}{36.8 \scriptsize\textbf{(86.5)}} &\multirow{2}{*}{81.0 \scriptsize\textbf{(61.2)}}\\
& $\varepsilon = 0.1$&  & &  &\\
&\greencheck& \multirow{2}{*}{62.2 \scriptsize\textbf{(73.1)}} & \multirow{2}{*}{85.4 \scriptsize\textbf{(60.8)}}& \multirow{2}{*}{36.4 \scriptsize\textbf{(81.8)}} &\multirow{2}{*}{80.8 \scriptsize\textbf{(56.9)}}  \\
& $\varepsilon = 0.4$& &  &  &\\
\bottomrule
\end{tabular}
\end{center}
\end{table}
\normalsize

\section{Related work}\label{sec:related_work}
\paragraph{Classical majority aggregation.}  
The study of majority voting as a mechanism for error reduction dates back to Condorcet’s jury theorem, which shows that under independence and competence above chance, majority aggregation recovers the correct decision with probability approaching one as the ensemble size grows \citep{condorcet1785essai}. Subsequent work has analysed correlated jurors \citep{ladha1992condorcet}, multiclass outcomes \citep{list2001epistemic}, and asymptotic behaviour \citep{boland1989majority}.  Concentration inequalities have long been used to control majority error in the binary case, providing simple finite-sample bounds on the probability of incorrect aggregation.   In this work, we build on these results with the aim of systematically understanding the multinomial setting relevant for LLM outputs, and to reinterpret the resulting bounds explicitly as \emph{certificates} of model reliability.  Various extensions to Condorcet's original formalism have been considered.   A closely related line of work models heterogeneous and possibly biased voters via the Dawid–Skene framework \citep{dawid_skene_79}, which introduces latent {confusion matrices} for each voter, estimating them via Expectation-Maximisation. This generalises majority vote to settings with unequal competence and asymmetric errors in the multiclass case. Subsequent extensions incorporate item difficulty and worker ability, yielding models akin to Item Response Theory \citep{bock1997brief}, Bayesian treatments and priors over confusion matrices \citep{raykar2010learning,liu2012variational,kim2012bayesian}.   These frameworks have been leveraged in the context of LLMs, both for assessing quality of data annotation e.g. \cite{whitehill2009whose,welinder2010multidimensional}, as well as for aggregation and combination of outputs from heterogeneous models \citep{yao2024bayesian, song2025irt}, or for uncertainty quantification \citep{kang2025uncertainty}.  Adapting our  anytime statistical certificates in these more general settings will be the scope of future work. 

\paragraph{Self-consistency and ensembles in LLMs.}  
In the context of chain-of-thought (CoT) prompting, majority voting is widely known as \emph{self-consistency} \citep{wang2022selfconsistency}.
By sampling multiple reasoning trajectories and returning the empirical mode, self-consistency significantly improves accuracy on reasoning benchmarks.  Extensions include iterative refinement and self-feedback loops
\citep{madaan2023selfrefine,shinn2023reflexion} and ensemble-style aggregation in large-scale systems such as PaLM~2 \citep{anil2023palm} and GPT-4~\citep{openai2023gpt4}.
These approaches demonstrate empirically that aggregation mitigates stochasticity in reasoning and that the marginal benefit of additional samples is highly instance-dependent.

More recent work has begun to address this dependency explicitly through
\emph{adaptive self-consistency}, where the number of sampled trajectories is
determined dynamically through a stopping rule, informed by model uncertainty or rollout agreement. \citep{aggarwal2023let,liescape, wan2025reasoning}.   
Difficulty-adaptive sampling schemes
\citep{wang2025make}
and early-stopping strategies such as
\emph{Self-Truncation Best-of-$N$} (ST-BoN; \cite{wang2025sampling})
aim to minimise test-time compute while maintaining accuracy by
halting when the vote distribution stabilises.
Related adaptive compute frameworks learn to predict, mid-generation, whether
further sampling would change the outcome
\citep{manvi2024adaptive,liu2024speculative},
thereby allocating more samples to difficult or ambiguous prompts and fewer to easy ones.
\\\\
While the above adaptive self-consistency strategies share the same goal of
halting rollouts when the empirical vote distribution stabilises, they provide no formal
control over reliability.  Our Martingale Majority Certificate (MMC) makes this
principle explicit by framing aggregation as an \emph{anytime-valid} hypothesis
test through $e$-values \citep{STA-002}.  This guarantees uniform, finite-sample error control for all stopping times,
offering a statistically grounded analogue to these heuristic adaptive-sampling
strategies.

\paragraph{Test-time training and reinforcement learning.}  
A complementary line of work has investigated \emph{test-time adaptation}, in which the model is updated online at inference time. Early approaches include entropy minimisation and self-training in computer vision. More recently, test-time reinforcement learning (TTRL) has been introduced for LLMs, where the model is adapted by optimising KL-regularised objectives with respect to its own rollouts \citep{zuo2025ttrl}. Related methods such as \cite{akyurek2025ttt} and \cite{prasadself} similarly adapt models at inference time to sharpen predictions and improve reliability.   Similarly, \cite{wen2025unsupervised}, propose a method called Internal Coherence Maximization (ICM), which fine-tunes pretrained language models without any external labels by maximising mutual predictability and logical consistency among the model's own generated labels.  In \cite{prabhudesai2025maximizing} and \cite{kang2025scalable} the authors use token-level negative entropy as a reward signal for test-time reinforcement learning.  Finally, \cite{shafayat2025large} explores RL post-training leveraging a consensus reward identical to \cite{zuo2025ttrl}, but without KL-regularisation with respect to the base measure,  demonstrating it can generate measurable improvements, before the inevitable collapse.

While these approaches empirically demonstrate measurable improvements, their mechanism has not been theoretically clarified. Firstly, our analysis provides a unifying perspective: KL-regularised TTRL objectives correspond to exponential tilting of the terminal distribution, and entropy-penalising rewards are equivalent to marginal tempering. This explains why such methods increase the mode margin and thereby reduce the number of samples required for certification.   Secondly, our work clarifies the essential role played by the KL-regularisation, without which the model eventually collapses under post-training.

\section{Discussion}\label{sec:discussion}
Our results combine several strands of recent work on reliable inference in LLMs, self-consistency,
adaptive compute allocation, and test-time reinforcement learning (TTRL), under a common
statistical perspective.  Through this lens, majority voting emerges naturally as a means of estimating the mode of the terminal distribution.  The validity of the majority vote as an estimate of the mode can be  {certified} by finite-sample and asymptotic bounds. The Martingale Majority Certificate (MMC)
extends this view by providing an {operational} test-time algorithm that determines, from model
rollouts alone, when a response is statistically self-consistent.  This recasts test-time scaling as a sequential decision problem with formal coverage guarantees, contrasting with heuristic
stopping rules based on agreement or entropy thresholds.
\\\\
Our analysis clarifies the underlying mechanism by which TTRL and related post-training
approaches improve reasoning reliability: KL-regularised optimisation corresponds to an
{exponential tilting} of the terminal law, sharpening it around its mode and increasing the
signal-to-noise ratio (SNR) of the margin variable, as has been observed in similar settings \citep{yang2024asymptotics} for verifier based approaches.  This insight explains empirical observations of
enhanced consistency after test-time adaptation, and motivates new label-free objectives such as
our SNR- and entropy-based rewards, which explicitly target this trade-off between sharpness and bias. 
\\\\
Beyond immediate applications to reasoning benchmarks, our framework offers a principled path
toward {certifiable reliability} in language models.  By unifying classical concentration theory,
martingale testing, and reinforcement-style post-training within one formal structure, we obtain
statistical interpretability for inference-time adaptation.  This could extend naturally to multi-agent
ensembles, verifier–generator systems, and other domains where LLMs operate under uncertainty.
Future work will explore applying anytime-valid certificates to correlated rollouts, structured output
spaces, and multi-verifier settings, as well as combining them with learned difficulty estimators for
dynamic compute scheduling.   
\\\\
We have also demonstrated the efficacy of anytime-valid certificates in the simplified setting of problems with discrete, multiple-choice outputs.   It is worth emphasising that the MMC does not require a-priori knowledge of the set of possible answers. While this would enable one to apply similar approaches to free-text answers, this would still require some degree of response canonicalisation.  Future work will explore alternative reformulations of MMC, which circumvent the need for `binning' similar responses, while still providing statistical certificates.

\section{Limitations}\label{app:limitations}
Our analysis assumes conditionally independent rollouts given a fixed prompt
and context, corresponding exactly to standard stochastic decoding (e.g.\
temperature or nucleus sampling).  This assumption holds for the inference
regime considered here, where each completion is sampled independently from the
model’s conditional distribution, but future extensions could address adaptive
or verifier-guided sampling strategies that introduce dependencies across
rollouts.  A second limitation concerns calibration: our SNR- and
entropy-based quantities rely on the model’s internal probabilities to reflect
true epistemic uncertainty, which may not hold for all models or decoding
temperatures.  Empirically, our evaluation focuses on single-turn reasoning
benchmarks; applying the framework to multi-turn dialogue, program synthesis,
and structured prediction remains an open direction.  Although
anytime-valid stopping improves expected efficiency, generating multiple
trajectories still incurs substantial compute cost.  Future work will explore
correlated-rollout models, calibration corrections, and hierarchical extensions
to improve the robustness and scalability of certified reasoning. 

\bibliography{references}

\begin{thebibliography}{74}
\providecommand{\natexlab}[1]{#1}
\providecommand{\url}[1]{\texttt{#1}}
\expandafter\ifx\csname urlstyle\endcsname\relax
  \providecommand{\doi}[1]{doi: #1}\else
  \providecommand{\doi}{doi: \begingroup \urlstyle{rm}\Url}\fi

\bibitem[Agarwal et~al.(2025)Agarwal, Zhang, Yuan, Han, and Peng]{agarwal2025unreasonableeffectivenessentropyminimization}
Shivam Agarwal, Zimin Zhang, Lifan Yuan, Jiawei Han, and Hao Peng.
\newblock {The Unreasonable Effectiveness of Entropy Minimization in LLM Reasoning}.
\newblock \emph{arXiv preprint arXiv:2505.15134}, 2025.

\bibitem[Aggarwal et~al.(2023)Aggarwal, Madaan, Yang, et~al.]{aggarwal2023let}
Pranjal Aggarwal, Aman Madaan, Yiming Yang, et~al.
\newblock {Let's Sample Step by Step: Adaptive-Consistency for Efficient Reasoning and Coding with LLMs}.
\newblock In \emph{The 2023 Conference on Empirical Methods in Natural Language Processing}, 2023.

\bibitem[Aky{\"u}rek et~al.(2025)Aky{\"u}rek, Damani, Zweiger, Qiu, Guo, Pari, Kim, and Andreas]{akyurek2025ttt}
Ekin Aky{\"u}rek, Mehul Damani, Adam Zweiger, Linlu Qiu, Han Guo, Jyothish Pari, Yoon Kim, and Jacob Andreas.
\newblock {The Surprising Effectiveness of Test-Time Training for Few-Shot Learning}.
\newblock In \emph{Proceedings of the 42nd International Conference on Machine Learning (ICML)}, 2025.

\bibitem[Anil et~al.(2023)Anil, Chi, Chowdhery, and et~al.]{anil2023palm}
Rohan Anil, Ed~H Chi, Aakanksha Chowdhery, and et~al.
\newblock {PaLM 2 Technical Report}.
\newblock \emph{arXiv preprint arXiv:2305.10403}, 2023.

\bibitem[Bahadur \& Rao(1960)Bahadur and Rao]{bahadur-rao}
R.~R. Bahadur and R.~Ranga Rao.
\newblock {On Deviations of the Sample Mean}.
\newblock \emph{The Annals of Mathematical Statistics}, 31\penalty0 (4):\penalty0 1015--1027, 1960.

\bibitem[Beirami et~al.()Beirami, Agarwal, Berant, D'Amour, Eisenstein, Nagpal, and Suresh]{beiramitheoretical}
Ahmad Beirami, Alekh Agarwal, Jonathan Berant, Alexander~Nicholas D'Amour, Jacob Eisenstein, Chirag Nagpal, and Ananda~Theertha Suresh.
\newblock Theoretical guarantees on the best-of-n alignment policy.
\newblock In \emph{Forty-second International Conference on Machine Learning}.

\bibitem[Bock(1997)]{bock1997brief}
R~Darrell Bock.
\newblock A brief history of item theory response.
\newblock \emph{Educational measurement: issues and practice}, 16\penalty0 (4):\penalty0 21--33, 1997.

\bibitem[Boland(1989)]{boland1989majority}
Philip~J. Boland.
\newblock {Majority Systems and the Condorcet Jury Theorem}.
\newblock \emph{Journal of the Royal Statistical Society. Series D (The Statistician)}, 38\penalty0 (3):\penalty0 181--189, 1989.
\newblock ISSN 00390526, 14679884.
\newblock URL \url{http://www.jstor.org/stable/2348873}.

\bibitem[Boucheron et~al.(2013)Boucheron, Lugosi, and Massart]{boucheronconcentration2013}
Stéphane Boucheron, Gábor Lugosi, and Pascal Massart.
\newblock \emph{Concentration Inequalities: A Nonasymptotic Theory of Independence}.
\newblock Oxford University Press, 2013.

\bibitem[Brown et~al.(2020)Brown, Mann, Ryder, Subbiah, Kaplan, Dhariwal, Neelakantan, Shyam, Sastry, Askell, et~al.]{brown2020gpt3}
Tom~B. Brown, Benjamin Mann, Nick Ryder, Melanie Subbiah, Jared Kaplan, Prafulla Dhariwal, Arvind Neelakantan, Pranav Shyam, Girish Sastry, Amanda Askell, et~al.
\newblock Language models are few-shot learners.
\newblock \emph{Advances in Neural Information Processing Systems}, 33, 2020.

\bibitem[Chan et~al.(2025)Chan, Nanni, Lazauskas, Wood, Yong, Tarassenko, Girolami, Geddes, and Duncan]{chan2025lean}
Ryan Sze-Yin Chan, Federico Nanni, Tomas Lazauskas, Rosie Wood, Penelope Yong, Lionel Tarassenko, Mark Girolami, James Geddes, and Andrew Duncan.
\newblock Retrieval-augmented reasoning with lean language models.
\newblock \emph{The Alan Turing Institute}, 2025.
\newblock \doi{10.5281/ZENODO.16408412}.

\bibitem[Cobbe et~al.(2021)Cobbe, Kosaraju, Bavarian, Chen, Jun, Kaiser, Plappert, Tworek, Hilton, Nakano, Hesse, and Schulman]{cobbe2021gsm8k}
Karl Cobbe, Vineet Kosaraju, Mohammad Bavarian, Mark Chen, Heewoo Jun, Lukasz Kaiser, Matthias Plappert, Jerry Tworek, Jacob Hilton, Reiichiro Nakano, Christopher Hesse, and John Schulman.
\newblock Training verifiers to solve math word problems.
\newblock \emph{arXiv preprint arXiv:2110.14168}, 2021.

\bibitem[Coulom(2007)]{coulom_mcts_2007}
R{\'e}mi Coulom.
\newblock Efficient selectivity and backup operators in monte-carlo tree search.
\newblock In H.~Jaap van~den Herik, Paolo Ciancarini, and H.~H. L. M.~(Jeroen) Donkers (eds.), \emph{Computers and Games}, pp.\  72--83, Berlin, Heidelberg, 2007. Springer Berlin Heidelberg.

\bibitem[Dawid \& Skene(1979)Dawid and Skene]{dawid_skene_79}
A.~P. Dawid and A.~M. Skene.
\newblock {Maximum Likelihood Estimation of Observer Error-Rates Using the EM Algorithm}.
\newblock \emph{Journal of the Royal Statistical Society. Series C (Applied Statistics)}, 28\penalty0 (1):\penalty0 20--28, 1979.

\bibitem[de~Condorcet(1785)]{condorcet1785essai}
Marie Jean Antoine Nicolas~Caritat de~Condorcet.
\newblock \emph{Essai sur l'application de l'analyse \`a la probabilit\'e des d\'ecisions rendues \`a la pluralit\'e des voix}.
\newblock Imprimerie Royale, Paris, 1785.
\newblock Reprint: AMS Chelsea, 1972.

\bibitem[Dembo \& Zeitouni(2010)Dembo and Zeitouni]{dembo2010ldp}
Amir Dembo and Ofer Zeitouni.
\newblock \emph{{Large Deviations Techniques and Applications}}.
\newblock Springer Berlin, Heidelberg, 2010.

\bibitem[Fu et~al.(2025)Fu, Wang, Tian, and Zhao]{fu2025deep}
Yichao Fu, Xuewei Wang, Yuandong Tian, and Jiawei Zhao.
\newblock Deep think with confidence.
\newblock \emph{arXiv preprint arXiv:2508.15260}, 2025.

\bibitem[Grattafiori et~al.(2024)Grattafiori, Dubey, Jauhri, Pandey, Kadian, Al-Dahle, Letman, Mathur, Schelten, Vaughan, Yang, et~al.]{grattafiori2024llama3herdmodels}
Aaron Grattafiori, Abhimanyu Dubey, Abhinav Jauhri, Abhinav Pandey, Abhishek Kadian, Ahmad Al-Dahle, Aiesha Letman, Akhil Mathur, Alan Schelten, Alex Vaughan, Amy Yang, et~al.
\newblock {The Llama 3 Herd of Models}.
\newblock \emph{arXiv preprint arXiv:2407.21783}, 2024.

\bibitem[Gui et~al.(2024)Gui, G{\^a}rbacea, and Veitch]{gui2024bonbon}
Lin Gui, Cristina G{\^a}rbacea, and Victor Veitch.
\newblock {BoNBoN alignment for large language models and the sweetness of best-of-n sampling}.
\newblock \emph{Advances in Neural Information Processing Systems}, 37:\penalty0 2851--2885, 2024.

\bibitem[Hendrycks et~al.(2021)Hendrycks, Burns, Kadavath, Arora, Basart, Tang, Song, and Steinhardt]{hendrycks2021measuring}
Dan Hendrycks, Collin Burns, Saurav Kadavath, Akul Arora, Steven Basart, Eric Tang, Dawn Song, and Jacob Steinhardt.
\newblock Measuring mathematical problem solving with the {MATH} dataset.
\newblock In \emph{Thirty-fifth Conference on Neural Information Processing Systems Datasets and Benchmarks Track (Round 2)}, 2021.

\bibitem[Howard et~al.(2021)Howard, Ramdas, McAuliffe, and Sekhon]{howard2021confidenceseq}
Steven~R Howard, Aaditya Ramdas, Jon McAuliffe, and Jasjeet Sekhon.
\newblock Time-uniform, nonparametric, nonasymptotic confidence sequences.
\newblock \emph{The Annals of Statistics}, 49\penalty0 (2):\penalty0 1055--1080, 2021.

\bibitem[Kadavath et~al.(2022)Kadavath, Conerly, Askell, Henighan, Drain, Perez, Schiefer, Dodds, Dassarma, Tran-Johnson, Johnston, El-Showk, Jones, Elhage, Hume, Chen, Bai, Bowman, Fort, Ganguli, Hernandez, Jacobson, Kernion, Kravec, Lovitt, Ndousse, Olsson, Ringer, Amodei, Brown, Clark, Joseph, Mann, McCandlish, Olah, and Kaplan]{Kadavath2022LanguageM}
Saurav Kadavath, Tom Conerly, Amanda Askell, T.~J. Henighan, Dawn Drain, Ethan Perez, Nicholas Schiefer, Zachary Dodds, Nova Dassarma, Eli Tran-Johnson, Scott Johnston, Sheer El-Showk, Andy Jones, Nelson Elhage, Tristan Hume, Anna Chen, Yuntao Bai, Sam Bowman, Stanislav Fort, Deep Ganguli, Danny Hernandez, Josh Jacobson, John Kernion, Shauna Kravec, Liane Lovitt, Kamal Ndousse, Catherine Olsson, Sam Ringer, Dario Amodei, Tom~B. Brown, Jack Clark, Nicholas Joseph, Benjamin Mann, Sam McCandlish, Chris Olah, and Jared Kaplan.
\newblock Language models (mostly) know what they know.
\newblock \emph{arXiv preprint arXiv:2207.05221}, 2022.

\bibitem[Kang et~al.(2025{\natexlab{a}})Kang, Bakman, Yaldiz, Buyukates, and Avestimehr]{kang2025uncertainty}
Sungmin Kang, Yavuz~Faruk Bakman, Duygu~Nur Yaldiz, Baturalp Buyukates, and Salman Avestimehr.
\newblock Uncertainty quantification for hallucination detection in large language models: Foundations, methodology, and future directions.
\newblock \emph{arXiv preprint arXiv:2510.12040}, 2025{\natexlab{a}}.

\bibitem[Kang et~al.(2025{\natexlab{b}})Kang, Zhao, and Song]{kang2025scalable}
Zhewei Kang, Xuandong Zhao, and Dawn Song.
\newblock Scalable best-of-n selection for large language models via self-certainty.
\newblock \emph{arXiv preprint arXiv:2502.18581}, 2025{\natexlab{b}}.

\bibitem[Karan \& Du(2025)Karan and Du]{karan2025reasoning}
Aayush Karan and Yilun Du.
\newblock Reasoning with sampling: Your base model is smarter than you think.
\newblock \emph{arXiv preprint arXiv:2510.14901}, 2025.

\bibitem[Kim \& Ghahramani(2012)Kim and Ghahramani]{kim2012bayesian}
Hyun-Chul Kim and Zoubin Ghahramani.
\newblock Bayesian classifier combination.
\newblock In \emph{Artificial Intelligence and Statistics}, pp.\  619--627. PMLR, 2012.

\bibitem[Kojima et~al.(2022)Kojima, Gu, Reid, Matsuo, and Iwasawa]{kojima2022zeroshot}
Takeshi Kojima, Shixiang~Shane Gu, Machel Reid, Yutaka Matsuo, and Yusuke Iwasawa.
\newblock Large language models are zero-shot reasoners.
\newblock \emph{Advances in Neural Information Processing Systems}, 35:\penalty0 22199--22213, 2022.

\bibitem[Ladha(1992)]{ladha1992condorcet}
K.K. Ladha.
\newblock {The Condorcet Jury Theorem, Free Speech and Correlated Votes}.
\newblock \emph{American Journal of Political Science}, 36\penalty0 (3):\penalty0 617--634, 1992.

\bibitem[Lewkowycz et~al.(2022)Lewkowycz, Andreassen, Dohan, Dyer, Michalewski, Ramasesh, Slone, Anil, Schlag, Gutman-Solo, Wu, Neyshabur, Gur-Ari, and Misra]{lewkowycz2022minerva}
Aitor Lewkowycz, Anders Andreassen, David Dohan, Ethan Dyer, Henryk Michalewski, Vinay Ramasesh, Ambrose Slone, Cem Anil, Imanol Schlag, Theo Gutman-Solo, Yuhuai Wu, Behnam Neyshabur, Guy Gur-Ari, and Vedant Misra.
\newblock Solving quantitative reasoning problems with language models.
\newblock In \emph{Advances in Neural Information Processing Systems}, 2022.

\bibitem[Li et~al.(2024{\natexlab{a}})Li, Beeching, Tunstall, Lipkin, Soletskyi, Huang, Rasul, Yu, Jiang, Shen, et~al.]{li2024numinamath}
Jia Li, Edward Beeching, Lewis Tunstall, Ben Lipkin, Roman Soletskyi, Shengyi Huang, Kashif Rasul, Longhui Yu, Albert~Q Jiang, Ziju Shen, et~al.
\newblock Numinamath: The largest public dataset in ai4maths with 860k pairs of competition math problems and solutions.
\newblock \emph{Hugging Face repository}, 13\penalty0 (9):\penalty0 9, 2024{\natexlab{a}}.

\bibitem[Li et~al.(2023)Li, Beirami, Sanjabi, and Smith]{li2023tilted}
Tian Li, Ahmad Beirami, Maziar Sanjabi, and Virginia Smith.
\newblock On tilted losses in machine learning: Theory and applications.
\newblock \emph{Journal of Machine Learning Research}, 24\penalty0 (142):\penalty0 1--79, 2023.

\bibitem[Li et~al.(2024{\natexlab{b}})Li, Yuan, Feng, Pan, Wang, Sun, Wang, and Li]{liescape}
Yiwei Li, Peiwen Yuan, Shaoxiong Feng, Boyuan Pan, Xinglin Wang, Bin Sun, Heda Wang, and Kan Li.
\newblock {Escape Sky-high Cost: Early-stopping Self-Consistency for Multi-step Reasoning}.
\newblock In \emph{The Twelfth International Conference on Learning Representations}, 2024{\natexlab{b}}.

\bibitem[Lightman et~al.(2023)Lightman, Kosaraju, Burda, Edwards, Baker, Lee, Leike, Schulman, Sutskever, and Cobbe]{lightman2023let}
Hunter Lightman, Vineet Kosaraju, Yuri Burda, Harrison Edwards, Bowen Baker, Teddy Lee, Jan Leike, John Schulman, Ilya Sutskever, and Karl Cobbe.
\newblock Let's verify step by step.
\newblock In \emph{The Twelfth International Conference on Learning Representations}, 2023.

\bibitem[List \& Goodin(2001)List and Goodin]{list2001epistemic}
Christian List and {Robert E.} Goodin.
\newblock Epistemic democracy: Generalizing the condorcet jury theorem.
\newblock \emph{Journal of Political Philosophy}, 9\penalty0 (3):\penalty0 277--306, 2001.
\newblock ISSN 0963-8016.
\newblock \doi{10.1111/1467-9760.00128}.

\bibitem[Liu et~al.(2024)Liu, Wang, Wang, and Cai]{liu2024speculative}
Jiahao Liu, Qifan Wang, Jingang Wang, and Xunliang Cai.
\newblock {Speculative Decoding via Early-exiting for Faster LLM Inference with Thompson Sampling Control Mechanism}.
\newblock \emph{arXiv preprint arXiv:2406.03853}, 2024.

\bibitem[Liu et~al.(2012)Liu, Peng, and Ihler]{liu2012variational}
Qiang Liu, Jian Peng, and Alexander~T Ihler.
\newblock Variational inference for crowdsourcing.
\newblock \emph{Advances in neural information processing systems}, 25, 2012.

\bibitem[Ma et~al.(2025)Ma, Liu, Jiang, Zhang, Ma, and Chen]{ma2025general}
Xueguang Ma, Qian Liu, Dongfu Jiang, Ge~Zhang, Zejun Ma, and Wenhu Chen.
\newblock General-reasoner: Advancing llm reasoning across all domains.
\newblock \emph{arXiv preprint arXiv:2505.14652}, 2025.

\bibitem[Madaan et~al.(2023)Madaan, Tandon, Gupta, Hallinan, Gao, Wiegreffe, Alon, Dziri, Prabhumoye, Yang, Gupta, Majumder, Hermann, Welleck, Yazdanbakhsh, and Clark]{madaan2023selfrefine}
Aman Madaan, Niket Tandon, Prakhar Gupta, Skyler Hallinan, Luyu Gao, Sarah Wiegreffe, Uri Alon, Nouha Dziri, Shrimai Prabhumoye, Yiming Yang, Shashank Gupta, Bodhisattwa~Prasad Majumder, Katherine Hermann, Sean Welleck, Amir Yazdanbakhsh, and Peter Clark.
\newblock {Self-Refine: Iterative Refinement with Self-Feedback}.
\newblock In \emph{Thirty-seventh Conference on Neural Information Processing Systems}, 2023.

\bibitem[Manvi et~al.(2024)Manvi, Singh, and Ermon]{manvi2024adaptive}
Rohin Manvi, Anikait Singh, and Stefano Ermon.
\newblock {Adaptive inference-time compute: LLMs can predict if they can do better, even mid-generation}.
\newblock \emph{arXiv preprint arXiv:2410.02725}, 2024.

\bibitem[Miller(1995)]{miller_1995}
George Miller.
\newblock Note on the bias of information estimates.
\newblock In \emph{Information Theory in Psychology: Problems and Methods}, pp.\  95--100, 1995.

\bibitem[Muennighoff et~al.()Muennighoff, Yang, Shi, Li, Fei-Fei, Hajishirzi, Zettlemoyer, Liang, Candes, and Hashimoto]{muennighoffs1}
Niklas Muennighoff, Zitong Yang, Weijia Shi, Xiang~Lisa Li, Li~Fei-Fei, Hannaneh Hajishirzi, Luke Zettlemoyer, Percy Liang, Emmanuel Candes, and Tatsunori Hashimoto.
\newblock s1: Simple test-time scaling.
\newblock In \emph{Workshop on Reasoning and Planning for Large Language Models}.

\bibitem[OpenAI(2023)]{openai2023gpt4}
OpenAI.
\newblock {GPT-4} technical report.
\newblock \emph{arXiv preprint arXiv:2303.08774}, 2023.

\bibitem[Prabhudesai et~al.(2025)Prabhudesai, Chen, Ippoliti, Fragkiadaki, Liu, and Pathak]{prabhudesai2025maximizing}
Mihir Prabhudesai, Lili Chen, Alex Ippoliti, Katerina Fragkiadaki, Hao Liu, and Deepak Pathak.
\newblock Maximizing confidence alone improves reasoning.
\newblock \emph{arXiv preprint arXiv:2505.22660}, 2025.

\bibitem[Prasad et~al.(2025)Prasad, Yuan, Pang, Xu, Fazel-Zarandi, Bansal, Sukhbaatar, Weston, and Yu]{prasadself}
Archiki Prasad, Weizhe Yuan, Richard~Yuanzhe Pang, Jing Xu, Maryam Fazel-Zarandi, Mohit Bansal, Sainbayar Sukhbaatar, Jason~E Weston, and Jane Yu.
\newblock Self-consistency preference optimization.
\newblock In \emph{Forty-second International Conference on Machine Learning}, 2025.

\bibitem[Ramdas \& Wang(2025)Ramdas and Wang]{STA-002}
Aaditya Ramdas and Ruodu Wang.
\newblock Hypothesis testing with e-values.
\newblock \emph{Foundations and Trends® in Statistics}, 1\penalty0 (1-2):\penalty0 1--390, 2025.
\newblock ISSN 2978-4212.
\newblock \doi{10.1561/3600000002}.

\bibitem[Raykar et~al.(2010)Raykar, Yu, Zhao, Valadez, Florin, Bogoni, and Moy]{raykar2010learning}
Vikas~C Raykar, Shipeng Yu, Linda~H Zhao, Gerardo~Hermosillo Valadez, Charles Florin, Luca Bogoni, and Linda Moy.
\newblock Learning from crowds.
\newblock \emph{Journal of machine learning research}, 11\penalty0 (4), 2010.

\bibitem[Shafayat et~al.(2025)Shafayat, Tajwar, Salakhutdinov, Schneider, and Zanette]{shafayat2025large}
Sheikh Shafayat, Fahim Tajwar, Ruslan Salakhutdinov, Jeff Schneider, and Andrea Zanette.
\newblock Can large reasoning models self-train?
\newblock \emph{arXiv preprint arXiv:2505.21444}, 2025.

\bibitem[Shao et~al.(2024)Shao, Wang, Zhu, Xu, Song, Bi, Zhang, Zhang, Li, Wu, and Guo]{shao2024deepseekmathpushinglimitsmathematical}
Zhihong Shao, Peiyi Wang, Qihao Zhu, Runxin Xu, Junxiao Song, Xiao Bi, Haowei Zhang, Mingchuan Zhang, Y.~K. Li, Y.~Wu, and Daya Guo.
\newblock {DeepSeekMath: Pushing the Limits of Mathematical Reasoning in Open Language Models}.
\newblock \emph{arXiv preprint arXiv:2402.03300}, 2024.

\bibitem[Sheng et~al.(2024)Sheng, Zhang, Ye, Wu, Zhang, Zhang, Peng, Lin, and Wu]{sheng2024hybridflow}
Guangming Sheng, Chi Zhang, Zilingfeng Ye, Xibin Wu, Wang Zhang, Ru~Zhang, Yanghua Peng, Haibin Lin, and Chuan Wu.
\newblock {HybridFlow: A Flexible and Efficient RLHF Framework}.
\newblock \emph{arXiv preprint arXiv: 2409.19256}, 2024.

\bibitem[Shinn et~al.(2023)Shinn, Cassano, Gopinath, Narasimhan, and Yao]{shinn2023reflexion}
Noah Shinn, Federico Cassano, Ashwin Gopinath, Karthik Narasimhan, and Shunyu Yao.
\newblock Reflexion: Language agents with verbal reinforcement learning.
\newblock \emph{Advances in Neural Information Processing Systems}, 36:\penalty0 8634--8652, 2023.

\bibitem[Song et~al.(2025)Song, Huang, Cheng, Gao, Xu, Zhao, Wang, and Wu]{song2025irt}
Wei Song, Zhenya Huang, Cheng Cheng, Weibo Gao, Bihan Xu, GuanHao Zhao, Fei Wang, and Runze Wu.
\newblock {IRT-Router: Effective and Interpretable Multi-LLM Routing via Item Response Theory}.
\newblock \emph{arXiv preprint arXiv:2506.01048}, 2025.

\bibitem[Song et~al.(2024)Song, Wang, Li, and Lin]{song2024good}
Yifan Song, Guoyin Wang, Sujian Li, and Bill~Yuchen Lin.
\newblock The good, the bad, and the greedy: Evaluation of llms should not ignore non-determinism.
\newblock \emph{arXiv preprint arXiv:2407.10457}, 2024.

\bibitem[Tang et~al.(2025)Tang, Zheng, Synnaeve, and Munos]{tang2025optimizing}
Yunhao Tang, Kunhao Zheng, Gabriel Synnaeve, and Remi Munos.
\newblock {Optimizing Language Models for Inference Time Objectives using Reinforcement Learning}.
\newblock In \emph{Forty-second International Conference on Machine Learning}, 2025.

\bibitem[Tao et~al.(2025)Tao, Kulikov, Saha, Wang, Xu, Li, Weston, and Yu]{tao2025hybrid}
Leitian Tao, Ilia Kulikov, Swarnadeep Saha, Tianlu Wang, Jing Xu, Yixuan Li, Jason~E Weston, and Ping Yu.
\newblock Hybrid reinforcement: When reward is sparse, it's better to be dense.
\newblock \emph{arXiv preprint arXiv:2510.07242}, 2025.

\bibitem[Valiant \& Valiant(2013)Valiant and Valiant]{NIPS2013_53c04118}
Paul Valiant and Gregory Valiant.
\newblock {Estimating the Unseen: Improved Estimators for Entropy and other Properties}.
\newblock In C.J. Burges, L.~Bottou, M.~Welling, Z.~Ghahramani, and K.Q. Weinberger (eds.), \emph{Advances in Neural Information Processing Systems}, volume~26. Curran Associates, Inc., 2013.

\bibitem[Ville(1939)]{ville1939collectif}
Jean Ville.
\newblock \emph{\'Etude critique de la notion de collectif}.
\newblock Gauthier-Villars, 1939.

\bibitem[Wan et~al.(2025)Wan, Wu, Chen, and Li]{wan2025reasoning}
Guangya Wan, Yuqi Wu, Jie Chen, and Sheng Li.
\newblock {Reasoning Aware Self-Consistency: Leveraging Reasoning Paths for Efficient LLM Sampling}.
\newblock In \emph{Proceedings of the 2025 Conference of the Nations of the Americas Chapter of the Association for Computational Linguistics: Human Language Technologies (Volume 1: Long Papers)}, pp.\  3613--3635, 2025.

\bibitem[Wang et~al.(2020)Wang, Hsieh, Chang, Chen, Pan, Wei, and Juan]{wang2020contextual}
Pei-Hsin Wang, Sheng-Iou Hsieh, Shih-Chieh Chang, Yu-Ting Chen, Jia-Yu Pan, Wei Wei, and Da-Chang Juan.
\newblock Contextual temperature for language modeling.
\newblock \emph{arXiv preprint arXiv:2012.13575}, 2020.

\bibitem[Wang et~al.(2025{\natexlab{a}})Wang, Feng, Li, Yuan, Zhang, Tan, Pan, Hu, and Li]{wang2025make}
Xinglin Wang, Shaoxiong Feng, Yiwei Li, Peiwen Yuan, Yueqi Zhang, Chuyi Tan, Boyuan Pan, Yao Hu, and Kan Li.
\newblock Make every penny count: Difficulty-adaptive self-consistency for cost-efficient reasoning.
\newblock In \emph{Findings of the Association for Computational Linguistics: NAACL 2025}, pp.\  6904--6917, 2025{\natexlab{a}}.

\bibitem[Wang et~al.(2022)Wang, Wei, Schuurmans, Bosma, Chi, Le, and Zhou]{wang2022selfconsistency}
Xuezhi Wang, Jason Wei, Dale Schuurmans, Maarten Bosma, Ed~H. Chi, Quoc~V. Le, and Denny Zhou.
\newblock Self-consistency improves chain of thought reasoning in language models.
\newblock \emph{arXiv preprint arXiv:2203.11171}, 2022.

\bibitem[Wang et~al.(2025{\natexlab{b}})Wang, Zhang, Huang, Yang, Zhang, Huang, and Wang]{wang2025sampling}
Yiming Wang, Pei Zhang, Siyuan Huang, Baosong Yang, Zhuosheng Zhang, Fei Huang, and Rui Wang.
\newblock Sampling-efficient test-time scaling: Self-estimating the best-of-n sampling in early decoding.
\newblock \emph{arXiv preprint arXiv:2503.01422}, 2025{\natexlab{b}}.

\bibitem[Wei et~al.(2022)Wei, Wang, Schuurmans, Bosma, Ichter, Xia, Chi, Le, and Zhou]{wei2022cot}
Jason Wei, Xuezhi Wang, Dale Schuurmans, Maarten Bosma, Brian Ichter, Fei Xia, Ed~H. Chi, Quoc~V. Le, and Denny Zhou.
\newblock Chain-of-thought prompting elicits reasoning in large language models.
\newblock \emph{Advances in Neural Information Processing Systems}, 35, 2022.

\bibitem[Welinder et~al.(2010)Welinder, Branson, Belongie, and Perona]{welinder2010multidimensional}
Peter Welinder, Steve Branson, Serge Belongie, and Pietro Perona.
\newblock The multidimensional wisdom of crowds.
\newblock \emph{Advances in neural information processing systems}, 23, 2010.

\bibitem[Wen et~al.(2025)Wen, Ankner, Somani, Hase, Marks, Goldman-Wetzler, Petrini, Sleight, Burns, He, et~al.]{wen2025unsupervised}
Jiaxin Wen, Zachary Ankner, Arushi Somani, Peter Hase, Samuel Marks, Jacob Goldman-Wetzler, Linda Petrini, Henry Sleight, Collin Burns, He~He, et~al.
\newblock Unsupervised elicitation of language models.
\newblock \emph{arXiv preprint arXiv:2506.10139}, 2025.

\bibitem[Whitehill et~al.(2009)Whitehill, Wu, Bergsma, Movellan, and Ruvolo]{whitehill2009whose}
Jacob Whitehill, Ting-fan Wu, Jacob Bergsma, Javier Movellan, and Paul Ruvolo.
\newblock Whose vote should count more: Optimal integration of labels from labelers of unknown expertise.
\newblock \emph{Advances in neural information processing systems}, 22, 2009.

\bibitem[Williams(1992)]{reinforce_92}
Ronald~J. Williams.
\newblock Simple statistical gradient-following algorithms for connectionist reinforcement learning.
\newblock \emph{Mach. Learn.}, 8\penalty0 (3–4):\penalty0 229–256, 1992.

\bibitem[Xie et~al.(2023)Xie, Kawaguchi, Zhao, Zhao, Kan, He, and Xie]{xie2023selfevaluation}
Yuxi Xie, Kenji Kawaguchi, Yiran Zhao, Xu~Zhao, Min-Yen Kan, Junxian He, and Qizhe Xie.
\newblock Self-evaluation guided beam search for reasoning.
\newblock In \emph{Thirty-seventh Conference on Neural Information Processing Systems}, 2023.

\bibitem[Xie et~al.(2024)Xie, Goyal, Zheng, Kan, Lillicrap, Kawaguchi, and Shieh]{xie2024monte}
Yuxi Xie, Anirudh Goyal, Wenyue Zheng, Min-Yen Kan, Timothy~P Lillicrap, Kenji Kawaguchi, and Michael Shieh.
\newblock Monte carlo tree search boosts reasoning via iterative preference learning.
\newblock \emph{arXiv preprint arXiv:2405.00451}, 2024.

\bibitem[Yang et~al.(2024{\natexlab{a}})Yang, Zhang, Hui, Gao, Yu, Li, Liu, Tu, Zhou, Lin, Lu, Xue, Lin, Liu, Ren, and Zhang]{yang2024qwen25mathtechnicalreportmathematical}
An~Yang, Beichen Zhang, Binyuan Hui, Bofei Gao, Bowen Yu, Chengpeng Li, Dayiheng Liu, Jianhong Tu, Jingren Zhou, Junyang Lin, Keming Lu, Mingfeng Xue, Runji Lin, Tianyu Liu, Xingzhang Ren, and Zhenru Zhang.
\newblock {Qwen2.5-Math Technical Report: Toward Mathematical Expert Model via Self-Improvement}.
\newblock \emph{arXiv preprint arXiv:2409.12122}, 2024{\natexlab{a}}.

\bibitem[Yang et~al.(2025)Yang, Yang, Zhang, Hui, Zheng, Yu, Li, Liu, Huang, Wei, Lin, Yang, Tu, Zhang, Yang, Yang, Zhou, Lin, Dang, Lu, Bao, Yang, Yu, Li, Xue, Zhang, Zhu, Men, Lin, Li, Tang, Xia, Ren, Ren, Fan, Su, Zhang, Wan, Liu, Cui, Zhang, and Qiu]{qwen2025qwen25technicalreport}
An~Yang, Baosong Yang, Beichen Zhang, Binyuan Hui, Bo~Zheng, Bowen Yu, Chengyuan Li, Dayiheng Liu, Fei Huang, Haoran Wei, Huan Lin, Jian Yang, Jianhong Tu, Jianwei Zhang, Jianxin Yang, Jiaxi Yang, Jingren Zhou, Junyang Lin, Kai Dang, Keming Lu, Keqin Bao, Kexin Yang, Le~Yu, Mei Li, Mingfeng Xue, Pei Zhang, Qin Zhu, Rui Men, Runji Lin, Tianhao Li, Tianyi Tang, Tingyu Xia, Xingzhang Ren, Xuancheng Ren, Yang Fan, Yang Su, Yichang Zhang, Yu~Wan, Yuqiong Liu, Zeyu Cui, Zhenru Zhang, and Zihan Qiu.
\newblock {Qwen2.5 Technical Report}.
\newblock \emph{arXiv preprint arXiv:2412.15115}, 2025.

\bibitem[Yang et~al.(2024{\natexlab{b}})Yang, Salamatian, Sun, Suresh, and Beirami]{yang2024asymptotics}
Joy~Qiping Yang, Salman Salamatian, Ziteng Sun, Ananda~Theertha Suresh, and Ahmad Beirami.
\newblock Asymptotics of language model alignment.
\newblock In \emph{2024 IEEE International Symposium on Information Theory (ISIT)}, pp.\  2027--2032. IEEE, 2024{\natexlab{b}}.

\bibitem[Yao et~al.(2024)Yao, Mathew, Singh, Firmani, and Barbosa]{yao2024bayesian}
Peiran Yao, Jerin~George Mathew, Shehraj Singh, Donatella Firmani, and Denilson Barbosa.
\newblock {A Bayesian Approach Towards Crowdsourcing the Truths from LLMs}.
\newblock In \emph{NeurIPS 2024 Workshop on Bayesian Decision-making and Uncertainty}, 2024.

\bibitem[Yao et~al.(2023)Yao, Yu, Zhao, Shafran, Griffiths, Cao, and Narasimhan]{yao2023treeofthoughts}
Shunyu Yao, Dian Yu, Jeffrey Zhao, Izhak Shafran, Thomas~L. Griffiths, Yuan Cao, and Karthik Narasimhan.
\newblock Tree of thoughts: Deliberate problem solving with large language models.
\newblock In \emph{Advances in Neural Information Processing Systems}, 2023.

\bibitem[Zuo et~al.(2025)Zuo, Zhang, Qu, Sheng, Zhu, Qi, Sun, Cui, Ding, and Zhou]{zuo2025ttrl}
Yuxin Zuo, Kaiyan Zhang, Shang Qu, Li~Sheng, Xuekai Zhu, Biqing Qi, Youbang Sun, Ganqu Cui, Ning Ding, and Bowen Zhou.
\newblock {TTRL: Test-Time Reinforcement Learning}.
\newblock \emph{arXiv preprint arXiv:2504.16084}, 2025.

\end{thebibliography}
\bibliographystyle{references}

\newpage
\appendix
\section{Proofs of Section \ref{sec:theoretical_bounds}}\label{app:proofs_for_theoretical_bounds}
A comprehensive reference that provides an in-depth discussion of concentration inequalities for functions of independent random variables is \cite{boucheronconcentration2013}.

\subsection{Small sample regime}\label{app:small_regime_bounds}
Hoeffding, Bernstein, and Chernoff–Markov bounds become less effective in the small-sample regime, i.e., when the number of voters satisfies $n\lesssim 50$. In this setting, the exact error probability can be obtained by leveraging the properties of the  multinomial distribution.
We provide below an efficient dynamic programming (DP) approach to compute this probability.

\subsubsection{Dynamic programming for exact multinomial probabilities}\label{app:subsubsec_dynamic_programming}
For each category $j$, define 
$$
P_j(x) =\frac{p_j^{x}}{x!} , \quad x = 0, \dots, n,
$$
and store the values $P_j= (P_j(0),\dots, P_j(n))$ in an array of length $n+1$.
The entries can be generated iteratively using the recurrence 
$$P_j(x+1) = \frac{p_j}{(x+1)}P_j(x).$$

After processing a subset of the rival categories, we define a state of the dynamic program as
$$
  \text{state }(t,m,s)\quad
  \text{with}\quad
  \begin{cases}
     t &=\text{votes for the true category }c^\star,\\
     m &=\text{current {\it maximum} vote count among the rivals processed so far,}\\
    s &=\text{{\it total} vote count allocated to processed categories.}
  \end{cases}
$$
Formally, the DP table is
$$
  \text{DP}_{i}(t,m,s)
  \;=\;\frac{1}{s!}\,
  \mathbb{P}\Bigl[N_{c^\star}^{(s)}=t,\;
           \max_{j\in\{1,\dots,i\}\setminus\{c^\star\}}N_j^{(s)} = m, \; \sum_{j\in\{c^\star, 1, \dots, i\}}N_i^{(s)} =s\Bigr],
$$
where $i$ denotes the number of categories processed.

\paragraph{Initial table.}
Before incorporating any rivals, we only consider the true category $c^\star$
$$
  \text{DP}_{1}(t,0,t)=P_{c^\star}(t),\qquad
  t=0,\dots,n.
$$
since the maximum vote count among zero rivals is naturally $0$.

\paragraph{Transition when adding a new rival $j$.}
Suppose we have already computed $\text{DP}_{i}(\cdot,\cdot, \cdot)$. 
We now incorporate category $j$.
For each triple $(t, m, s)$, we consider vote counts  $x=0,\dots,n-s$ drawn from $P_j(x)$ and define
$$
  \text{newMax}=\max\{m,x\}.
$$
The DP table is updated according to
$$
  \text{DP}_{i+1}(t,\text{newMax}, s+x)
  \;+\!=\;
  \text{DP}_{i}(t,m,s)\;P_j(x).
$$

This implementation stores states $(t,m,s)$ with $t+m\leq s\leq n$, and transitions $x=0, \dots, n-s$. 
Because of these constraints, the total number of reachable states is $\mathcal{O}(n^{3})$, rather than the naive $\mathcal{O}(n^{4})$. Iterating over all $k$ categories therefore yields a worst-case time complexity of
$$\mathcal{O}(kn^{3}).$$
The memory complexity is $\mathcal{O}(n^{3})$.

After processing all $k-1$ rivals, the DP table $\text{DP}_{k}(\cdot,\cdot,\cdot)$ is complete.
The error probability is then obtained by summing over all states where the true category does not have a strict majority and the total number of votes is equal to $n$, 
$$
  \mathbb{P}(\hat{c}_n \neq c^\star)
  \;=\;
  \sum_{t=0}^{n}\,
  \sum_{m=t}^{n}
  n!\;\text{DP}_{k}(t,m,n).
$$
For large values of $n$, factorial terms may cause numerical underflow or overflow. To prevent this, we compute the entries of the DP table in log space.

\subsection{Hoeffding bound}\label{app:hoeffding_bound}
\begin{proof}
For each rival $j\neq c^\star$, consider the random variable
\begin{equation}\label{eq:random_variable_sum_auxiliary}
    Z_j^{(n)} = N_{c^\star}^{(n)} -  N_j^{(n)}= \sum_{i=1}^n\Bigl(\mathbf 1\{X_i=c^\star\}-\mathbf 1\{X_i=j\}\Bigr).
\end{equation}
The summands $Y_i^j = \mathbf 1\{X_i=c^\star\}-\mathbf 1\{X_i=j\}$ are independent identically distributed random variables, bounded in $[-1,1]$, with expected value $\mu_j = p_{c^\star}-p_j>0$.
Applying Hoeffding's inequality we obtain
\small
    $$
    \mathbb{P}\bigl[Z_j^{(n)} \leq 0\bigr] \leq \mathbb{P}\left[\frac{Z_j^{(n)}}{n}- \mu_j\leq -\mu_j\right] = \mathbb{P}\left[-Z_j^{(n)}+ \mathbb{E}[Z_j^{(n)}]\geq \mathbb{E}[Z_j^{(n)}]\right] \leq \exp\left(-\frac{n}{2}(p_{c^\star}-p_j)^2\right).
    $$
\normalsize
The event $\{\hat{c}_n\neq c^\star\}$ implies $Z_j^{(n)}\le 0$ for some $j\neq c^\star$. Thus,
$$
    \mathbb{P}\bigl[\hat{c}_n\neq c\bigr]
    \le \sum_{j\neq c}\mathbb{P}\bigl[Z_j^{(n)}\le 0\bigr]
    \le \sum_{j\neq c}\exp\left(-\frac{n}{2}(p_{c^\star}-p_j)^2\right),
$$
which establishes the exponential bound. This can be further simplified as 
$$
    \mathbb{P}\bigl[\hat{c}_n\neq c\bigr]
    \le (k-1)\exp\left(-\frac{n}{2}\min_{j\neq c}(p_{c^\star}-p_j)^2\right).
$$
Since the upper bound decays to~$0$ as $n\to \infty$, and $k$ is finite, we obtain
$$
    \mathbb{P}[\hat{c}_n=c^\star]\to 1\quad \text{as\; $n\to\infty$}.
$$
\end{proof}

\subsubsection{Weighted majority vote for experts with different accuracies}\label{app:weighted_majority_vote}
We now consider a setting where we have access to multiple models with varying expertise: some are cheaper but less accurate, while others are more expensive but more precise. To capture this heterogeneity, we assign each expert a weight, denoted by $\omega_\ell$, that reflects its reliability. Specifically, we assume there are $L$ experts, and expert $\ell$ contributes $n_\ell$ samples.

Instead of using a simple majority vote, we aggregate predictions via a weighted majority vote
$$
\hat{c}_n^\omega = \arg\max_j \sum_{\ell =1}^L \omega_\ell { N_j^{(n_\ell)}},
$$
where $N_j^{(n_\ell)}= \sum_{i=1}^{n^\ell} \mathbf 1\!\bigl\{X^{(\ell)}_i=j\bigr\}$ is the number of samples from expert $\ell$ predicting label $j$. The total sample size is $n=\sum_\ell n_\ell$.

In this setting, the data across experts are non-exchangeable, since each expert has a different own distribution over labels.

\paragraph{Error bound for weighted majority voting.}
We derive an error bound using Hoeffding’s inequality.
For every rival $j\neq c^\star$, define the weighted margin
$$
Z_{j,\omega}^{(n)}
= N_{c,\omega}^{(n)} - N_{j,\omega}^{(n)} 
    = \sum_{\ell=1}^L\sum_{i=1}^{n_\ell}
      \omega_\ell \left(\mathbf 1\{X_i^{(\ell)}=c^\star\}-\mathbf 1\{X_i^{(\ell)}=j\}\right).
$$
Each summand $Y_{i,\ell}^{j}=\omega_\ell \left(\mathbf 1\{X_i^{(\ell)}=c^\star\}-\mathbf 1\{X_i^{(\ell)}=j\}\right)$ is independent and bounded between $-\omega_\ell$ and $\omega_\ell$.
The sum $Z_{j,\omega}^{(n)}$ has mean
$$
\mathbb{E}\left[Z_{j,\omega}^{(n)}\right] = \sum_{\ell=1}^Ln_\ell \,\omega_\ell\left(p_{c^\star}^{\ell}-p_j^\ell \right).
$$
Applying Hoeffding’s inequality, we obtain
\begin{align*}
    \mathbb{P}\left[\hat{c}_n^\omega\neq c^\star\right]
    &\le \sum_{j\neq c}\mathbb{P}\left[Z_{j,\omega}^{(n)}\le 0\right]\le \sum_{j\neq c^\star}\exp\left(-\frac{1}{2}\frac{\left(\sum_{\ell=1}^L{n_\ell\,\omega_\ell\left(p_{c^\star}^\ell-p_j^\ell\right)}\right)^2}{\sum_{\ell = 1}^Ln_\ell \,\omega_\ell^2}\right)\\
    &\le(k-1) \exp\left(-\frac{1}{2}\frac{\left(\sum_{\ell=1}^L{n_\ell\,\omega_\ell\min_{j\neq c^\star}\left(p_{c^\star}^\ell-p_j^\ell\right)}\right)^2}{\sum_{\ell = 1}^Ln_\ell \,\omega_\ell^2}\right)\\
    &\le (k-1)\exp\left(-\frac{1}{2}\sum_{\ell=1}^Ln_\ell\frac{n_\ell\,\omega_\ell^2}{\sum_{\ell =1}^{L} n_\ell\,\omega_\ell^2}\min_{j\neq c^\star}\left(p_{c^\star}^\ell-p_j^\ell\right)^2\right).    
\end{align*}
If each expert contributes the same number of sample $(n_\ell = n)$, the previous bound can be simplified as
\begin{align*}
    \mathbb{P}\left[\hat{c}_n^\omega\neq c^\star\right]
    &\le \sum_{j\neq c^\star}\mathbb{P}\left[Z_{j,\omega}^{(n)}\le 0\right]\le \sum_{j\neq c^\star}\exp\left(-\frac{n}{2}\frac{\left(\sum_{\ell=1}^L\omega_\ell\left(p_{c^\star}^\ell-p_j^\ell\right)\right)^2}{\sum_{\ell=1}^L \omega_\ell^2}\right)\\
    &\le (k-1)\exp\left(-\frac{n}{2}\sum_{\ell=1}^L\left(\frac{\omega_\ell^2}{\sum_{\ell=1}^L\omega_\ell^2}\right)\min_{j\neq c^\star}\left(p_{c^\star}^\ell-p_j^\ell\right)^2\right).    
\end{align*}

\paragraph{Optimal weights based on expert accuracy.}
Recall that our decision rule maps the set of expert responses $X$ to a final answer. We say a decision rule is \emph{optimal} if it minimises the probability of error. Formally, letting $D$ denote the rule, we want to minimise
$$
\mathbb{P}\left[D(X)\neq c^\star\right], \quad X = (X_{i_\ell}^{(\ell)}),\,\, \ell = 1,\dots, L,\,\,\,\, i_\ell = 1,\dots,n_\ell,
$$
where $c^\star$ is the true answer.
To derive the optimal decision rule, we make the following assumptions.
\begin{assumption}\label{assumption:independence_between_draws}
    Independence: conditioned on the ground-truth label $c^\star$, the random variables $X_i^{(\ell)}$, corresponding to the $i$-th response from expert $\ell$, are mutually independent across both experts and repetitions.
\end{assumption}
\begin{assumption}\label{assumption:unbiased_truth}
    Unbiased truth: the ground-truth label is uniformly distributed, i.e. $\mathbb{P}[c^\star=j] = 1/k$ for $j=1, \dots, k$.
\end{assumption}

Suppose that we know the confusion matrix $\left(C_{ij}^{(\ell)}\right)_{ij}$ for each expert $\ell$, where $$C_{ij}^{(\ell)} = \mathbb{P}\left[X^{(\ell)} = i\big \vert c^\star=j\right]$$ denotes the probability that model $\ell$ will record value $i$ given $j$ is the true response. 
Then, the decision rule that minimises the Bayes risk coincides with the Maximum a Posteriori (MAP) rule,
$$
D^{\text{OPT}}(X)=\arg\max_j\; \log \mathbb{P}[c^\star=j\vert X].
$$
By Bayes' theorem we have
\begin{align*}
  \arg\max_j \,\,\mathbb{P}[c^\star=j\vert X] &= \arg\max_j \,\,\mathbb{P}[c^\star=j] \mathbb{P}[X\vert c^\star=j]\\
  &= \arg\max_j \,\, \prod_{\ell=1}^L \prod_{i=1}^{n_\ell}\mathbb{P}\left[X_i^{(\ell)}\big \vert c^\star=j\right] = \prod_{\ell=1}^L \prod_{i=1}^{n_\ell} C_{j\ X_i^{(\ell)}}^{(\ell)},  
\end{align*}
which results into
$$
D^{\text{OPT}}(X)=\arg\max_j\; \sum_{\ell=1}^L \sum_{i=1}^{n_\ell}\log C_{j\ X_i^{(\ell)}}^{(\ell)}.
$$
Now, imagine that we only know each expert’s overall competence level $q_\ell \in (0,1)$, defined as the probability of correctly predicting the true label,
$$q_\ell = \mathbb{P}\left[X^{(\ell)}= j\big\vert c^\star=j\right],$$
but not the full confusion matrix.
A natural approximation is to assume that errors are distributed uniformly across the $k-1$ incorrect labels, i.e.
$$\mathbb{P}[X^{(\ell)}= i\big\vert c^\star= j\neq i] = \frac{1-q_\ell}{k-1}.$$
Without this approximation, one would need to estimate the full confusion matrices. This can be done, for example, via the Expectation–Maximisation algorithm \citep{dawid_skene_79}.

\subsection{Bernstein bound}\label{app:bernstein_bound}
\begin{proof}
Let the random variable $Y_i^{j} = \mathbf 1\{X_i=c^\star\}-\mathbf 1\{X_i=j\}$. We have that 
$$\mathbb{E}\left[Y_i^j- (p_{c^\star}-p_j)\right] = 0,$$
    $$
    \left \lvert Y_i^j - (p_{c^\star}-p_j)\right\rvert = \left \lvert\mathbf 1\{X_i=c\}-\mathbf 1\{X_i=j\}- (p_{c^\star}-p_j)  \right\rvert\leq 1+(p_{c^\star}-p_j),
    $$ 
    and
    $$
    \sigma_j^2 = \mathbb{E}\left[\left(Y_i^j- (p_{c^\star}-p_j)\right)^2\right] = p_{c^\star} + p_j - (p_{c^\star}-p_j)^2.
    $$
    Consider $Z_j^{(n)}=\sum_i^n Y_i^j$. 
    Applying Bernstein's inequality gives
    $$
      \mathbb{P}\bigl[Z_j^{(n)} \leq 0\bigr] \leq \mathbb{P}\left[-\frac{Z_j^{(n)}}{n}+ \mu_j\geq \mu_j\right]  \leq \exp\left(-\frac{n(p_{c^\star}-p_j)^2}{2\sigma_j^2 + \frac{2}{3}(p_{c^\star}-p_j) + \frac{2}{3}(p_{c^\star}-p_j)^2}\right).
    $$
    Since the event $\{\hat{c}_n\neq c^\star\}$ implies $Z_j^{(n)}\le 0$ for some $j\neq c^\star$, we obtain the bound
$$
    \mathbb{P}\bigl[\hat{c}_n\neq c^\star\bigr]
    \le \sum_{j\neq c^\star}\mathbb{P}\bigl[Z_j^{(n)}\le 0\bigr]
    \le \sum_{j\neq c^\star}\exp\left(-\frac{n(p_{c^\star}-p_j)^2}{2\sigma_j^2 + \frac{2}{3}(p_{c^\star}-p_j) + \frac{2}{3}(p_{c^\star}-p_j)^2}\right).
$$ 
Noting that $p_{c^\star} + p_j \leq 1$, we can obtain a simpler but looser bound 
$$
\mathbb{P}[\hat{c}_n \neq c^\star] \leq \sum_{j \neq c^\star}\exp\left(-\frac{n (p_{c^\star} - p_j)^2}{2\left(1 - \frac{2}{3}(p_{c^\star} - p_j)^2\right) + \frac{2}{3}(p_{c^\star} - p_j)}\right),
$$
which only depends on the probability gaps $\delta_j = p_{c^\star} - p_j$.
\end{proof}

\subsection{Chernoff-Markov bound}\label{app:chernoff-markov_bound}
\begin{proof}
    Using the Chernoff-Markov inequality, for any $\lambda<0$ we have
    \begin{align*}
    \mathbb{P}\left[Z_j^{(n)}\leq 0\right] &= \mathbb{P}\left[e^{\lambda Z_j^{(n)}}\geq 1\right] \leq \mathbb{E}\left[e^{\lambda Z_j^{(n)}}\right] 
    = \left(\mathbb{E}\left[e^{\lambda Y_1^j}\right]\right)^n,
    \end{align*}
where we have used that the random variables $Y_i^j$ are independent and identically distributed. The moment generating function of $Y_1^j$ is 
$$
M(\lambda) = \mathbb{E}\left[e^{\lambda Y_1^j}\right] = p_{c^\star} e^{\lambda} + p_j e^{-\lambda} + 1- (p_{c^\star}+p_j).
$$
We now optimise $M(\lambda)$ over $\lambda<0$.
Since $p_{c^\star} > p_j$, there is no minimiser in $\lambda > 0$, so we can just extend the optimisation to $\lambda \in \mathbb{R}$.
    Setting the derivative to zero,
    $$M'(\lambda) = p_{c^\star} e^{\lambda} - p_j e^{-\lambda} = 0$$
    gives the minimiser 
    \begin{equation*}
    \lambda^\star = \frac{1}{2}\log(p_j/p_{c^\star}) < 0
    \end{equation*}
    with corresponding value
$$
M(\lambda^*) = 1-(p_{c^\star}+p_j) + 2\sqrt{p_{c^\star}p_j}.
$$
Thus, the Chernoff-Markov bound becomes
\begin{align*}  
\mathbb{P}\left[Z_j^{(n)}\leq 0\right]&\leq \left(1-(p_{c^\star}+p_j) + 2\sqrt{p_{c^\star}p_j}\right)^n = \exp\left(n\log\left(1-(p_{c^\star}+p_j) + 2\sqrt{p_{c^\star}p_j}\right)\right)\\
& = \exp\left(n\log\left(1-(\sqrt{p_{c^\star}}-\sqrt{p_j})^2\right)\right).
\end{align*}
Consequently, we have
$$
\mathbb{P}\bigl[\hat{c}_n\neq c^\star\bigr]
\le \sum_{j\neq c^\star}\mathbb{P}\bigl[Z_j^{(n)}\le 0\bigr]
\le \sum_{j\neq c^\star} \exp\left(n\log\left(1-(\sqrt{p_{c^\star}}-\sqrt{p_j})^2\right)\right).
$$
\end{proof}

\subsection{CLT bound}\label{app:majority-clt}
\begin{proof}
For each rival $j$, consider the random variable $Z_j^{(n)}$ defined in (\ref{eq:random_variable_sum_auxiliary}), which is a sum of independent and identically distributed random variables with mean 
$$\mu_j = p_{c^\star}-p_j$$
and variance
$$
\sigma_j^2 = p_{c^\star}+p_j - (p_{c^\star}-p_j)^2.
$$
By the central limit theorem,
$$
\frac{Z_j^{(n)}-n(p_{c^\star}-p_j)}{\sqrt{n}}\xrightarrow{d} \mathcal{N}(0,\sigma_j^2).
$$
Therefore, as $n\to \infty$, we have
\small
$$
\mathbb P\left[Z_j^{(n)}\le 0\right]
= \mathbb P\left[\frac{Z_j^{(n)}-n(p_{c^\star}-p_j)}{\sigma_j\sqrt n}
                 \le -\,\frac{(p_{c^\star}-p_j)\sqrt n}{\sigma_j}\right]
= \Phi\Bigl(-\,\frac{(p_{c^\star}-p_j)\sqrt n}{\sigma_j}\Bigr)
  \,[1+O(n^{-1/2})],
$$
\normalsize
where $\Phi$ denotes the CDF of a standard Gaussian random variable.

Majority voting fails if $Z_j^{(n)}\le 0$ for some $j \neq c^\star$.
Applying the union bound, we obtain
$$
\mathbb P[\hat{c}_n\neq c^\star]
   \;\le\;\sum_{j \neq c^\star}\mathbb P\left[Z_j^{(n)}\le 0\right]
   \;=\;\sum_{j \neq c^\star}
        \Phi\Bigl(-\,\frac{(p_{c^\star} - p_j)\sqrt n}{\sigma_j}\Bigr)
        [1+O(n^{-1/2})].
$$
To bound the Gaussian tail, we use Craig's formula
$$
\Phi(-x) = \frac{1}{\pi}\int_0^{\pi/2}\exp\left(-\frac{x^2}{2\sin^2\theta}\right)d\theta\le \frac{1}{2}e^{-\frac{x^{2}}{2}}, \quad\,\,\text{for $x>0$}.
$$
Substituting this bound gives
$$
\mathbb P[\hat{c}_n\neq c^\star]
   \;\le\;\frac{1}{2}\sum_{j \neq c^\star} \exp\left(-\frac{n}{2}\left(\frac{p_{c^\star}-p_j}{\sigma_j}\right)^2\right)\leq \frac{1}{2}(k-1) \exp\left(-\frac{n}{2}\min_{j \neq c^\star}\left(\frac{p_{c^\star}-p_j}{\sigma_j}\right)^2\right).
$$
For fixed $p_{c^\star}$, the ratio $\frac{p_{c^\star}-p_j}{\sigma_j}$ is monotonically decreasing in $p_j$. 
Therefore, the smallest value, and hence the slowest exponential decay, is attained at the rival with the largest probability among the competitors. Denoting this \emph{second-largest} vote  probability by
$$p_{j^\star} = \max_{j \neq c^\star} p_j,$$
the convergence rate in the exponential bound above is determined by 
$$
\kappa = \frac{\delta}{2 p_{c^\star} - \delta-\delta^2},\quad\,\,\ \delta = p_{c^\star}-p_{j^\star}.
$$
Thus, the competitor that most threatens the accuracy of majority voting is precisely the category with the second–highest support.

The previous bound derived from the central limit theorem can be sharpened by incorporating two classical corrections.
The first correction is the continuity term, that is, the correction term due to discreteness. Since the random variable $Z_j^{(n)}$ takes values in the discrete set $\{-n, \dots, n\}$, the event $Z_j^{(n)}\leq 0$ is equivalent to $Z_j^{(n)}\leq 1/2$. Hence,
$$
\mathbb{P}\left[Z_j^{(n)}\leq 0\right] = \mathbb{P}\left[Z_j^{(n)}\le 1/2\right].
$$
Applying the central limit theorem approximation then yields, as $n\to\infty$,
\begin{align*}
\mathbb P\left[Z_j^{(n)}\le 1/2\right]
&= \mathbb P\left[\frac{Z_j^{(n)}-n(p_{c^\star}-p_j)}{\sigma_j\sqrt n}
                 \le\frac{1}{2\sigma_j\sqrt{n}} -\frac{\sqrt n(p_{c^\star}-p_j)}{\sigma_j}\right]\\
&\approx \Phi\left(\frac{\sqrt n(p_{c^\star}-p_j)-1/(2\sqrt{n})}{\sigma_j}\right) = \frac{1}{2}\text{erfc} \left(\frac{\sqrt n(p_{c^\star}-p_j)-1/(2\sqrt{n})}{\sqrt{2}\,\sigma_j}\right).
\end{align*}
A further refinement comes from the Berry–Esseen theorem, which quantifies the uniform error of the central limit theorem approximation. In particular, for all $n$
$$
\left\vert \mathbb{P}\left[\frac{Z_j^{(n)}-n(p_{c^\star}-p_j)}{\sigma_j\sqrt n}\le x\right] -\Phi(x)\right\vert \leq \frac{C \rho_j}{\sigma_j^3\sqrt{n}},
$$
where $\rho_j$ denotes the third central moment,
$$
\rho_j = \mathbb{E}\left[\left(Y_i^j-(p_{c^\star}-p_j)\right)^3\right] = (p_{c^\star} - p_{j})(1-3(p_{c^\star} + p_{j}) + 2(p_{c^\star} - p_{j})^2)
$$
and $C\le 0.56$ is a universal constant. Incorporating both corrections, we obtain the refined bound
$$
\mathbb P[\hat{c}_n\neq c^\star]\leq \sum_{j \neq c^\star}\frac{1}{2}\text{erfc}\left(\frac{{\sqrt{n}(p_{c^\star}-p_j)} - {1}/{(2\sqrt{n}})}{\sqrt{2}\,\sigma_j}\right) + \frac{0.56\rho_j}{\sigma_j^3\sqrt{n}}.
$$
\end{proof}

\subsection{Large-deviations regime}\label{app:sanov_bound}
\begin{proof}[Proof]
Let $\mathbf{p} = (p_1, \dots, p_k)$ denote the true probability distribution, and let $\mathbf{\hat{p}_n} = (\hat{{p}}_{n,1}, \dots, \hat{{p}}_{n,k})$ be the empirical measure, where $\hat{p}_{n,j}$ are the empirical frequencies for each category
$$
\hat p_{n,j}=\frac1n\sum_{i=1}^n\mathbf 1\{X_i=j\}.
$$
Recall that $p_{c^\star} = \max_j p_j$.
Define the set $\mathcal{B}\subseteq\Delta_k$ by
\begin{equation}\label{eq:bad_set}
\mathcal B
=\left\{\mathbf q\in\Delta_k:\;q_{c^\star}\le\max_{j \neq c^\star}q_j\right\}= \left\{ \mathbf q \in \Delta_k : q_{c^\star} \leq q_j, \mbox{ for some } j \neq c^\star\right\}.    
\end{equation}

    \paragraph{Step 1. Sanov upper bound.} 
Sanov’s theorem (large--deviation principle for types) states that for
any Borel set $\mathcal A\subseteq\Delta_k$,
$$
-\inf_{\mathbf{q}\in\mathring{\mathcal A}}D_{\mathrm{KL}}(\mathbf{q}\|\mathbf p)
\le
\liminf_{n\to\infty}\frac1n\log
\mathbb P\left(\hat{\mathbf p}_n\in\mathcal A\right)
\le
\limsup_{n\to\infty}\frac1n\log
\mathbb P\left(\hat{\mathbf p}_n\in\mathcal A\right)
\le
-\inf_{\mathbf{q}\in\overline{\mathcal A}}D_{\mathrm{KL}}(\mathbf q\|\mathbf p),
$$
where $\mathring{\mathcal A}$ and $\overline{\mathcal A}$ denote the interior and closure of $\mathcal{A}$, respectively. 

For our purposes, let $\mathcal{A} = \mathcal{B}$ as defined in Eq. (\ref{eq:bad_set}). Then
$$
\mathring{\mathcal B} = \{\mathbf q \in \Delta_k : q_{c^\star} < q_j \text{ for some } j \neq c^\star \}, 
\quad \overline{\mathcal B} = \mathcal B,
$$
since $\mathcal{B}$ is closed.

    \paragraph{Step 2. Error event as a type set.}
    The majority rule is \emph{incorrect} (i.e. $\hat{c}_n = \arg\max_j \,\hat{{p}}_{n,j}\neq c^\star$) 
    exactly when $\hat{\mathbf p}_n\in\mathcal B$. 
Hence, applying Sanov’s bounds yields
$$
\mathbb P\left[\hat{c}_n\neq c^\star\right] = 
\exp\left(-n\,\inf_{\mathbf q\in\mathcal B}D_{\mathrm{KL}}(\mathbf q\|\mathbf p)
+o(n)\right).
$$
    \paragraph{Step 3. Positivity of the exponent.}
    If $p_{c^\star}>p_j$ for every $j \neq c^\star$, then the true distribution satisfies $\mathbf p\notin\mathcal B$.
    The infimum of the KL divergence over $\mathcal B$ is therefore attained on the boundary, i.e., at some $\mathbf q^\star \in \mathcal B$ with $q^\star_{c^\star} = q^\star_{j}$ for some $j \neq c^\star$. 
Thus, the large-deviation exponent is
$$
I^\star (\mathbf{p})
=\min_{j \neq c^\star}\;
\inf_{\mathbf{q}:\,q_{c^\star}=q_j}D_{\mathrm{KL}}(\mathbf q\|\mathbf p)
>0,
$$
and the error probability decays exponentially
$$
\mathbb P[\hat{c}_n \neq c^\star] = \exp(-n I^\star(\mathbf p) + o(n)).
$$
\end{proof}
\subsubsection{Sanov exponent}
The Sanov exponent $I^\star(\mathbf{p})$ admits a closed-form expression. We provide a detailed derivation below.

Recall that our objective is to compute
\begin{equation}\label{eq:bad_set_sanov_exponent}
    I^\star(\mathbf{p}) = \inf_{\mathbf{q}\in\mathcal{B}} D_\text{KL}(\mathbf{q}\Vert \mathbf{p}),\quad\,\, \mathcal{B} = \{\mathbf{q}\in \Delta_k:\,q_{c^\star}\leq \max_{j \neq c^\star} q_j\}.
\end{equation}
Let the runner-up (second-largest competitor) be
$${j^\star} = \arg\max_{j \neq c^\star}\,\, p_j.$$ 
Then the optimisation problem can be equivalently written as
$$
I^\star(\mathbf{p}) = \min_{\mathbf q \in \Delta_k}\left[\sum_{i=1}^k q_i \log\left (\frac{q_i}{p_i}\right) \,: q_{j^\star} \geq q_{c^\star}\right].
$$
Introducing Lagrange multipliers, we define the Lagrangian
$$
\mathcal{L}(q, \lambda, \mu) = \sum_{i=1}^kq_i\log \frac{q_i}{p_i} + \lambda\left(\sum_{i=1}^k q_i - 1\right) + \mu(q_{c^\star} - q_{j^\star}),
$$
where $\lambda\in\mathbb{R}$ and $\mu\geq 0$, to enforce $q_{c^\star}\leq q_{j^\star}$. 
The first-order conditions yield
$$
\partial_{q_i}\mathcal{L} = \log q_i + 1 - \log p_i + \lambda = 0, \quad \,\, \text{for \,\,\,}i \neq c^\star, j^\star,
$$
which implies
$$
q_i = p_i e^{-(1 + \lambda)}, \quad \,\, \text{for \,\,\,}i \neq c^\star, j^\star.
$$
Similarly, for $q_{c^\star}$ and $q_{j^\star}$ we obtain
$$
\begin{aligned}
q_{c^\star} = p_{c^\star} e^{-(1+\lambda+\mu)},\qquad
q_{j^\star}=p_{j^\star}e^{-(1+\lambda-\mu)}.
\end{aligned}
$$
Defining $Z = e^{-(1+\lambda)}$ and $s = e^{\mu}$, we can rewrite the solution as
$$
\begin{aligned}
q_i &= p_i Z, \quad\,\, i \neq c^\star, j^\star\\
q_{c^\star} &= p_{c^\star} Z/s \\
q_{j^\star} &= p_{j^\star}Zs.
\end{aligned}
$$
Solving for $s$, we have
$$
s = \sqrt{\frac{q_{j^\star}}{p_{j^\star}} \frac{p_{c^\star}}{q_{c^\star}}}.
$$
Enforcing the constraint $q_{j^\star} \geq q_{c^\star}$ gives
$$
s \geq \sqrt{\frac{p_{c^\star}}{p_{j^\star}}}.
$$
On the other hand, enforcing the simplex constraint $\sum_i q_i = 1$ gives
$$
Z\left[(1 - p_{c^\star} - p_{j^\star}) + \frac{p_{c^\star}}{s} + p_{j^\star}s\right] = 1.
$$
    Note that $Z>0$. Substituting this, the KL divergence can be expressed as a function of $s$ (since $Z$ itself depends on $s$)
    \begin{align*}
    D_{\text{KL}}(\mathbf{q}(s)\Vert \mathbf{p} ) &= Z\log Z\left[(1-p_{c^\star}-p_{j^\star})+\frac{p_{c^\star}}{s}+p_{j^\star}s\right]
    +Z\log s\left(p_{j^\star}s-\frac{p_{c^\star}}{s}\right) 
    \\&= \log Z + Z \log s\left(p_{j^\star} s - \frac{p_{c^\star}}{s}\right).
    \end{align*}
    Minimising over $\mathbf q$ is equivalent to optimising over $s \geq \sqrt{p_{c^\star}/p_{j^\star}}$.   
    Focusing on the first term, we observe that
    $$
    \frac{d}{ds}\log Z = -\frac{-p_{c^\star}/s^2 + p_{j^\star}}{\left((1 - p_{c^\star} - p_{j^\star}) + p_{c^\star}/s + p_{j^\star}s\right)} \leq 0, \quad\mbox{ for } s \geq \sqrt{p_{c^\star}/p_{j^\star}}.
    $$
    Therefore, $\log Z$ is strictly decreasing for $s\in\left(\sqrt{p_{c^\star}/p_{j^\star}}, \infty\right)$. 

Furthermore, the derivative of the KL divergence with respect to $s$ is
\begin{align*}
\frac{d}{ds}D_{\text{KL}}(\mathbf{q}(s)\Vert \mathbf{p} ) &= Z \log s\left(p_{j^\star}  + \frac{p_{c^\star}}{s^2} -Zs\left(p_{j^\star}-\frac{p_{c^\star}}{s^2}\right)^2\right), \quad\,\, s \geq \sqrt{p_{c^\star}/p_{j^\star}}> 0.
\end{align*}
Note that 
\begin{align*}
 s\left[\left(p_{j^\star}+\frac{p_{c^\star}}{s^{2}}\right)
- Z\left(p_{j^\star}-\frac{p_{c^\star}}{s^2}\right)^2
\right] &\geq \left[\left(p_{j^\star}s+\frac{p_{c^\star}}{s}\right)
-
\frac{(p_{j^\star} s-p_{c^\star}/s)^{2}}{\,p_{j^\star} s + {p_{c^\star}}/{s}}
\right] \\&\geq \left[\left(p_{j^\star}s+\frac{p_{c^\star}}{s}\right)
\;-\;
\frac{(p_{j^\star} s+p_{c^\star}/s)^{2}}{\,p_{j^\star} s + p_{c^\star}/s}
\right]  \geq 0.
\end{align*}
In particular, for $s>\sqrt{p_{c^\star}/p_{j^\star}}$
$$
 s\left[\left(p_{j^\star}+\frac{p_{c^\star}}{s^{2}}\right)
- Z\left(p_{j^\star}-\frac{p_{c^\star}}{s^2}\right)^2
\right] >0.
$$
Using this together with the fact that $Z>0$ and $\log s>0$ (since $s> 1$), it follows that
$$
\frac{d}{ds}D_{\text{KL}}(\mathbf{q}(s)\Vert \mathbf{p} )>0, \quad\mbox{ for } s > \sqrt{p_{c^\star}/p_{j^\star}}.
$$
Hence, $D_{\text{KL}}(\mathbf{q}(s)\Vert \mathbf{p} )$ is strictly increasing for $s\in(\sqrt{p_{c^\star}/p_{j^\star}}, \infty)$. The minimum is therefore attained at $s = \sqrt{p_{c^\star}/p_{j^\star}}$, which leads to
\begin{align*}
I^\star(\mathbf{p}) &=\min_{\substack{\mathbf{q}\in\Delta_k\\q_{j^\star}\ge q_{c^\star}}} 
D_{\text{KL}}(\mathbf{q}\Vert \mathbf{p})
=-\log \left(1-p_{c^\star}-p_{j^\star}+2\sqrt{p_{c^\star} p_{j^\star}}\right) \\
&= -\log\left(1-\left(\sqrt{p_{c^\star}}-\sqrt{p_{j^\star}}\right)^2\right).
\end{align*}

\begin{remark}
\leavevmode
\begin{enumerate}
\item If there are multiple runners-up, there may be several minimisers $\mathbf q^\star$ in the set $\mathcal{B}$ defined in Eq. (\ref{eq:bad_set_sanov_exponent}), but they all yield the same KL value. Therefore $I^\star(\mathbf{p})$ remains unchanged regardless of the number of runners-up.
\item If $p_{c^\star} = p_{j^\star}$, then $I^\star(\mathbf{p}) = 0$.
\end{enumerate}
\end{remark}
\vspace{8pt}
\begin{remark}
By expanding the Sanov exponent in terms of the probability gap, we recover the error rate obtained via direct application of the central limit theorem. Specifically, for $\delta = p_{c^\star} - p_{j^\star} \ll p_{j^\star}$,
$$
I^\star(\mathbf{p}) = \frac{\delta^2}{4 p_{j^\star}}\left[1 + O(\delta/p_{j^\star})\right].
$$
On the other hand, since $\sigma_{j^\star}^2 = 2p_{j^\star} + \delta - \delta^2$, we have $
2 \sigma_{j^\star}^2 = 4 p_{j^\star} + O(\delta).
$
Therefore,
$$
\frac{\delta^2}{2\sigma_{j^\star}^2} = \frac{\delta^2}{4p_{j^*}}[1 + O(\delta/p_{j^*})],
$$
which yields
$$
I^\star(\mathbf{p}) = \frac{\delta^2}{2\sigma_{j^\star}^2} + O(\delta^3).
$$
\end{remark}

\subsubsection{Bahadur-Rao correction}
The random variable $Y_i^j = \mathbf 1\{X_i=c^\star\}-\mathbf 1\{X_i=j\} \in\{-1,0,1\}$ is integer-valued with span $d=1$ and has logarithmic moment generating function
$$
\Lambda_Y(\lambda) = \log M(\lambda) = \log \mathbb{E}\left[e^{\lambda Y_i^j}\right] = \log\left(p_{c^\star} e^{\lambda} + p_j e^{-\lambda} + 1- (p_{c^\star}+p_j)\right).
$$
Consider $S_j^{(n)}=-\sum_i^n Y_i^j = -Z_j^{(n)}$. We have that 
$$
\Lambda_{-Y}(\lambda) = \Lambda_Y(-\lambda). 
$$
Let $\lambda^\star =\tfrac{1}{2}\log(p_j/p_{c^\star}) < 0$ denote the minimiser of the moment generating function $M(\lambda)$ and define $\eta = -\lambda^\star>0$. 
Then, 
$$
\Lambda_{-Y}'(\eta) = -\Lambda'_Y(-\eta) = -\Lambda'_Y(\lambda^\star) =0.
$$
We are interested in the event $S_j^{(n)}\geq q$, where $q=\Lambda_{-Y}'(\eta)=0$. 
By \citet[Theorem 3.7.4. (b)]{dembo2010ldp}, as $n\to \infty$, we obtain the exact asymptotics 
$$
\mathbb{P}\left[S_j^{(n)}\geq nq\right] = \frac{1+o(1)}{(1-\exp(\lambda^\star))\sqrt{2\pi n\Lambda''_{-Y}(-\lambda^\star)}}\exp\left(-n\ \Lambda^\star_{-Y}(q)\right),
$$
where $\Lambda^\star_{-Y}(q)$ is the Legendre transform given by
\begin{align*}
 \Lambda^\star_{-Y}(q) &= \eta q - \Lambda_{-Y}(\eta) = -\Lambda_{-Y}(-\lambda^\star) = -\Lambda_{Y}(\lambda^\star)\\
 &=-\log\left(1-\left(\sqrt{p_{c^\star}}-\sqrt{p_j}\right)^2\right)
\end{align*}
and 
$$
\Lambda''_{-Y}(-\lambda^\star) = \Lambda_Y(\lambda^\star) = \frac{M''(\lambda^\star)}{M(\lambda^\star)} = \frac{2\sqrt{p_{c^\star} p_j}}{1-(\sqrt{p_{c^\star}}- \sqrt{p_j})^2}:=\tilde \sigma_j^2.
$$
Finally, we have that as $n\to\infty$,
\begin{align*}
    \mathbb{P}\bigl[\hat{c}_n&\neq c^\star\bigr]
    \le \sum_{j\neq c^\star}\mathbb{P}\bigl[Z_j^{(n)}\le 0\bigr]
    =\sum_{j\neq c^\star}\mathbb{P}\bigl[S_j^{(n)}\ge 0\bigr]\\
    &=(1+o(1))\sum_{j\neq c^\star}\frac{1}{\sqrt{2\pi n}(1-\exp(1/2\log(p_j/p_{c^\star})))\tilde\sigma_j}\exp\left(n\log(1-(\sqrt{p_{c^\star}}-\sqrt{p_j})^2)\right)\\
    &\sim \frac{1}{\sqrt{2\pi n}(1-\exp(1/2\log(p_{j^\star}/p_{c^\star})))\tilde\sigma_{j^\star}}\exp\left(n\log(1-(\sqrt{p_{c^\star}}-\sqrt{p_{j^\star}})^2)\right)\\
    &\sim \frac{1}{\sqrt{2\pi n}(1-\exp(1/2\log(p_{j^\star}/p_{c^\star})))\tilde\sigma_{j^\star}}\exp\left(-n I^\star(\mathbf{p})\right),
\end{align*}
where
${j^\star} = \arg\max_{j \neq c^\star}\,\, p_j$ is the runner-up.

\subsection{Comparison of the different bounds}\label{app:subsec_comparison_bounds}
We perform numerical experiments on synthetic examples to empirically verify the tightness of the derived bounds. 
We consider a categorical probability distribution with  $k=3$ categories and a small probability gap $\delta = p_{{c}^\star}-p_{j^\star}$. In particular, we set $p_1 = 0.38$, $p_2 = 0.35$,  $p_3 = 0.27$.   
To compute the empirical estimates of $P[\widehat{c}_n \neq c^\star]$, we employ a Monte Carlo approach with $10^6$ samples.  

\begin{figure}[!h]
\begin{center}
\includegraphics[width=0.9\linewidth]{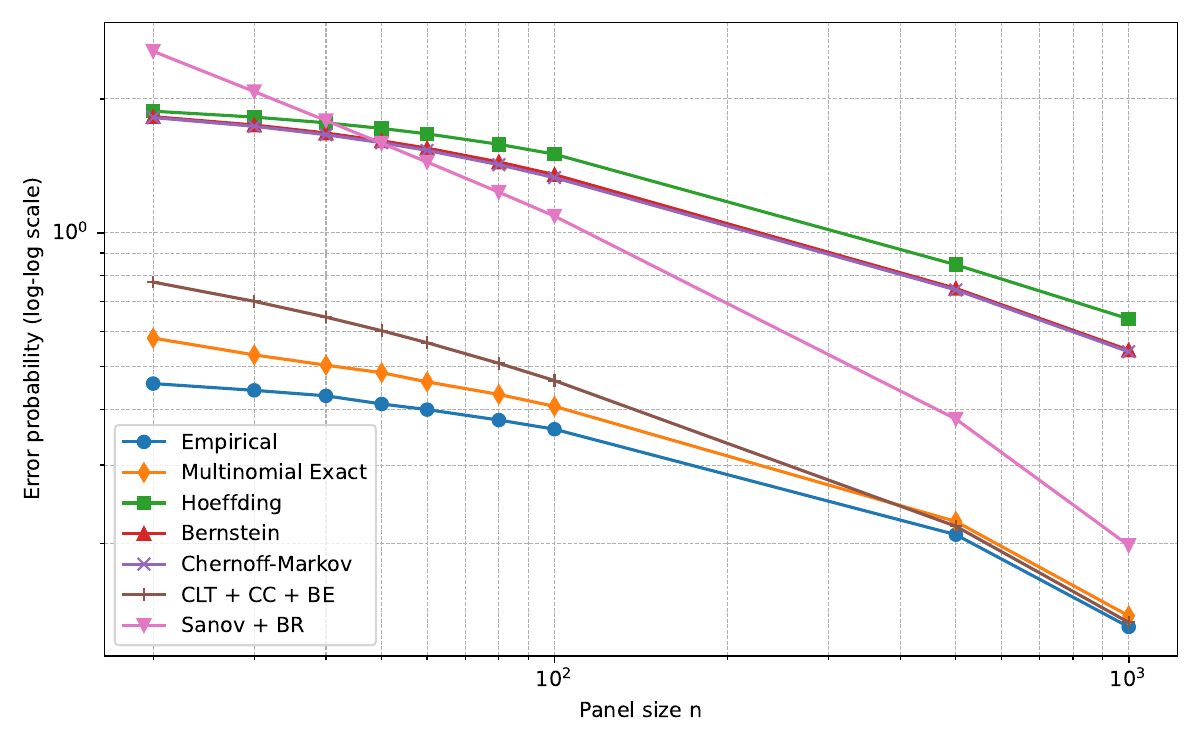}
\end{center}
\caption{Comparison of empirical and theoretical bounds on $P[\widehat{c}_n \neq c^\star]$ for the probability distribution $\mathbf{p} = (0.38, 0.35, 0.27)$.}
\label{fig:comparison_bounds_majority_vote}
\end{figure}

The results are shown in Figure \ref{fig:comparison_bounds_majority_vote}. We observe that the CLT bound, with continuity and Berry–Esseen corrections (CLT + CC + BE), provides a very tight estimate that converges to the exact multinomial result as the panel size of voters increases.
In contrast, the Hoeffding, Bernstein and Chernoff bounds are noticeably looser. Finally, the Sanov bound with Bahadur-Rao correction (Sanov + BR) is expected to become increasingly tight for larger panels. 

\section{Analysis of the stopping rule}\label{app:details_stopping_rule}
We provide a more detailed analysis of the stopping rule introduced in Section \ref{sec:stopping_rule}.
For a detailed treatment of $e$-values for hypothesis testing, see \cite{STA-002}.

\subsection{Anytime-valid $e$-processes}
\label{subsec_app:mmc_recursive}

Recall the predictable, recursive counts
\[
A_{n-1}:=\widehat c_{\,n-1},\quad B_{n-1}:=j^\star_{\,n-1},\qquad
\begin{aligned}
&s_n=s_{n-1}+\mathbf 1\{X_n=A_{n-1}\},\\
&f_n=f_{n-1}+\mathbf 1\{X_n=B_{n-1}\},\\
&o_n=o_{n-1}+\mathbf 1\{X_n\notin\{A_{n-1},B_{n-1}\}\},
\end{aligned}
\]
and the effective sizes \(M_n:=s_n+f_n\) (A vs B) and \(T_n:=s_n+o_n\) (A vs others).

Let $(\pi^{\mathrm{run}}_n)_{n\ge1}$ and $(\pi^{\mathrm{oth}}_n)_{n\ge1}$ be  {predictable} priors (each $\pi_n$ is $\mathcal F_{n-1}$-measurable) 
supported on $(1/2,1]$.
Define the two mixture $e$-processes recursively (with optional skipping) by
\begin{align*}
e^{\mathrm{run}}_n
&=\begin{cases}
e^{\mathrm{run}}_{n-1}\cdot 2\!\displaystyle\int \theta\,\pi^{\mathrm{run}}_n(d\theta), & X_n = A_{n-1},\\[1mm]
e^{\mathrm{run}}_{n-1}\cdot 2\!\displaystyle\int (1-\theta)\,\pi^{\mathrm{run}}_n(d\theta), & X_n = B_{n-1},\\[1mm]
e^{\mathrm{run}}_{n-1}, & \text{otherwise,}
\end{cases}\\[2mm]
e^{\mathrm{oth}}_n
&=\begin{cases}
e^{\mathrm{oth}}_{n-1}\cdot 2\!\displaystyle\int \lambda\,\pi^{\mathrm{oth}}_n(d\lambda), & X_n = A_{n-1},\\[1mm]
e^{\mathrm{oth}}_{n-1}\cdot 2\!\displaystyle\int (1-\lambda)\,\pi^{\mathrm{oth}}_n(d\lambda), & X_n \notin \{A_{n-1},B_{n-1}\},\\[1mm]
e^{\mathrm{oth}}_{n-1}, & \text{if } X_n = B_{n-1},
\end{cases}
\end{align*}
with $e^{\mathrm{run}}_0=e^{\mathrm{oth}}_0=1$.
By aggregating the per-round factors, we have the equivalent expression (\ref{eq:e_process_A_vs_B}) and (\ref{eq:e_process_A_vs_OTHERS}).

Before proving Theorem \ref{thm:mmc_eprocess_recursive}, we introduce the following auxiliary lemma.

\begin{lemma}[One-step mixture bound]\label{lem:onestep_recursive}
Let $\pi$ be any probability measure on $(1/2,1]$ and define $m:=\int u\,\pi(du)\in(1/2,1]$.
If $Y\sim\mathrm{Ber}(\vartheta)$ then
\[
\mathbb E\!\Big[\,2\!\int u^{Y}(1-u)^{1-Y}\,\pi(du)\,\Big] \;=\; 2\big(1-m+\vartheta(2m-1)\big),
\]
which is increasing in $\vartheta$ and $\le 1$ for all $\vartheta\le \tfrac12$, with equality at $\vartheta=\tfrac12$.
\end{lemma}

\begin{proof}
$\mathbb E[u^{Y}(1-u)^{1-Y}]=\vartheta u + (1-\vartheta)(1-u) = (1-u)+\vartheta(2u-1)$; integrating over $\pi$ yields the result.
\end{proof}

\begin{theorem}[Theorem \ref{thm:mmc_eprocess_recursive} restated]
Let $p_j=\mathbb P[X=j\mid pr]$. For the \emph{A vs B} test (leader vs runner-up), define
$\theta_n = \tfrac{p_{A_{n-1}}}{p_{A_{n-1}}+p_{B_{n-1}}}$ and the one-sided composite null
\[
H^{\mathrm{run}}_0:\quad \theta_n \le \tfrac12 \ \big(\text{equivalently $p_{A_{n-1}}\le p_{B_{n-1}}$}\big) \,\,\text{at every round $n$.}
\]
For the \emph{A vs others} test, define
$\lambda_n = \tfrac{p_{A_{n-1}}}{p_{A_{n-1}}+\sum_{j\notin\{A_{n-1},B_{n-1}\}}p_j} = \tfrac{p_{A_{n-1}}}{1-p_{B_{n-1}}}$
and the composite null
\[
H^{\mathrm{oth}}_0:\quad \lambda_n \le \tfrac12 \ \big(\text{equivalently $p_{A_{n-1}}\le \scriptstyle{\sum_{j\notin\{A_{n-1},B_{n-1}\}}}$$\, p_j$}\big) \,\,\text{at every round $n$.}
\]
Then $\{e^{\mathrm{run}}_n\}_{n\ge0}$ and $\{e^{\mathrm{oth}}_n\}_{n\ge0}$ defined in (\ref{eq:e_process_A_vs_B}), (\ref{eq:e_process_A_vs_OTHERS}) are non-negative test
\emph{supermartingales} w.r.t.\ $\{\mathcal F_n\}$, even with predictable, data-dependent priors and optional skipping.
Under the boundary (simple) nulls ($\theta_n\equiv\tfrac12$ or $\lambda_n\equiv\tfrac12$ on their informative rounds),
they are test \emph{martingales}. Consequently, by Ville’s inequality, for any stopping time,
\[
\sup_{\mathbb P\in H^{\mathrm{run}}_0}\ \mathbb P\Big(\sup_{n\ge0}e^{\mathrm{run}}_n\ge 1/\varepsilon\Big)\le\varepsilon,
\qquad
\sup_{\mathbb P\in H^{\mathrm{oth}}_0}\ \mathbb P\Big(\sup_{n\ge0}e^{\mathrm{oth}}_n\ge 1/\varepsilon\Big)\le\varepsilon.
\]
\end{theorem}

\begin{proof}
Fix $n\ge1$ and condition on $\mathcal F_{n-1}$, then $A_{n-1},B_{n-1}$ and the priors
$\pi^{\mathrm{run}}_n,\pi^{\mathrm{oth}}_n$ are deterministic. For the \emph{A vs B} process,
\[
\frac{e^{\mathrm{run}}_n}{e^{\mathrm{run}}_{n-1}}
=\begin{cases}
2\!\int \theta'\,\pi^{\mathrm{run}}_n(d\theta'), & X_n=A_{n-1},\\
2\!\int (1-\theta')\,\pi^{\mathrm{run}}_n(d\theta'), & X_n=B_{n-1},\\
1, & \text{otherwise.}
\end{cases}
\]
Write $q_n:=p_{A_{n-1}}+p_{B_{n-1}}$ and $\theta_n:=p_{A_{n-1}}/q_n$ (if $q_n=0$ the step is skipped a.s.).
Then, under $H^{\mathrm{run}}_0$ we have $\theta_n\le \tfrac12$ and
\[
\mathbb E\!\left[\frac{e^{\mathrm{run}}_n}{e^{\mathrm{run}}_{n-1}}\ \Big|\ \mathcal F_{n-1}\right]
= q\cdot \mathbb E\!\Big[\,2\!\int \theta'^{Y}(1-\theta')^{1-Y}\,\pi^{\mathrm{run}}_n(d\theta')\,\Big]
\;+\; (1-q)\cdot 1,
\]
where $Y\sim\mathrm{Ber}(\theta_n)$ on the informative event. By Lemma~\ref{lem:onestep_recursive},
the bracketed expectation is $\le 1$ for $\theta_n\le 1/2$, hence the whole conditional expectation is $\le 1$.
Thus $\{e^{\mathrm{run}}_n\}$ is a test supermartingale (and a martingale at $\theta_n=1/2$).

Similarly, for the \emph{A vs others} process,
\[
\frac{e^{\mathrm{oth}}_n}{e^{\mathrm{oth}}_{n-1}}
=\begin{cases}
2\!\int \lambda'\,\pi^{\mathrm{oth}}_n(d\lambda'), & X_n=A_{n-1},\\
2\!\int (1-\lambda')\,\pi^{\mathrm{oth}}_n(d\lambda'), & X_n\notin\{A_{n-1},B_{n-1}\},\\
1, & \text{if } X_n=B_{n-1}.
\end{cases}
\]
Let $r_n:=1-p_{B_{n-1}}$ and $\lambda_n:=p_{A_{n-1}}/r_n$. Under $H^{\mathrm{oth}}_0$, $\lambda_n\le 1/2$ and the same calculation gives
\[
\mathbb E\!\left[\frac{e^{\mathrm{oth}}_n}{e^{\mathrm{oth}}_{n-1}}\ \Big|\ \mathcal F_{n-1}\right]
= r\cdot \mathbb E\!\Big[\,2\!\int \lambda'^{Z}(1-\lambda')^{1-Z}\,\pi^{\mathrm{oth}}_n(d\lambda')\,\Big]
\;+\; (1-r)\cdot 1 \;\le\; 1,
\]
where $Z\sim\mathrm{Ber}(\lambda_n)$ on the informative event. Ville’s inequality yields the stated time-uniform bounds.
\end{proof}

\subsection{Estimation of $\widehat{\varepsilon}\geq \mathbb{P}[\hat{c}_n\neq c^\star]$}\label{app:subsec_estimator_proability}
For convenience, we describe below how to compute $1-\hat\varepsilon$, which provides a lower bound $1-\hat{\varepsilon}\leq\mathbb{P}[\widehat{c}_n = c^\star]$.
Before doing so, recall that if $a$ and $b$ denote two possible outcomes of a multinomial distribution, then
$$
\mathbb{P}[p_a>p_b] = \mathbb{P}\left[\theta_{ab}=\tfrac{p_b}{p_a+p_b}<\tfrac{1}{2}\right].
$$
This probability can be estimated using a Beta approximation. Assuming a Beta prior on $\theta_{ab}$ with parameters $(1,1)$, and letting $N_a$ and $N_b$ denote the observed counts for each outcome, we obtain
$$
\mathbb{P}[\theta_{ab}<\tfrac{1}{2}] = \frac{\Gamma(N_a,N_b)}{\Gamma(N_a)\Gamma(N_b)}\int_0^{1/2} \theta^{N_a-1} (1-\theta)^{N_b-1}\, d\theta :=I_{1/2}(N_a, N_b).
$$
Therefore, we have
\begin{align}
    \mathbb{P}[\hat{c}_n = c^\star]&\gtrsim \min\left(\mathbb{P}(p_{\hat{c}_n}>p_{{j^\star_n}}), \mathbb{P}(p_{\hat{c}_n}>p_{{\hat{o}_n}})\right)\nonumber\\
    &\approx \min\left(I_{1/2}(f_n + 1, s_n+1), I_{1/2}({o}_n + 1, s_n+1)\right).\label{eq:lower_bound_for_estimator_correct_no_true}
\end{align}

\subsection{Stopping time}\label{app:subsec_stopping_time}
When the prior is of the form 
$$
\Pi_n(d\bm\theta)=\prod_{i=1}^n\delta_{\theta^\star}(d\theta_i)
$$
the corresponding $e$-process is given by
\[
e_n
= 2^{M}(\theta^\star)^{s}(1-\theta^\star)^{f}.
\] 
If the data-generating process follows a Bernoulli distribution with parameter $\theta^\star$, then $s \approx M \theta^\star$, yielding
\begin{align*}
\log e_{n} &= M \left(\frac{s}{M}\log(2\theta^\star) +\left(1-\frac{s}{M}\right)\log 2(1-\theta^\star)\right)\nonumber\\
&\approx M \left(\theta^\star\log(2\theta^\star) +\left(1-\theta^\star\right)\log 2(1-\theta^\star)\right)\nonumber\\
&=M D_\text{KL}(\text{Ber}(\theta^\star)\Vert \text{Ber}(1/2)).
\end{align*}
Therefore, the number of informative rounds required until stopping is
\begin{align*}
M_\tau &= \inf\{M: \ \log e_n\geq \log(1/\varepsilon)\} \\
&= \inf\left \{M: \ M D_\text{KL}(\text{Ber}(\theta^\star)\Vert \text{Ber}(1/2))\geq \log(1/\varepsilon)\right\} \\
&\approx \frac{\log(1/\varepsilon)}{D_\text{KL}(\text{Ber}(\theta^\star)\Vert \text{Ber}(1/2))}.
\end{align*}
Note that when $\theta^\star$ is close to $1/2$, we can approximate $D_\text{KL}(\text{Ber}(1/2 +\varepsilon)\Vert \text{Ber}(1/2)) \approx 2\varepsilon^2$, which leads to
\small
\begin{equation*}
M_{\text{lead, runner-up}} \approx \tfrac{2\left(p_{\hat{c}} + p_{j^\star}\right)^2}{\left(p_{\hat{c}} - p_{j^\star}\right)^2}\log(1/\varepsilon), \quad\quad M_{\text{lead,others}} \approx \tfrac{2\left(1- p_{j^\star}\right)^2}{\left(2p_{\hat{c}} + p_{j^\star}-1\right)^2}\log(1/\varepsilon).
\end{equation*}
\normalsize
Finally, since the expected number of rounds until an informative one occurs is
\small
\begin{equation*}
K_{\text{lead, runner-up}} = \tfrac{1}{p_{\hat{c}} + p_{j^\star}}, \quad\quad K_{\text{lead,others}} = \tfrac{1}{1- p_{j^\star}},
\end{equation*}
\normalsize
due to the properties of the geometric distribution, we find that the total number of rounds required is approximately $N = K\cdot M$
\small
\begin{equation*}
N_{\text{lead, runner-up}} \approx \tfrac{2\left(p_{\hat{c}} + p_{j^\star}\right)}{\left(p_{\hat{c}} - p_{j^\star}\right)^2}\log(1/\varepsilon), \quad\quad N_{\text{lead,others}} \approx \tfrac{2\left(1- p_{j^\star}\right)}{\left(2p_{\hat{c}} + p_{j^\star}-1\right)^2}\log(1/\varepsilon).
\end{equation*}
\normalsize
Moreover, when $ p_{\hat{c}}-p_{j^\star}\ll p_{j^\star}$, the ratio $\scriptstyle \left(p_{\hat{c}} + p_{j^\star}\right)/{\left(p_{\hat{c}} - p_{j^\star}\right)^2}$ is approximately $\text{SNR}(\Delta_{j^\star})^{-1}$, where $\Delta_{j^\star}=\scriptstyle{ \mathbf 1\{X=\hat c\}-\mathbf 1\{X=j^\star\}}$.

\subsection{Algorithms for truncated \texorpdfstring{$\mathrm{Beta}(a,b)$}{Beta(a,b)} and updating point prior}
We provide pseudocode for implementing the MMC stopping rule with the truncated $\mathrm{Beta}(a,b)$ shared-parameter prior (Algorithm \ref{alg:mmc_truncbeta}) and the shared-parameter point prior presented in B.1 (Algorithm \ref{alg:mmc_shared_point}).

\begin{algorithm}[t]
\small{
\caption{MMC stopping rule with truncated $\mathrm{Beta}(a,b)$ prior}
\label{alg:mmc_truncbeta}
\begin{algorithmic}[1]
\Require confidence level $\varepsilon$, budget $N_{\text{budget}}$, hyperparameters $a,b>0$; deterministic tie-break rule
\State \textbf{Init:} $n\gets 0$; for all $j\in\{1,\dots,k\}$ set label counts $N_j\gets 0$; 
$s_0=f_0=o_0\gets 0$; $e^{\mathrm{run}}_0=e^{\mathrm{oth}}_0\gets 1$
\State Define $\mathsf{B}_{>1/2}(a,b)=\int_{1/2}^1 t^{a-1}(1-t)^{b-1}\,dt$
\While{True}
  \State \textbf{Predictable top-2:} set $A_n\gets\arg\max_j N_j$, $B_n\gets$ second largest (ties broken deterministically)
  \State \textbf{Cache counts (pre-update):} $\tilde s\gets s_n$, $\tilde f\gets f_n$, $\tilde o\gets o_n$
  \State \textbf{Draw a new vote:} sample $X\sim\mathbb P[\,\cdot\,| pr]$ 
  \State \textbf{Per-round ratio (A vs B):}
  \[
  \rho_{\mathrm{run}} \;=\;
  \begin{cases}
    2\,\dfrac{\mathsf{B}_{>1/2}(a+\tilde s+1,\; b+\tilde f)}{\mathsf{B}_{>1/2}(a+\tilde s,\; b+\tilde f)}, & X=A_n,\\[2mm]
    2\,\dfrac{\mathsf{B}_{>1/2}(a+\tilde s,\; b+\tilde f+1)}{\mathsf{B}_{>1/2}(a+\tilde s,\; b+\tilde f)}, & X=B_n,\\[2mm]
    1,& \text{otherwise.}
  \end{cases}
  \]
  \State \textbf{Per-round ratio (A vs others):}
  \[
  \rho_{\mathrm{oth}} \;=\;
  \begin{cases}
    2\,\dfrac{\mathsf{B}_{>1/2}(a+\tilde s+1,\; b+\tilde o)}{\mathsf{B}_{>1/2}(a+\tilde s,\; b+\tilde o)}, & X=A_n,\\[2mm]
    2\,\dfrac{\mathsf{B}_{>1/2}(a+\tilde s,\; b+\tilde o+1)}{\mathsf{B}_{>1/2}(a+\tilde s,\; b+\tilde o)}, & X\notin\{A_n,B_n\},\\[2mm]
    1,& X=B_n.
  \end{cases}
  \]
  \State \textbf{Update $e$-values:} $e^{\mathrm{run}}_{n+1}\gets e^{\mathrm{run}}_{n}\cdot \rho_{\mathrm{run}}$, \ \ $e^{\mathrm{oth}}_{n+1}\gets e^{\mathrm{oth}}_{n}\cdot \rho_{\mathrm{oth}}$
  \State \textbf{Update recursive counts:}
  \[
  (s_{n+1},f_{n+1},o_{n+1})=
  \begin{cases}
    (\tilde s+1,\tilde f,\tilde o), & X=A_n,\\
    (\tilde s,\tilde f+1,\tilde o), & X=B_n,\\
    (\tilde s,\tilde f,\tilde o+1), & \text{otherwise.}
  \end{cases}
  \]
  \State \textbf{Update label counts:} $N_X\gets N_X+1$; \ $n\gets n+1$
  \State \textbf{Check stop:} \textbf{if} $e^{\mathrm{run}}_{n}\ge 1/\varepsilon$ \textbf{and} $e^{\mathrm{oth}}_{n}\ge 1/\varepsilon$ \textbf{then}
     \State \hspace{1.5em} set $\hat c\gets \arg\max_j N_j$; \Return $(\hat c,\ \text{stopped})$
  \State \textbf{Budget:} \textbf{if} $n\ge N_{\text{budget}}$ \textbf{then} \Return $(\arg\max_j N_j,\ \text{abstained})$
\EndWhile
\end{algorithmic}}
\normalsize
\end{algorithm}

\begin{algorithm}[!ht]
{\small
\caption{MMC stopping rule with  updating point prior}
\label{alg:mmc_shared_point}
\begin{algorithmic}[1]
\Require Confidence level $\varepsilon$; budget $N_{\text{budget}}$; Dirichlet smoothing $(\alpha_A,\alpha_B,\alpha_O)>0$; clipping $\varepsilon\in(0,10^{-3}]$; deterministic tie-break rule
\State \textbf{Init:} $n\leftarrow 0$; for all $j\in\{1,\dots,k\}$ set label counts  $N_j\leftarrow 0$; $s_0=f_0=o_0\leftarrow 0$; $e^{\mathrm{run}}_0=e^{\mathrm{oth}}_0\leftarrow 1$
\While{True}
\State \textbf{Predictable top–2:} set $A_n \leftarrow \arg\max_j N_j$,\ \ $B_n \leftarrow$ second largest (break ties deterministically)
\State \textbf{Predictable total counts:} $L\gets s_n +  f_n +  o_n$
\State \textbf{Shared multinomial plug–in (Dirichlet–smoothed):}
$$\displaystyle \hat p_{A}\leftarrow \frac{ s_n+\alpha_A}{L+\alpha_A+\alpha_B+\alpha_O},\quad
         \hat p_{B}\leftarrow \frac{ f_n+\alpha_B}{L+\alpha_A+\alpha_B+\alpha_O}$$
$$\displaystyle \theta^\star_{n}\leftarrow \operatorname{clip}\!\Big(\frac{\hat p_{A}}{\hat p_{A}+\hat p_{B}},\;\tfrac12+\varepsilon,\;1-\varepsilon\Big),\quad
         \lambda^\star_{n}\leftarrow \operatorname{clip}\!\Big(\frac{\hat p_{A}}{1-\hat p_{B}},\;\tfrac12+\varepsilon,\;1-\varepsilon\Big)$$
\State \textbf{Draw a new vote:} sample $X\sim \mathbb P[\,\cdot\,| pr]$
\State \textbf{Update recursive counts:}
\[
(s_{n+1},f_{n+1},o_{n+1}) \!=\!
\begin{cases}
(s_n\!+\!1, f_n, o_n), & X=A_n,\\
(s_n, f_n\!+\!1, o_n), & X=B_n,\\
(s_n, f_n, o_n\!+\!1), & \text{otherwise}
\end{cases}
\]
\State \textbf{Update e–values:} $$e^{\mathrm{run}}_n
= 2^{s_{n+1}+f_{n+1}}(\theta_n^\star)^{s_{n+1}}(1-\theta_n^\star)^{f_{n+1}},\qquad
{e^{\mathrm{oth}}_n
= 2^{s_{n+1}+o_{n+1}}(\lambda_n^\star)^{s_{n+1}}(1-\lambda_n^\star)^{o_{n+1}}}.$$
\State \textbf{Update label counts:} $N_X\leftarrow N_X+1$;\quad $n\leftarrow n+1$
\State \textbf{Check stop:} \textbf{if} $e^{\mathrm{run}}_n\ge 1/\varepsilon$ \textbf{and} $e^{\mathrm{oth}}_n\ge 1/\varepsilon$ \textbf{then}
\State \hspace{1.2em} set $\hat c \leftarrow \arg\max_j N_j$;\; \Return $(\hat c,\ \text{stopped})$
\State \textbf{Budget:} \textbf{if} $n\ge N_{\text{budget}}$ \textbf{then} \Return $(\arg\max_j N_j,\ \text{abstained})$
\EndWhile
\end{algorithmic}}
\normalsize
\end{algorithm}

\subsection{General stopping rule}\label{app:subsec_general_stopping_rule}
The proposed stopping rule exploits the fact that although the space of possible LLM outputs may be large, the true distribution $\mathbb{P}[\cdot|pr]$ (for a given prompt $pr$) is typically concentrated on a subset of $m$ classes, with $m \ll k$, where $\{1, \dots, k\}$ is the total support.
We further assume that the conditional distribution over these top-$m$ classes is approximately uniform.
Based on this, we design a strategy that performs pairwise comparisons: between the leader and each of the top-$(m-1)$ runner-ups, and between the leader and the remaining classes. 

Similarly to the case $m=2$, at round $n\!\ge\!1$, \emph{before} observing $X_n$, set the predictable top-$m$ labels as
\[
A_{n-1}:=\widehat c_{\,n-1},\qquad B_{n-1}^{\, i}:=j^\star_{\,n-1, i}, \qquad B_{n-1}:=\{B_{n-1}^{\, 1}, \dots, B_{n-1}^{m-1}\},
\]
which are measurable w.r.t.\ $\mathcal F_{n-1}=\sigma(X_1,\dots,X_{n-1})$ (ties broken deterministically).
We maintain the following \emph{recursive, predictable} counts
\[
\begin{aligned}
&\textbf{Leader hits:}&& s_n \;=\; s_{n-1} + \mathbf 1\{X_n = A_{n-1}\}, \quad s_0=0,\\
&\textbf{$i$-th runner-up hits (for the A vs $\mathbf{B^{i}}$ test):}&& f_n^i \;=\; f_{n-1}^i + \mathbf 1\{X_n = B_{n-1}^{\,i}\}, \quad f_0^i=0,\\
&\textbf{Others hits (for the A vs others test):}&& o_n \;=\; o_{n-1} + \mathbf 1\{X_n \notin \{A_{n-1},B_{n-1}\}\}, \quad o_0=0.
\end{aligned}
\]
Thus the sample sizes are
\[
M_n^i := s_n + f_n^i,\qquad
T_n := s_n + o_n.
\]
Let $(\pi^{\mathrm{run}, i}_{n})_{n\ge1}$, $i=1, \dots, m-1$, and $(\pi^{\mathrm{oth}}_{n})_{n\ge1}$ be  {predictable} priors (i.e., $\mathcal F_{n-1}$-measurable) 
supported on $(1/2,1]$.

We define the $m$ mixture $e$-processes recursively (with optional skipping) by
\begin{align*}
e^{\mathrm{run}, i}_n
&=\begin{cases}
e^{\mathrm{run}, i}_{n-1}\cdot 2\!\displaystyle\int \theta\,\pi^{\mathrm{run}, i}_n(d\theta), & X_n = A_{n-1},\\[1mm]
e^{\mathrm{run}, i}_{n-1}\cdot 2\!\displaystyle\int (1-\theta)\,\pi^{\mathrm{run}, i}_n(d\theta), & X_n = B_{n-1}^{\, i},\\[1mm]
e^{\mathrm{run}, i}_{n-1}, & \text{otherwise,}
\end{cases}\\[2mm]
e^{\mathrm{oth}}_n
&=\begin{cases}
e^{\mathrm{oth}}_{n-1}\cdot 2\!\displaystyle\int \lambda\,\pi^{\mathrm{oth}}_n(d\lambda), & X_n = A_{n-1},\\[1mm]
e^{\mathrm{oth}}_{n-1}\cdot 2\!\displaystyle\int (1-\lambda)\,\pi^{\mathrm{oth}}_n(d\lambda), & X_n \notin \{A_{n-1},B_{n-1}\},\\[1mm]
e^{\mathrm{oth}}_{n-1}, & \text{if } X_n = B_{n-1},
\end{cases}
\end{align*}
with $e^{\mathrm{run}, i}_0=e^{\mathrm{oth}}_0=1$.
Thanks to Theorem \ref{thm:mmc_eprocess_recursive} $\{e^{\mathrm{run}, i}_n\}$, $i=1, \dots, m-1$, and $\{e^{\mathrm{oth}}_n\}$ are
non-negative test supermartingales under their respective composite nulls, and test martingales under the boundary nulls.

\subsection{Analysis of the stopping rule on synthetic data}\label{app:subsec_mmc_synthetic_data}
We analyse the performance of the proposed MMC stopping rule on synthetic data, focusing on the impact of the prior distribution choice.
To do so, we simulate different probability distributions over
$k=26$ classes and evaluate the performance as a function of the probability gap $\delta=p_{c^\star}- p_{j^\star}$, where $c^\star$ and $j^\star$ denote the true majority vote and the runner-up, respectively. 

Following Algorithm \ref{alg:stopping_rule}, we set the algorithm parameters to $\varepsilon=0.1$ (confidence level) and $N_{\text{budget}}=64$ (maximum budget).
This ensures that, at the final iteration, either the budget is reached or the following guarantee holds
$$
\mathbb{P}\left[\widehat{c}_n\neq c^\star\right]\leq \varepsilon.
$$

\begin{figure}[h!]
  \centering
  \begin{subfigure}{0.49\textwidth}
      \centering
      \includegraphics[width=\textwidth]{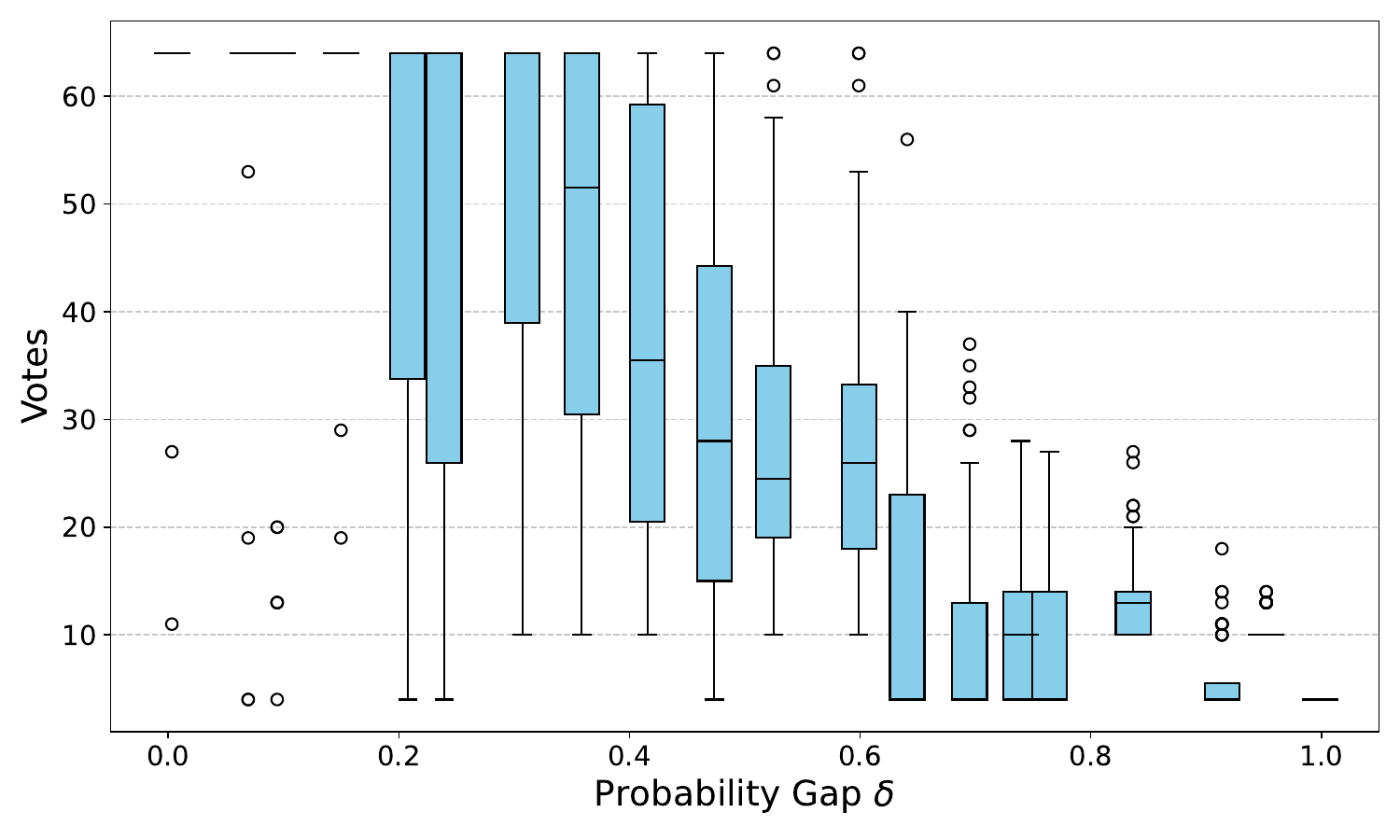}
      \caption{Truncated Beta prior (A).}
      \label{fig:votes_truncated_beta_prior}
  \end{subfigure}
  \hfill
  \begin{subfigure}{0.49\textwidth}
      \centering
      \includegraphics[width=\textwidth]{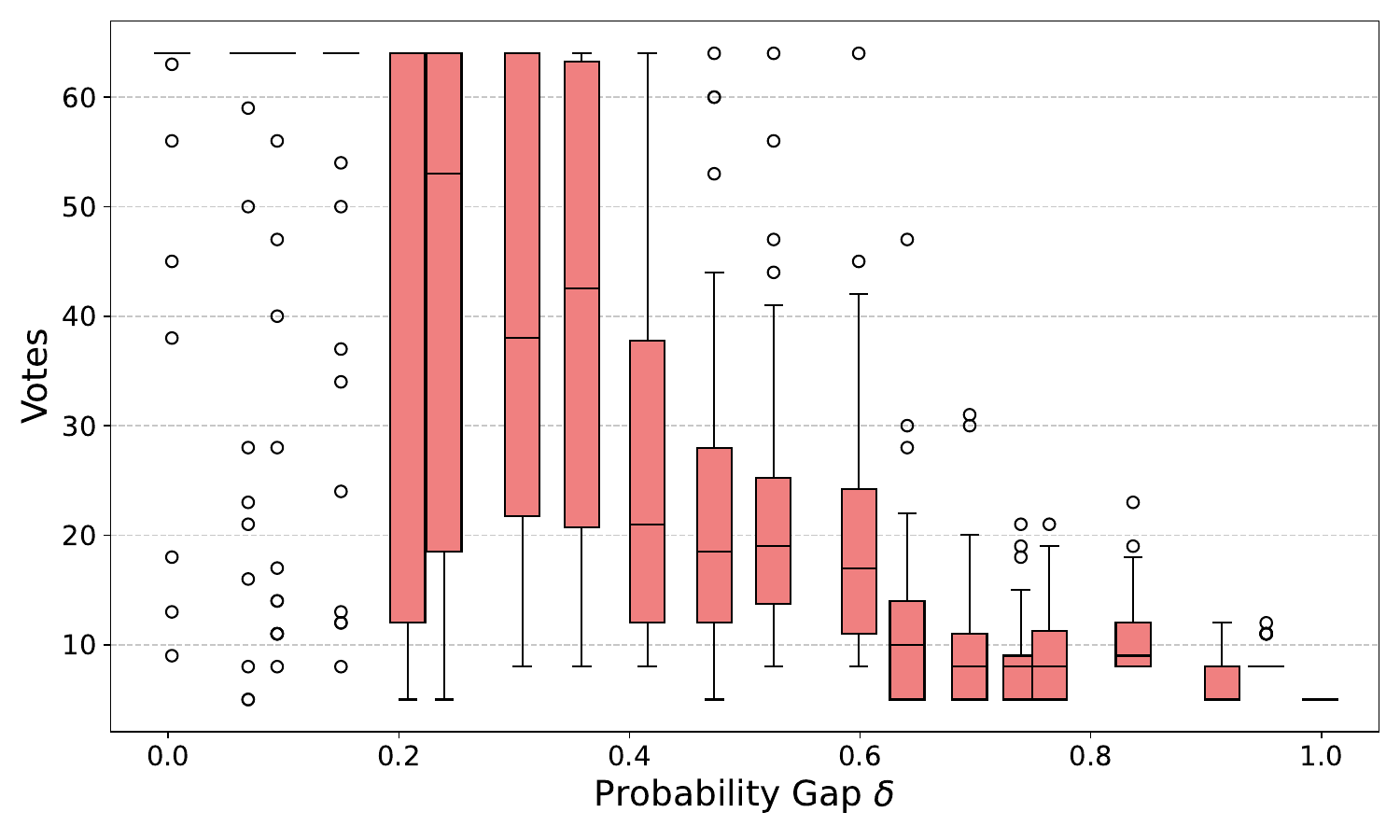}
        \caption{Updating point prior (B.1).}
      \label{fig:votes_delta_dirac_prior}
  \end{subfigure}
  \vfill
  \vspace{8pt}
    \begin{subfigure}{0.49\textwidth}
      \centering
      \includegraphics[width=\textwidth]{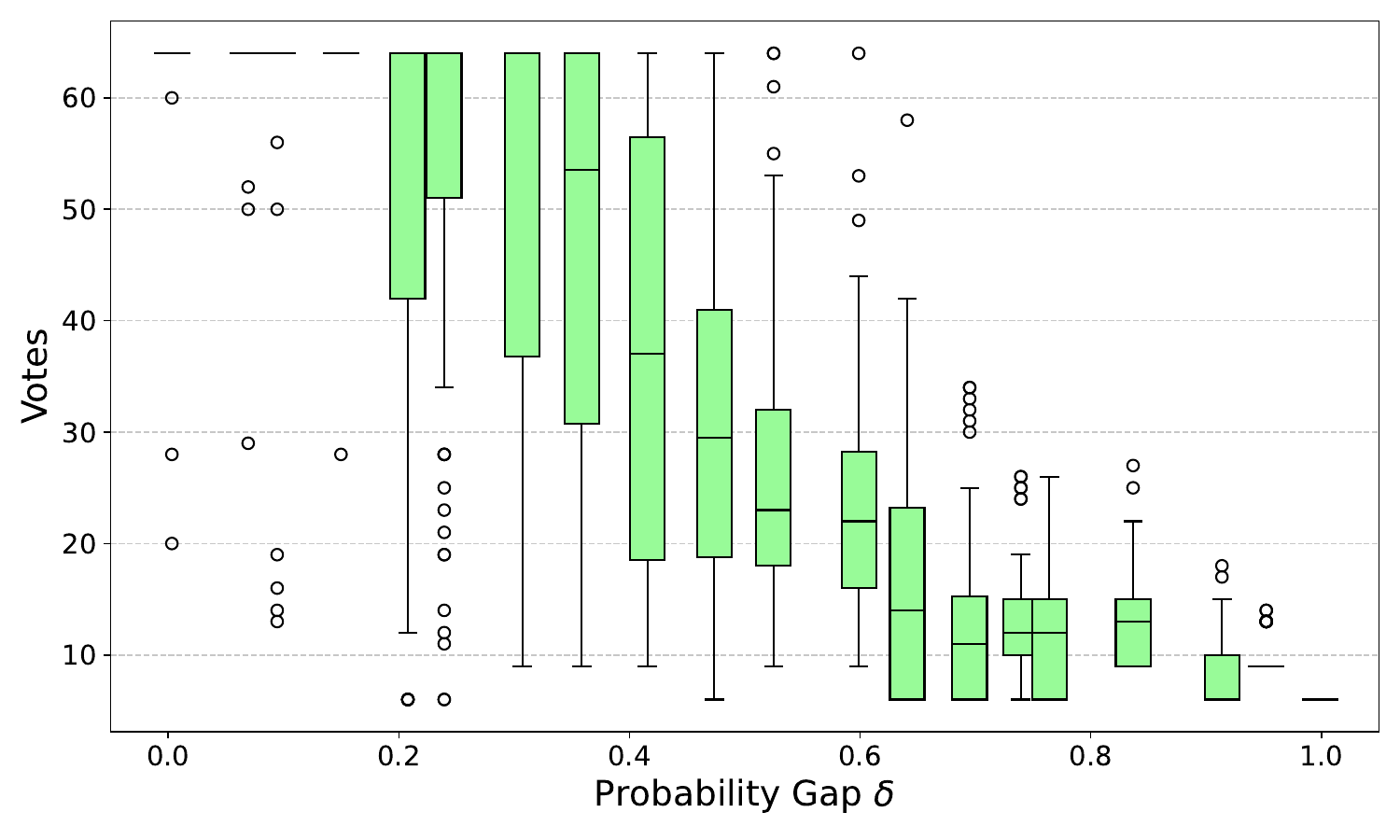}
        \caption{Updating point prior (B.2).}
      \label{fig:votes_delta_dirac_prior_ratio_updates}
  \end{subfigure}
  \caption{Boxplots showing the distribution of the number of votes required until stopping under the MMC rule as a function of the probability gap $\delta = p_{c^\star}- p_{j^\star}$. Results are shown for $\varepsilon = 0.1$ with a maximum budget of 64 votes.}
  \label{fig:synthetic_data_stopping_rule}
\end{figure}

Figure \ref{fig:synthetic_data_stopping_rule} presents boxplots of the number of votes required to stop under the MMC rule as a function of the probability gap $\delta$. 
For small values of $\delta$, the number of votes saturates at the maximum budget. 
As $\delta$ increases, the average number of votes required to guarantee the correctness of the majority vote decreases. 
Comparing the three prior choices for the same value of $\delta$, we observe that using an updating point prior with shared parameter, as presented in B.1 (Fig. \ref{fig:votes_delta_dirac_prior}) results in fewer votes to achieve statistical guarantees than either a truncated Beta prior with shared parameter (Fig. \ref{fig:votes_truncated_beta_prior}) or an updating point prior based on ratio updates, presented in B.2 (Fig. \ref{fig:votes_delta_dirac_prior_ratio_updates}).

\section{Test-time training objectives}\label{app:details_test_time_training_loss}

\subsection{Test-time reinforcement learning (TTRL)}\label{app:subsec_analysis_TTRL}
TTRL leverages majority voting over $n$ responses $X_1, \dots, X_n$ as a proxy for the correct answer, and defines the reward function $r_n(Y_i) = \mathbf{1}\{X_i = \widehat{c}_n\}$, where $\widehat{c}_n$ is the majority vote. 
The regularised objective it minimises is of the form
$$
L(\pi) := -\mathbb{E}_{pr \sim Q}\ \mathbb{E}_{Y\sim \pi(\cdot | pr)}[\mathbf{1}\{X=\widehat{c}_n\}] + \beta\ \KL(\pi(\cdot | pr) || \pi_{\text{ref}}(\cdot | pr)),
$$
where $\pi$ is the candidate distribution and $\pi_{\text{ref}}$ is a pre-trained reference model. Note that $L$ is strictly convex and therefore admits a unique global minimiser.

\paragraph{Optimisation of the regularised objective.} To compute the optimiser, we introduce a Lagrange multiplier $\lambda$ to enforce normalisation and consider a perturbation
$\pi_\varepsilon = \pi + \varepsilon\varphi$ with $\int\varphi\,d\mu = 0$.
The directional derivative at $\varepsilon=0$ is
$$
\left.\frac{d}{d\varepsilon}
\left[ L[\pi_\varepsilon] + \lambda \int \pi_\varepsilon\,d\mu \right]
\right|_{\varepsilon=0}
=
\int_{\Omega} \left[-\delta_{\widehat{c}_n} + \beta\left(1+\log\frac{\pi}{\pi_{\text{ref}}}\right) + \lambda\right]\varphi\,d\mu .
$$
Since this must vanish for all admissible $\varphi$, we obtain the pointwise stationarity condition
$$
-\mathbf{1}\{x=\widehat{c}_n\} + \beta\left(1+\log\frac{\pi(x)}{\pi_{\text{ref}}(x)}\right) + \lambda = 0.
$$
Solving this yields the tilted distribution
\begin{equation*}
\pi^\star(y | pr) \propto e^{\mathbf{1}\{x=\widehat{c}_n\}/\beta}\pi_{\text{ref}}(y | pr) = \left(1 + \mathbf{1}\{x=\widehat{c}_n\}\left(e^{\frac{1}{\beta}}-1\right)\right)\pi_{\text{ref}}(y|pr).
\end{equation*}
As $\beta\rightarrow 0$, the model converges to a Dirac delta centred at $\widehat{c}_n$.   For non-zero regularisation values $\beta$, the solution retains some  structure from the reference model. Assuming $\pi_{\text{ref}}$ is normalised and $e^{1/\beta} > 1$, we can write
$$
\pi^\star(y | pr) = \frac{e^{\mathbf{1}\{x=\widehat{c}_n\}/\beta}\pi_{\text{ref}}(y| pr)}{\pi_{\text{ref}}(\widehat{c}_n|pr)e^{1/\beta} +\sum_{x'\neq \widehat{c}_n} \pi_{\text{ref}}(y'|pr)} =  \frac{e^{\mathbf{1}\{x=\widehat{c}_n\}/\beta}\pi_{\text{ref}}(y | pr)}{1 + \pi_{\text{ref}}(\widehat{c}_n|pr)(e^{1/\beta}-1)}. 
$$

Let $\kappa = 1/\beta$ and $p_j=\pi_{\text{ref}}(j)$. We now analyse the behaviour of $\text{SNR}({\Delta_{j^\star}})$ as a function of $\kappa$. To do so, we compute its derivative with respect to $\kappa$
\begin{align*}
    \frac{d}{d\kappa}\text{SNR}_{\Delta_{j^\star}}(\kappa) =& \frac{d}{d\kappa}\frac{(\pi_{\hat{c}}^\star- \pi^\star_{j^\star})^2}{(\pi^\star_{\hat{c}}+\pi^\star_{j^\star})-(\pi_{\hat{c}}^\star- \pi^\star_{j^\star})^2}\\
    =& \frac{d}{d\kappa}\frac{(p_{\hat{c}}e^\kappa - p_{j^\star})^2}{(p_{\hat{c}}e^\kappa+p_{j^\star})(1+(e^\kappa-1)p_{\hat{c}})-(p_{\hat{c}}e^\kappa - p_{j^\star})^2}\\
    =& \frac{2(p_{\hat{c}}e^\kappa - p_{j^\star})p_{\hat{c}}e^\kappa\left[(p_{\hat{c}}e^\kappa+p_{j^\star})(1+(e^\kappa-1)p_{\hat{c}})-(p_{\hat{c}}e^\kappa - p_{j^\star})^2\right]}{((p_{\hat{c}}e^\kappa+p_{j^\star})(1+(e^\kappa-1)p_{\hat{c}})-(p_{\hat{c}}e^\kappa - p_{j^\star})^2)^2} \\
    &- \frac{(p_{\hat{c}}e^\kappa - p_{j^\star})^2p_{\hat{c}}e^\kappa\left[(1+(e^\kappa-1)p_{\hat{c}}) + (p_{\hat{c}}e^\kappa+p_{j^\star})\right]}{\left((p_{\hat{c}}e^\kappa+p_{j^\star})(1+(e^\kappa-1)p_{\hat{c}})-(p_{\hat{c}}e^\kappa - p_{j^\star})^2\right)^2}\\
    &+ \frac{2(p_{\hat{c}}e^\kappa-p_{j^\star})p_{\hat{c}}e^\kappa(p_{\hat{c}}e^\kappa - p_{j^\star})^2}{\left((p_{\hat{c}}e^\kappa+p_{j^\star})(1+(e^\kappa-1)p_{\hat{c}})-(p_{\hat{c}}e^\kappa - p_{j^\star})^2\right)^2}\\
    =& \frac{(\square)}{((p_{\hat{c}}e^\kappa+p_{j^\star})(1+(e^\kappa-1)p_{\hat{c}})-(p_{\hat{c}}e^\kappa - p_{j^\star})^2)^2}.
\end{align*}
The denominator is clearly positive, so we focus on the numerator. After cancelling out common terms, the numerator reduces to
\begin{align*}
       (\square) =& (p_{\hat{c}}e^\kappa - p_{j^\star})p_{\hat{c}}e^\kappa\Big[2(p_{\hat{c}}e^\kappa+p_{j^\star})(1+(e^\kappa-1)p_{\hat{c}}) - (1+(e^\kappa-1)p_{\hat{c}})(p_{\hat{c}}e^\kappa-p_{j^\star}) \\
       &- (p_{\hat{c}}e^\kappa+p_{j^\star})(p_{\hat{c}}e^\kappa-p_{j^\star})\Big]\\
       =& (p_{\hat{c}}e^\kappa - p_{j^\star})p_{\hat{c}}e^\kappa\Big[2p_{j^\star}(1+(e^\kappa-1)p_{\hat{c}})+(p_{\hat{c}}e^\kappa+p_{j^\star})(1-p_{\hat{c}}+p_{j^\star})\Big].
\end{align*}
Since $\kappa\geq 0$ and $0\leq p_{j^\star}\leq p_{\hat{c}}\leq 1$, it follows that $(\square)\geq 0$, with equality if and only if $p_{\hat{c}}=1$. Therefore, for $0<p_{\hat{c}}<1$, $ \frac{d}{d\kappa}\text{SNR}_{\Delta_{j^\star}}(\kappa)>0$, which implies that  $\text{SNR}_{\Delta_{j^\star}}(\kappa)$ is an increasing function of $\kappa$. This demonstrates that optimising the TTRL objective reduces the number of samples required to achieve statistical certificates. 

\subsection{SNR-based test-time RL objective}\label{app:details_new_TTT_loss}
Let $\mathbf{X} = (X_1, \dots, X_n)$ be a collection of answers to a given prompt corresponding to rollouts $\mathbf{Y} = (Y_1, \dots, Y_n)$, with $\widehat{c}_n$ denoting the majority vote and $j_n^\star$  the runner-up. We propose to directly maximise $\text{SNR}(\Delta_{j^\star_n})$ by using the group-level reward function $r_n^{(1)}$ defined in Eq. (\ref{eq:snr_based_reward}).
Our objective (without the KL-regularisation) takes the form
\begin{align*}
&\max_\phi \mathbb{E}_{Y_1, \dots, Y_n\sim \pi_\phi(\cdot|pr)}\left[r_n^{(1)}(\mathbf{Y})\right] = \max_\phi \mathbb{E}_{Y_1, \dots, Y_n\sim \pi_\phi(\cdot|pr)}\left[\reallywidehat{\text{SNR}}(\Delta_{j_n^\star})(\mathbf X)\right]\\
&= \max_\phi \mathbb{E}_{Y_1, \dots, Y_n\sim \pi_\phi(\cdot|pr)} \left[\frac{(N_{\hat{c}_n}+N_{{j}_n^\star})^2}{n(N_{\hat{c}_n}-N_{{j}_n^\star}) - (N_{\hat{c}_n}-N_{{j}_n^\star})^2}\right],
\end{align*}
where $N_j= \sum_i\mathbf{1}\{X_i=j\}$.
It is important to note that $\reallywidehat{\text{SNR}}(\Delta_{j_n^\star})(\mathbf X)$ is a biased estimator of ${\text{SNR}}(\Delta_{j_n^\star})$, however in the large-sample limit we obtain the approximation
\begin{align*}
&\max_\phi \mathbb{E}_{Y_1, \dots, Y_n\sim \pi_\phi(\cdot|pr)}\left[r_n^{(1)}(\mathbf{Y})\right] = \max_\phi \mathbb{E}_{Y_1, \dots, Y_n\sim \pi_\phi(\cdot|pr)}\left[\reallywidehat{\text{SNR}}(\Delta_{j_n^\star})(\mathbf X)\right]\\
&\approx \max_\phi \frac{(q_{\hat{c}_n}-q_{j^\star_n})^2}{q_{\hat{c}_n}+q_{j^\star_n}-(q_{\hat{c}_n}-q_{j^\star_n})^2} = \max_\phi {\text{SNR}}(\Delta_{j_n^\star}).
\end{align*}

As discussed in the main text, to reduce the variance of the gradient estimate of the group-level reward, we adopt a leave one-out control variate approach \citep{tang2025optimizing}, resulting in the following effective advantage function for $Y_i$  when using the REINFORCE algorithm \citep{reinforce_92}
\begin{equation}\label{eq:effective_advantage_reinforce}
A_i = \reallywidehat{\text{SNR}}(\Delta_{j_n^\star})(\mathbf X) - \reallywidehat{\text{SNR}}(\Delta_{j_n^\star})(\mathbf X_{-i}). 
\end{equation}

Under the GRPO algorithm \citep{shao2024deepseekmathpushinglimitsmathematical}, the effective advantage for $Y_i$ becomes
\begin{equation}\label{eq:effective_advantage_GRPO}
\hat A_i = \reallywidehat{\text{SNR}}(\Delta_{j_n^\star})(\mathbf X) - \reallywidehat{\text{SNR}}(\Delta_{j_n^\star})(\mathbf X_{-i}) - \frac{1}{n}\sum_i A_i, 
\end{equation}
which further reduces the variance of the gradient estimate at the expense of introducing some bias \citep{tang2025optimizing}.
In addition, we regularise the objective with a KL term that penalises deviations from a reference model $\pi_{\text{ref}}$. 

\paragraph{Optimisation of the regularised objective.} Let $\pi_{\text{ref}} = (p_1, \dots, p_k)$.  We optimise over categorical distributions $\pi = (q_1, \dots, q_k)$.
To enforce the normalisation constraint $\sum_i q_i=1$, we introduce a Lagrange multiplier $\lambda$, yielding the Lagrangian
$$
L(\pi, \lambda) = -\frac{(q_{\hat{c}_n}-q_{j^\star_n})^2}{q_{\hat{c}_n}+q_{j^\star_n}-(q_{\hat{c}_n}-q_{j^\star_n})^2} + \beta \sum_i q_i \log \frac{q_i}{p_i} + \lambda \left( \sum_i q_i - 1 \right),
$$
under the large-sample limit approximation described above.

The stationary points satisfy 
$$
\frac{\partial {L}}{\partial q_i} = 0, \quad \forall i.
$$
Define
$$
g(x,y) = -\frac{(x - y)^2}{x + y - (x - y)^2}
$$
and denote by $g_x$, $g_y$ its partial derivatives with respect to $x$ and $y$, respectively.
The optimality conditions are given by
$$
g_x(q_{\hat{c}_n}, q_{{j}_n^\star}) + \beta \left( 1 + \log \frac{q_{\hat{c}_n}}{p_{\hat{c}_n}} \right) + \lambda = 0,
$$
$$
g_y(q_{\hat{c}_n}, q_{{j}_n^\star}) + \beta \left( 1 + \log \frac{q_{{j}_n^\star}}{p_{j_n^\star}} \right) + \lambda = 0,
$$
$$
\beta \left( 1 + \log \frac{q_{i}}{p_{i}} \right) + \lambda = 0\Longrightarrow q_i\propto p_{i},\ \ \ i\neq \hat{c}_n, j^\star_n.
$$
In general, these equations do not admit a closed-form solution and must be solved numerically.

\subsection{Entropy-based test-time RL objective}\label{app:subsec_analysis_TTTEntropy}
Let $\mathbf{X} = (X_1, \dots, X_n)$ denote the set of i.i.d. answers to a given prompt corresponding to rollouts  $\mathbf{Y} = (Y_1, \dots, Y_n)$. Define $N_j = \sum_i \mathbf{1}\{X_i = j\}$. In the main text, we proposed a group-level reward function based on the plug-in estimator of the negative entropy
\begin{equation*}
    r_n^{(2)}(\mathbf{Y})=\sum_{j:\ N_j>0}\frac{N_j}{n}\log\left(\frac{N_j}{n}\right).
\end{equation*}
This estimator is known to overestimate $\mathbb{E}[\log X]$, with an error of approximately $(k-1)/(2n)$, where $k$ is the total number of classes of the distribution \citep{miller_1995}. An alternative approach is to introduce a Dirichlet prior on the class probabilities, $(p_1, \dots, p_k)\sim\text{Dir}(k, \alpha, \dots, \alpha)$.
Since the data are multinomial, the posterior distribution of the probabilities is also Dirichlet. After $n$ observations we obtain
$$
(p_1, \dots, p_k)|\mathbf{Y},\alpha\sim \text{Dir}(k, \alpha + N_1, \dots, \alpha+N_k)
$$
This leads to the alternative estimator
\begin{equation*}
    \hat r_n^{(2)}(\mathbf{Y}) = \sum_{j}\frac{N_j+\alpha}{n+\alpha}\log\left(\frac{N_j+\alpha}{n+\alpha}\right).
\end{equation*}
Because our ensembles of voters are typically small, this Bayesian smoothing can help mitigate fluctuations, especially when prior information is available.
Alternative estimators have been proposed in \cite{NIPS2013_53c04118}.

By using the reward functions $r_n^{(2)}(\mathbf{Y})$ or $\hat r_n^{(2)}(\mathbf{Y})$, the goal is to minimise the entropy of the answer distribution. 
In particular, our objective (without regularisation) is
\begin{align*}
&\max_\phi \mathbb{E}_{Y_1, \dots, Y_n\sim \pi_\phi(\cdot|pr)}\left[r_n^{(2)}(\mathbf{Y})\right]= \max_\phi \mathbb{E}_{Y_1, \dots, Y_n\sim \pi_\phi(\cdot|pr)}\left[\sum_{j:\ N_j>0}\frac{N_j}{n}\log\left(\frac{N_j}{n}\right)\right].
\end{align*}
As in the previous section, $r_n^{(2)}(\mathbf{Y})$ is a biased estimator of the negative entropy $\mathbb{E}[\log X]$.  However, in the large-sample limit we obtain the approximation
\begin{align*}
&\max_\phi \mathbb{E}_{Y_1, \dots, Y_n\sim \pi_\phi(\cdot|pr)}\left[r_n^{(2)}(\mathbf{Y})\right]\approx \max_\phi \sum_{j:\ p_{j,\phi}>0} p_{j,\phi}\log p_{j,\phi}= \max_\phi \mathbb{E}_{Y\sim  \pi_{\phi}(\cdot|pr)}[\log X].
\end{align*}
To reduce the variance of the gradient estimates of the group-level reward, we also employ the effective advantage functions introduced in (\ref{eq:effective_advantage_reinforce}) and (\ref{eq:effective_advantage_GRPO}), for the REINFORCE and GRPO algorithms, respectively.

\paragraph{Optimisation of the regularised objective.} 
Let $\pi_{\mathrm{ref}}(Y_{0:\tau})$ denote the reference distribution over reasoning trajectories with terminal variable
$X=g(Y_{\tau:})$, and write $p_{\mathrm{ref}}(x)=\pi_{\mathrm{ref}}(X=x)$ for its induced marginal.
As mentioned in the main text, the KL-regularised variational problem over the base measure reduces to one over the marginal $q(x) = \pi_\phi(x)$ alone, with the following loss
\begin{align*}
    L(q) &=  H( q) + \beta\ \KL(q || p_{\text{ref}})\\
    &=  \beta\left(1/\beta\ H( q) + \beta\ \KL(q || p_{\text{ref}})\right)\\
    &\propto \mathbb{E}_{pr \sim Q}\ \mathbb{E}_{X\sim q(\cdot | pr)}\left[-1/\beta \log q(X | pr)\right] + \KL(q(\cdot|pr) || p_{\text{ref}}(\cdot | pr))\\
    & =\mathbb{E}_{pr \sim Q}\ \mathbb{E}_{X\sim q(\cdot | pr)}\left[(1-1/\beta) \log q(X | pr)  -\log p_{\text{ref}}(X | pr))\right],
\end{align*}
where $\beta>1$. 
Since the mapping $q\mapsto (1-1/\beta)\int q\log q$ is {strictly} convex, and the second
term is linear, it follows that $L$ is strictly convex on the space of probability distributions. Consequently, any stationary point is necessarily the unique global minimiser.

As in Section \ref{app:subsec_analysis_TTRL}, to compute the optimiser we introduce a Lagrange multiplier $\lambda$ to enforce normalisation and consider a perturbation
$q_\varepsilon = q + \varepsilon\varphi$ with $\int\varphi\,d\mu = 0$.
The directional derivative at $\varepsilon=0$ is
$$
\left.\frac{d}{d\varepsilon}
\left[ L[q_\varepsilon] + \lambda\!\int q_\varepsilon\,d\mu \right]
\right|_{\varepsilon=0}
=
\int_{\Omega}\left[(1-1/\beta)(1+\log q) - \log p_{\text{ref}} + \lambda\right]\ \varphi\,d\mu .
$$
Since this must vanish for all admissible $\varphi$, the pointwise stationarity condition is
$$
(1-1/\beta)\bigl(1+\log q(x)\bigr) - \log p_{\text{ref}}(x) + \lambda = 0.
$$
Solving for $q$ yields
$$
\log q(x)
=
\frac{\log p_{\text{ref}}(x) - \lambda - (1-1/\beta)}{1-1/\beta}
\;=\;
\kappa\log p_{\text{ref}}(x) + C,
$$
where $\kappa = \beta/(\beta-1)>1$ and 
$C=\bigl[-\lambda - (1-1/\beta)\bigr]/(1-1/\beta)$ is a constant.
Exponentiating and renormalising gives the tempered distribution
$$
q(x)=\frac{e^{C} p_{\text{ref}}(x)^{\kappa}}{\int_{\Omega}e^{C} p_{\text{ref}}(x)^{\kappa}\,d\mu}
     \;=\;\frac{ p_{\text{ref}}(x)^{\kappa}}{Z_\beta},
$$
with $Z_\beta$ the normalisation constant. 

Let $p_j=p_{\text{ref}}(j)$. Under the optimal distribution $q^\star$, the signal-to-noise ratio $\text{SNR}_{\Delta_{j^\star}}$ takes the form
\begin{align*}
    \text{SNR}_{\Delta_{j^\star}}(\kappa) =&\frac{(q_{\hat{c}}^\star- q^\star_{j^\star})^2}{(q^\star_{\hat{c}}+q^\star_{j^\star})-(q_{\hat{c}}^\star- q^\star_{j^\star})^2}
    = \frac{(p_{\hat{c}}^\kappa - p_{j^\star}^\kappa)^2}{(p_{\hat{c}}^\kappa+p_{j^\star}^\kappa)\sum_i p_{i}^\kappa -(p_{\hat{c}}^\kappa - p_{j^\star}^\kappa)^2} \\=& \frac{(p_{\hat{c}}^\kappa - p_{j^\star}^\kappa)^2}{4p_{\hat{c}}^\kappa p_{j^\star}^\kappa + (p_{\hat{c}}^\kappa + p_{j^\star}^\kappa)\sum_{i\neq \hat{c}, j^\star} p_{i}^\kappa}
    = \frac{\left(\left(\tfrac{p_{\hat{c}}}{p_{j^\star}}\right)^\kappa-1\right)^2}{4\left(\tfrac{p_{\hat{c}}}{p_{j^\star}}\right)^\kappa + \left(\left(\tfrac{p_{\hat{c}}}{p_{j^\star}}\right)^\kappa+ 1\right)\sum_{i\neq \hat{c}, j^\star} \left(\tfrac{p_{i}}{p_{j^\star}}\right)^\kappa}.
\end{align*}
To study the behaviour of $\text{SNR}_{\Delta_{j^\star}}$ as a function of $\kappa$, we calculate its derivative with respect to $\kappa$. 
To do so define 
$$s(\kappa) = \left(\frac{p_{\hat{c}}}{p_{j^\star}}\right)^\kappa\geq 1\quad \text{and} \quad r(\kappa) = \sum_{i\neq \hat{c}, j^\star} \left(\frac{p_{i}}{p_{j^\star}}\right)^\kappa\geq 0.$$ Differentiating $\text{SNR}_{\Delta_{j^\star}}(\kappa)$ with respect to $\kappa$ gives

\begin{align*}
    \frac{d}{d\kappa}\text{SNR}_{\Delta_{j^\star}}(\kappa) =& s'(\kappa)\frac{2(s(\kappa)-1)\left(4s(\kappa) + \left(s(\kappa)+ 1\right)r(\kappa)\right)-(s(\kappa)-1)^2(4+r(\kappa))}{\left(4s(\kappa) + \left(s(\kappa)+ 1\right)r(\kappa)\right)^2}\\
    &- r'(\kappa)\frac{(s(\kappa)-1)^2(s(\kappa)+1)}{\left(4s(\kappa) + \left(s(\kappa)+ 1\right)r(\kappa)\right)^2}\\
    =& s'(\kappa)(s(\kappa)-1)\frac{4s(\kappa) + 3r(\kappa)+ s(\kappa)r(\kappa)+4}{\left(4s(\kappa) + \left(s(\kappa)+ 1\right)r(\kappa)\right)^2}\\
    &- r'(\kappa)\frac{(s(\kappa)-1)^2(s(\kappa)+1)}{\left(4s(\kappa) + \left(s(\kappa)+ 1\right)r(\kappa)\right)^2}\\
\end{align*}
Since $s(\kappa)-1 \geq 0$ and $r(\kappa)\geq 0$, it is sufficient to show that $s'(\kappa)\geq 0$ and $r'(\kappa)\leq 0$ in order to conclude that $\frac{d}{d\kappa}\text{SNR}_{\Delta_{j^\star}}(\kappa)\geq 0$.
Indeed,
\begin{align*}
       s'(\kappa) =  \left(\frac{p_{\hat{c}}}{p_{j^\star}}\right)^\kappa \ln  \left(\frac{p_{\hat{c}}}{p_{j^\star}}\right) \geq 0
\end{align*}
and 
\begin{align*}
       r'(\kappa) =  \sum_{i\neq \hat{c}, j^\star}\left(\frac{p_{i}}{p_{j^\star}}\right)^\kappa \ln  \left(\frac{p_{i}}{p_{j^\star}}\right) \leq 0,
\end{align*}
since $p_i\leq p_{j^\star}$ for $i\neq \hat{c}, j^\star$.

This implies that $\text{SNR}_{\Delta_{j^\star}}(\kappa)$ is non-decreasing for $\kappa\geq1$, showing that entropy-penalising rewards reduce the number of samples required for certification.

\paragraph{Differences from existing entropy-penalising methods.}
We highlight how our approach differs from that of \citep{agarwal2025unreasonableeffectivenessentropyminimization}.
Their method minimises individual rewards for a trajectory $(Y_t)_{t\geq 0}$, corresponding to an answer $X = g(Y_{\tau:})$ where $\tau$ is a random stopping time. Specifically, they define two entropy-based reward functions
\begin{itemize}
    \item Negative trajectory-level entropy estimator. The reward for a full trajectory $(Y_t)_{t\geq 0}$ is
    $$
    {r}_{\text{traj}}(Y_t)=\sum_{t=1}^{|Y^i_t|}\log\pi(Y_t^i|Y_{<t}^i).
    $$
    \item  Negative token level entropy. In this case, the reward is of the form
    $$
    r_{\text{tok}}(Y_t) =\sum_{t=1}^{|Y^i_t|}\sum_{j\in\mathcal{V}}\pi(j|Y_{<t}^i)\log\pi(j|Y_{<t}^i),
    $$
    where $\mathcal{V}$ denotes the vocabulary.
\end{itemize}
While both trajectory-level and token-level rewards aim to minimise entropy, they influence RL training differently: minimising trajectory entropy encourages policies with lower entropy over entire trajectories, whereas minimising token-level entropy encourages policies with low entropy at each generation step. 
In contrast, our group-level reward function targets the entropy of the final answer distribution, directly improving the model’s confidence in its final output while allowing exploration of diverse pathways during the chain-of-thought reasoning process.

\section{Experimental details}\label{app:numerical_experiments_details}
\subsection{Experimental setup}
We adopt the data and evaluation pipeline from the TTRL codebase \citep{zuo2025ttrl}, which is built on the VERL framework \citep{sheng2024hybridflow}.
The final answer of the language model is the string inside the last \emph{\textbackslash boxed\{\}}.

\paragraph{Implementation details.} 
We use hyperparameters similar to those in TTRL \citep{zuo2025ttrl} and report them here for completeness. 
A cosine learning rate schedule is applied, with a peak value of $5 \times 10^{-7}$, and the AdamW optimiser is used for the policy model with a learning rate of $9 \times 10^{-6}$. 
The KL-regularisation parameter in the RL objective is set to $0.001$.

We sample 64 responses per prompt using a temperature of $0.6$ ($1.0$ for Qwen2.5-Math models) for voting-based label estimation, and downsample $32$ responses per prompt for training. The maximum generation length is fixed at $3072$ tokens for all models. The number of episodes is set to $80$, $30$, and $10$ for AIME 2024, AMC, and MATH-500, respectively, reflecting dataset size. We also apply early stopping with a tolerance of $5\times 10^{-3}$ and a patience of $10$ iterations, evaluated on both metrics (pass@1 and majority).

All other hyperparameters not explicitly mentioned here are set to their default values in the VERL framework.
For the TTRL \citep{zuo2025ttrl} baseline, we adopt the hyperparameters reported in the paper.

\paragraph{Evaluation details.}
We also set the maximum generation length to $3072$ tokens during evaluation.
Following \citet{zuo2025ttrl}, we report the pass@1 score using non-zero temperature
sampling. Specifically, for each prompt $pr$, we generate $N=16$ responses using a
temperature of $0.6$ and a top-p value of $0.95$. 
The pass@1 score is then computed as
$$
\text{pass@1}= \frac{1}{QN}\sum_{pr}\sum_{i=1}^N \mathbf{1}\{X_i(pr)=\text{correct}\},
$$
where $X_i(pr)$ denotes the $i$-th generated response for prompt $pr$ and $Q$ is the total number of prompts.

We also report majority vote accuracy, which indicates whether the most frequent answer among the $N = 16$ responses per prompt matches the ground truth
$$
\text{majority}= \frac{1}{Q}\sum_{pr} \mathbf{1}\{\text{majority vote}(X_1(pr), \dots, X_N(pr))=\text{correct}\}.
$$

\paragraph{Computation time.}
All experiments were conducted on 8$\times$H100 Nvidia GPUs, each with 96GB of memory.

\subsection{Additional results}\label{app:subsec_additional_results}
Table~\ref{tab:test-time-training-results-pass1-format-score} expands upon the results presented in Table~\ref{tab:test-time-training-results}. 
It reports the pass@1 performance for both the score and the format score before and after applying test-time training with our proposed reward functions. 
We observe that, for Qwen2.5 models, the improvement in score is notably larger than that in format score, suggesting that test-time training effectively uncovers latent knowledge already present in the model rather than merely correcting format errors.
In contrast, for the Llama-3.1-8B model, we hypothesise that the mode of the model's final answer distribution does not coincide with the true answer, therefore, test-time training incorrectly shifts the model's output distribution. That is, the model lacks the necessary mathematical knowledge, and our test-time training strategies serve to reveal rather than create new knowledge.
Table~\ref{tab:test-time-training-results-majority-format-score} presents analogous results for majority vote accuracy, leading to similar conclusions.

\begin{table}[b!]
\caption{Comparison of pass@1 performance for the score and format score (using 16 samples per prompt) before and after applying test-time training.}
\label{tab:test-time-training-results-pass1-format-score}
\begin{center}
\footnotesize
\begin{tabular}{lcccccc}
\toprule
     & \multicolumn{2}{c}{\textbf{AIME}} & \multicolumn{2}{c}{\textbf{AMC}} & \multicolumn{2}{c}{\textbf{Math\scriptsize-500\footnotesize}} \\
\midrule
     & \textbf{Score} & \textbf{Format score}  & \textbf{Score} & \textbf{Format score}  & \textbf{Score} & \textbf{Format score}\\
\midrule
\textbf{Qwen2.5-7B} &9.4&  84.6 &31.2& 84.6 &59.1& 90.2\\
\\
\multirow{2}{*}{SNR}  &23.3& 100.0 &51.8& 99.5 &80.3 &98.9 \\
  &\textcolor{green}{+13.9}  & \textcolor{green}{+15.4}   &\textcolor{green}{+20.6}& \textcolor{green}{+14.9}&\textcolor{green}{+21.2}& \textcolor{green}{+8.7} \\
 \\
\multirow{2}{*}{Entropy}  &20.0& 100.0 &49.2& 99.5 &77.6& 100.0\\
 &\textcolor{green}{+10.6}& \textcolor{green}{+15.4} &\textcolor{green}{+18.0}& \textcolor{green}{+14.9} &\textcolor{green}{+18.5}& \textcolor{green}{+9.8} \\
\midrule
\textbf{Llama-3.1-8B} & 4.4& 60.0 & 21.8 & 72.0 &48.2& 83.8\\
\\
\multirow{2}{*}{SNR}  &13.4& 99.6 &29.3& 100.0 &59.2& 100.0 \\
 &\textcolor{green}{+9.0}& \textcolor{green}{+39.6} &\textcolor{green}{+7.5}& \textcolor{green}{+28.0}&\textcolor{green}{+11.0}& \textcolor{green}{+16.2} \\
 \\
\multirow{2}{*}{Entropy}    &13.3& 99.8 &27.0& 100.0 &55.4& 100.0\\
  &\textcolor{green}{+8.9}& \textcolor{green}{+39.8}&\textcolor{green}{+5.2}&  \textcolor{green}{+28.0} &\textcolor{green}{+7.2}& \textcolor{green}{+16.2}\\
\midrule
\textbf{Qwen2.5-Math-7B}  &10.6&  73.5 &31.0& 85.4 &47.1& 90.2\\
\\
\multirow{2}{*}{SNR}   & 36.7& 85.4 &65.0& 88.8 &84.5& 97.5 \\
  &\textcolor{green}{+26.1}& \textcolor{green}{+11.9}   &\textcolor{green}{+34.0}& \textcolor{green}{+3.4} &\textcolor{green}{+37.4}& \textcolor{green}{+7.3}   \\
 \\
\multirow{2}{*}{Entropy}    &38.3& 97.5 &65.4& 99.9  &82.4 & 99.3\\
  &\textcolor{green}{+27.7}& \textcolor{green}{+24.0} &\textcolor{green}{+34.4}&  \textcolor{green}{+14.5} &\textcolor{green}{+35.3}&  \textcolor{green}{+9.1}\\
\midrule
\textbf{Qwen2.5-Math-1.5B}  & 7.1& 74.2 & 28.1 & 80.1 &31.4 & 66.4\\
\\
\multirow{2}{*}{SNR}  &16.3 & 91.9& 45.4& 92.2& 72.0 & 97.7\\
 &\textcolor{green}{+9.2}& \textcolor{green}{+17.7} &\textcolor{green}{+17.3}& \textcolor{green}{+12.1} &\textcolor{green}{+40.6}& \textcolor{green}{+11.3} \\
 \\
\multirow{2}{*}{Entropy} &15.6  & 88.3 &45.9& 96.2 &70.8& 98.1\\
  &\textcolor{green}{+8.5}& \textcolor{green}{+14.1}&\textcolor{green}{+17.8}& \textcolor{green}{+16.1} &\textcolor{green}{+39.4}& \textcolor{green}{+11.7}\\
\bottomrule
\end{tabular}
\end{center}
\end{table}
\normalsize

\begin{table}[ht!]
\caption{Comparison of majority vote accuracy for the  score and format score (using 16 samples per prompt) before and after applying test-time training.}
\label{tab:test-time-training-results-majority-format-score}
\begin{center}
\footnotesize
\begin{tabular}{lcccccc}
\toprule
     & \multicolumn{2}{c}{\textbf{AIME}} & \multicolumn{2}{c}{\textbf{AMC}} & \multicolumn{2}{c}{\textbf{Math\scriptsize-500\footnotesize}} \\
\midrule
     & \textbf{Score} & \textbf{Format score}  & \textbf{Score} & \textbf{Format score}  & \textbf{Score} & \textbf{Format score}\\
\midrule
\textbf{Qwen2.5-7B} & 16.7 & 72.8 &41.8  & 79.6 &73.5 &  93.4\\
\\
\multirow{2}{*}{SNR} &23.3& 100.0 &51.2&100.0 &81.0& 98.9\\
 &\textcolor{green}{+6.6} & \textcolor{green}{+27.2}  &\textcolor{green}{+9.4} & \textcolor{green}{+20.4}&\textcolor{green}{+7.5} & \textcolor{green}{+5.5} \\
 \\
\multirow{2}{*}{Entropy} &20.0& 100.0 &49.4&100.0 &79.0& 100.0\\
 & \textcolor{green}{+3.3}& \textcolor{green}{+27.2}   &\textcolor{green}{+7.6} & \textcolor{green}{+20.4}&\textcolor{green}{+5.5} & \textcolor{green}{+6.6}\\
\midrule
\textbf{Llama-3.1-8B} &4.6& 26.8 &27.4& 53.4 &57.7 & 77.2\\
\\
\multirow{2}{*}{SNR}  &13.3 & 99.8 &28.6& 100.0&60.3& 100.0\\
& \textcolor{green}{+8.7} & \textcolor{green}{+73.0}  &\textcolor{green}{+1.2}& \textcolor{green}{+46.6} &\textcolor{green}{+2.6} & \textcolor{green}{+22.8}\\
 \\
\multirow{2}{*}{Entropy}   & 13.3 & 100.0 & 29.3 & 100.0&57.6& 100.0\\
&\textcolor{green}{+8.7}& \textcolor{green}{+73.2} &\textcolor{green}{+1.9}& \textcolor{green}{+46.6} & \textcolor{red}{-0.1} & \textcolor{green}{+22.8}\\
\midrule
\textbf{Qwen2.5-Math-7B} &16.5& 56.3 &41.5 & 78.4 &59.5& 87.8\\
\\
\multirow{2}{*}{SNR}  &37.8& 77.9 &67.2 & 87.9 &85.7& 99.5\\
&\textcolor{green}{+21.3} & \textcolor{green}{+21.6} &\textcolor{green}{+25.7}& \textcolor{green}{+9.5}&\textcolor{green}{+26.2} & \textcolor{green}{+11.7} \\
 \\
\multirow{2}{*}{Entropy}   &36.7& 97.3 & 66.1& 100.0&84.3 & 99.5\\
&\textcolor{green}{+20.2} & \textcolor{green}{+41.0}  &\textcolor{green}{+24.6} & \textcolor{green}{+22.6} &\textcolor{green}{+24.8} & \textcolor{green}{+11.7} \\
\midrule
\textbf{Qwen2.5-Math-1.5B} &11.7& 60.0 &37.2&  70.8 &36.5 & 57.4\\
\\
\multirow{2}{*}{SNR}  &23.7 & 90.5 &53.3&  90.9&78.9 & 97.8\\
& \textcolor{green}{+12.0} &\textcolor{green}{+30.5} & \textcolor{green}{+16.1}&   \textcolor{green}{+20.1}&\textcolor{green}{+42.4}& \textcolor{green}{+40.4}\\
 \\
\multirow{2}{*}{Entropy}  &23.2  &  81.3 &52.8& 95.7 &77.4 & 98.2\\
 &\textcolor{green}{+11.5} & \textcolor{green}{+21.3} &\textcolor{green}{+15.6}&   \textcolor{green}{+24.9} &\textcolor{green}{+40.9} & \textcolor{green}{+40.8}\\
\bottomrule
\end{tabular}
\end{center}
\end{table}
\normalsize

Table~\ref{tab:ratio_adaptive_sampling_pre_post_trained} provides evidence that the model becomes more confident in its outputs after applying test-time training strategies.
Specifically, the required number of samples for the MMC stopping rule, denoted as $N_{\text{adaptive}}$ is lower after test-time training compared to the pre-trained model.

The relationship between $N_{\text{adaptive}}$ and $N_{\text{budget}}$ can be accurately modelled by a linear regression of the form $N_{\text{adaptive}} = \alpha + \beta N_{\text{budget}}$ with a coefficient of determination $R^2$ very close to 1. We therefore report the estimated value of $\beta$ obtained via least squares fitting.
Since $0\leq N_{\text{adaptive}}\leq N_{\text{budget}}$, it follows that $\beta\leq1$. 

We observe that the estimated slope for the pre-trained model, $\hat\beta_{\text{pre}}$, is larger than that of the test-time trained model, $\hat\beta_{\text{post}}$. This reduction is particularly pronounced for the smaller 1.5B model, suggesting that larger models experience diminishing returns from test-time training.

\begin{figure}[h!]
  \centering
  \begin{subfigure}{0.98\textwidth}
      \centering
      \includegraphics[width=\textwidth]{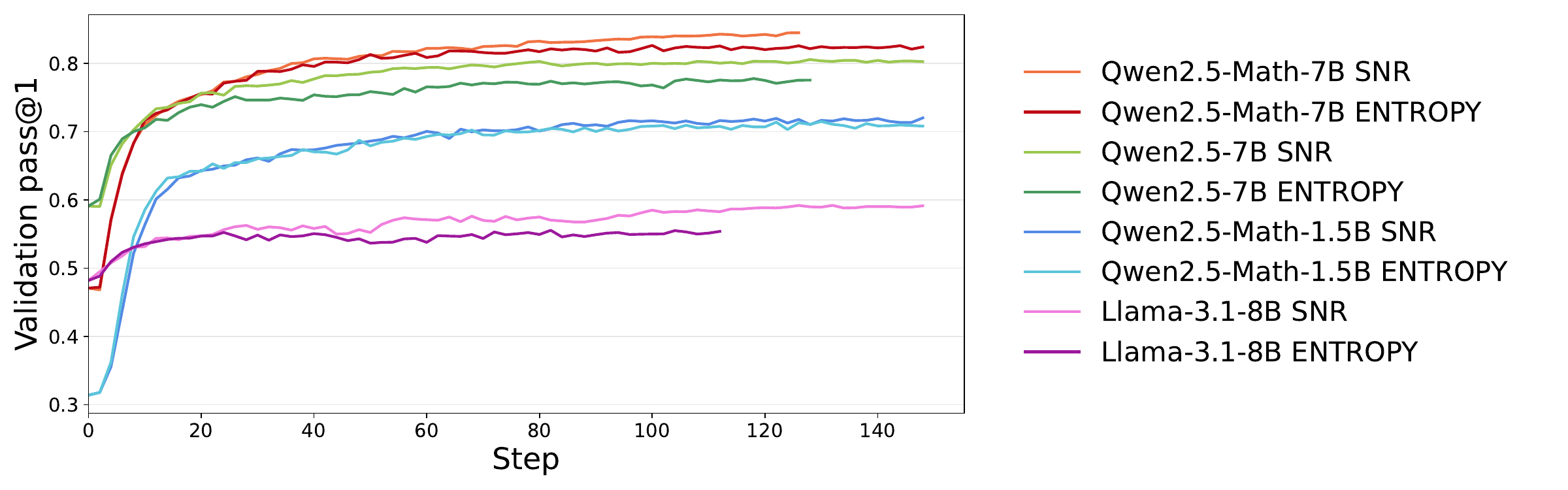}
      \label{fig:evolution_validation_pass@1}
  \end{subfigure}
  \vfill
  \begin{subfigure}{0.49\textwidth}
      \centering
      \includegraphics[width=\textwidth]{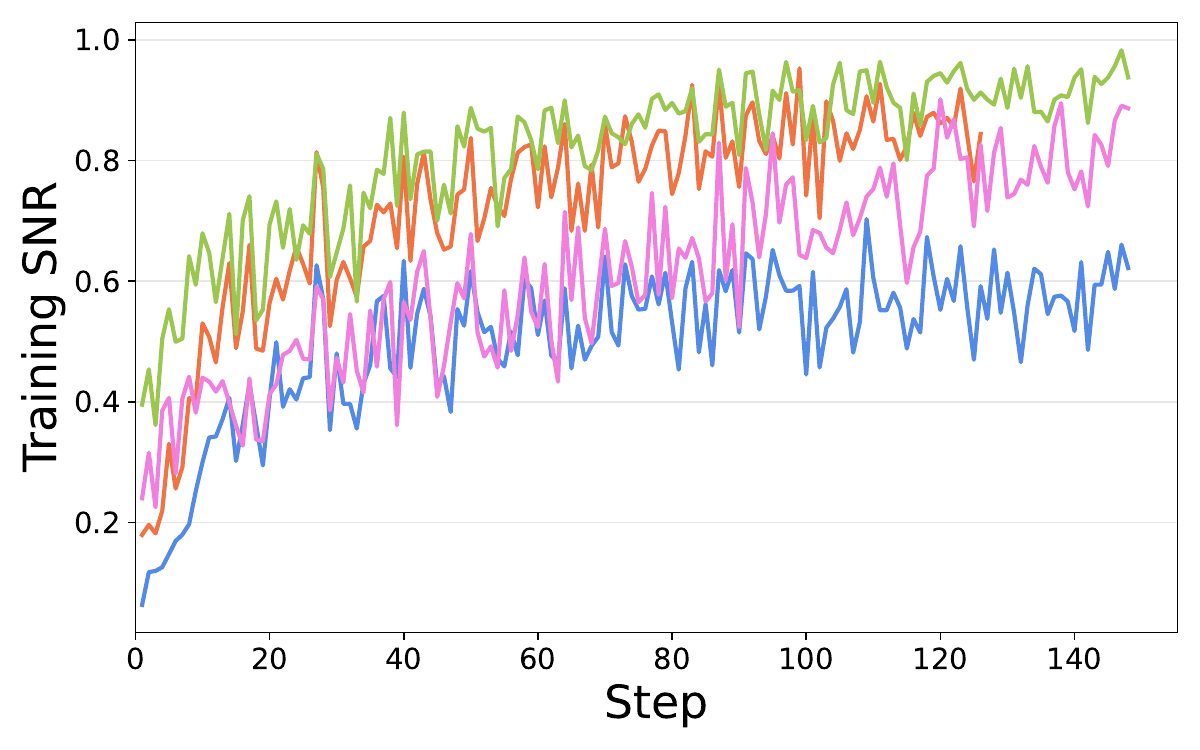}
      \label{fig:evolution_training_snr}
  \end{subfigure}
  \hfill
  \begin{subfigure}{0.49\textwidth}
      \centering
      \includegraphics[width=\textwidth]{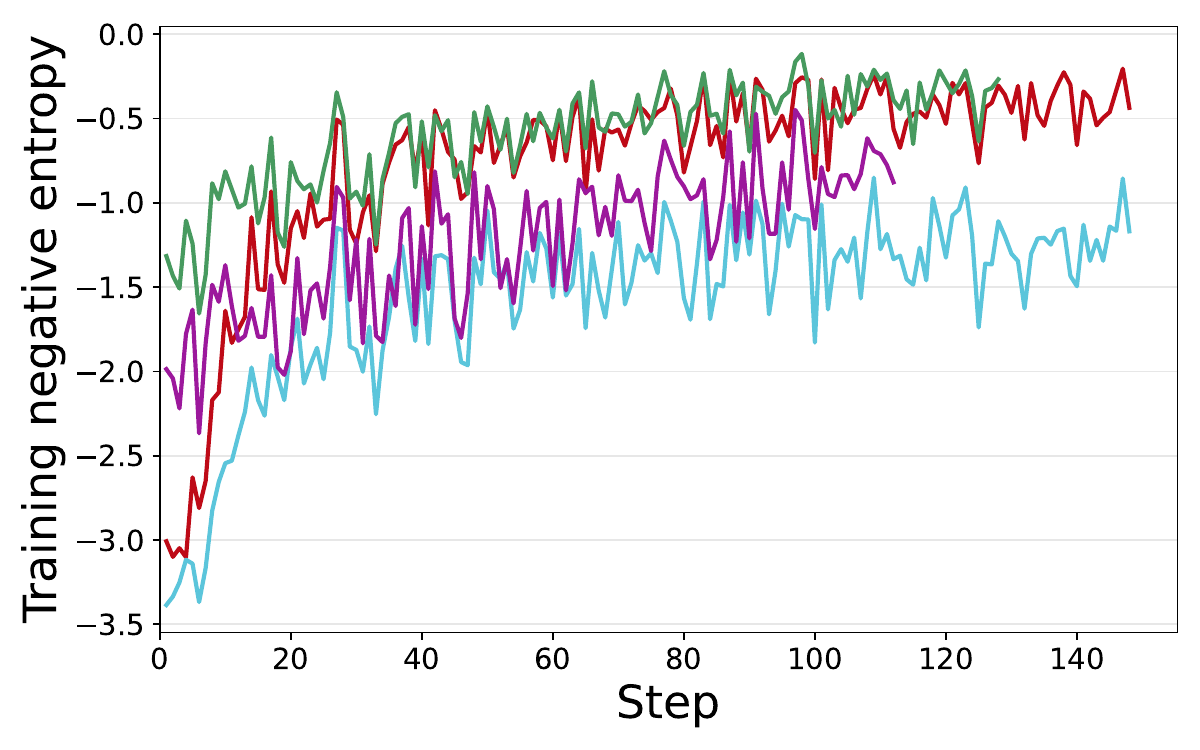}
      \label{fig:evolution_training_negative_entropy}
  \end{subfigure}
  \caption{Evolution of different training and validation metrics on the MATH-500 dataset.}
  \label{fig:evolution_during_training}
\end{figure}

These results are consistent with the larger increase in the estimated ${\text{SNR}}(\Delta_{j^\star_n})$ observed during training.
Recall from (\ref{eq:expected_number_samples}) that the required number of samples for the MMC stopping rule is approximately inversely proportional to the ${\text{SNR}}(\Delta_{j^\star_n})$. 
Figure~\ref{fig:evolution_during_training} shows the evolution of the estimated ${\text{SNR}}(\Delta_{j^\star_n})$ when using SNR-based rewards, as well as the negative entropy when training with entropy-based rewards, measured on the training dataset. We also include the evolution of the pass@1 performance on the validation dataset.

\begin{table}[ht!]
\caption{Regression coefficients from fitting the required number of samples under the MMC stopping rule as a function of the budget, ${N}_{\text{adaptive}} = \alpha + \beta {N}_{\text{budget}}$, for $\varepsilon = 0.1$ and $0.4$. Results contrast the pre-trained model with the model after test-time training using SNR-based rewards.}
\label{tab:ratio_adaptive_sampling_pre_post_trained}
\begin{center}
\footnotesize
\begin{tabular}{ccccccccc}
\toprule
  & \multicolumn{2}{c}{\textbf{Qwen2.5-Math-7B} }&\multicolumn{2}{c}{\textbf{Qwen2.5-Math-1.5B}} &\multicolumn{2}{c}{\textbf{Qwen2.5-7B}}&\multicolumn{2}{c}{\textbf{Llama-3.1-8B}}\\
\cmidrule(lr){2-3}
\cmidrule(lr){4-5}
\cmidrule(lr){6-7}
\cmidrule(lr){8-9}
 $\bm \varepsilon$ &0.1&0.4 & 0.1&0.4& 0.1&0.4& 0.1&0.4\\
\midrule
$\bm{\hat\beta}_{\text{pre}}$ pre-trained model  &0.725& 0.711&0.848 &0.798 & 0.627& 0.589& 0.645& 0.590\\
\\
$\bm{\hat\beta}_{\text{post}}$ test-time trained model &0.631 & 0.568& 0.570& 0.533 & 0.472&0.392& 0.564& 0.488\\
\\
$\nabla = \bm{\hat\beta}_{\text{pre}}-\bm{\hat\beta}_{\text{post}}$ & 0.094& 0.143& 0.237&0.265& 0.155& 0.197& 0.081& 0.102\\
\bottomrule
\end{tabular}
\end{center}
\end{table}
\normalsize

\begin{figure}[h!]
  \centering
  \begin{subfigure}{0.49\textwidth}
      \centering
      \includegraphics[width=\textwidth]{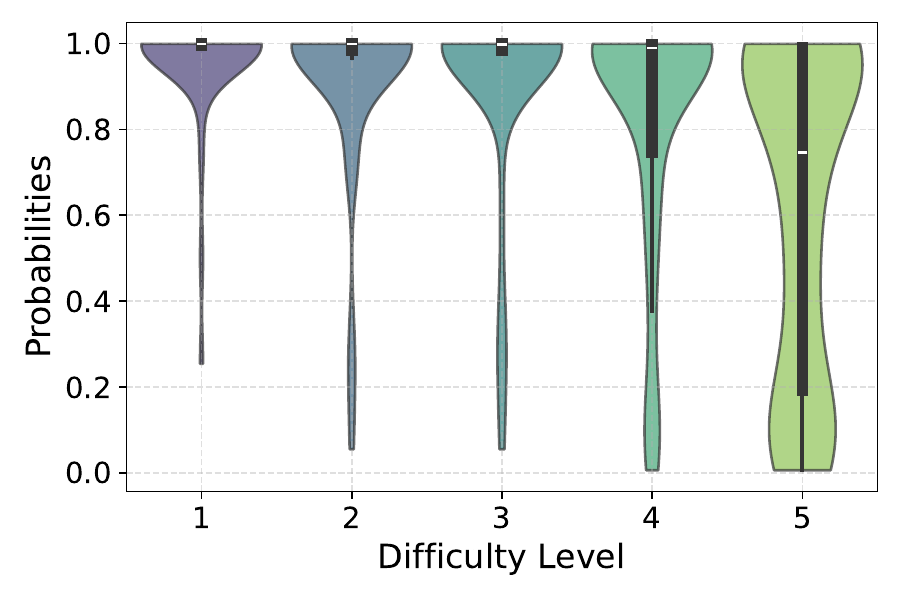}
      \caption{Qwen2.5-Math-1.5B, $N_{\text{budget}}=10$.}
      \label{fig:QWEN-MATH-1.5B_budget_10_NO_01}
  \end{subfigure}
  \hfill
  \begin{subfigure}{0.49\textwidth}
      \centering
      \includegraphics[width=\textwidth]{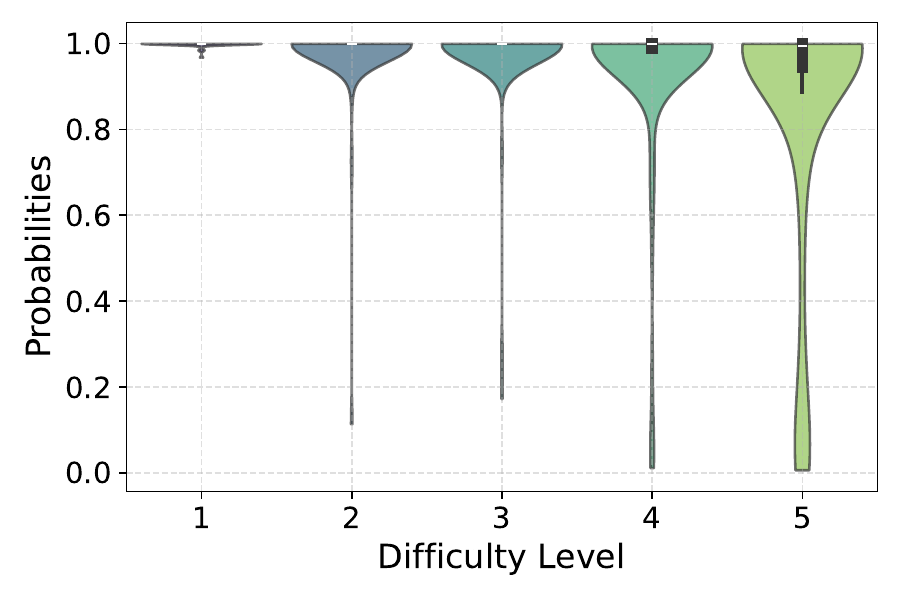}
        \caption{Qwen2.5-Math-7B, $N_{\text{budget}}=10$.}
      \label{fig:QWEN-MATH-7B_budget_10_NO_01}
  \end{subfigure}
  \vfill
  \begin{subfigure}{0.49\textwidth}
      \centering
      \includegraphics[width=\textwidth]{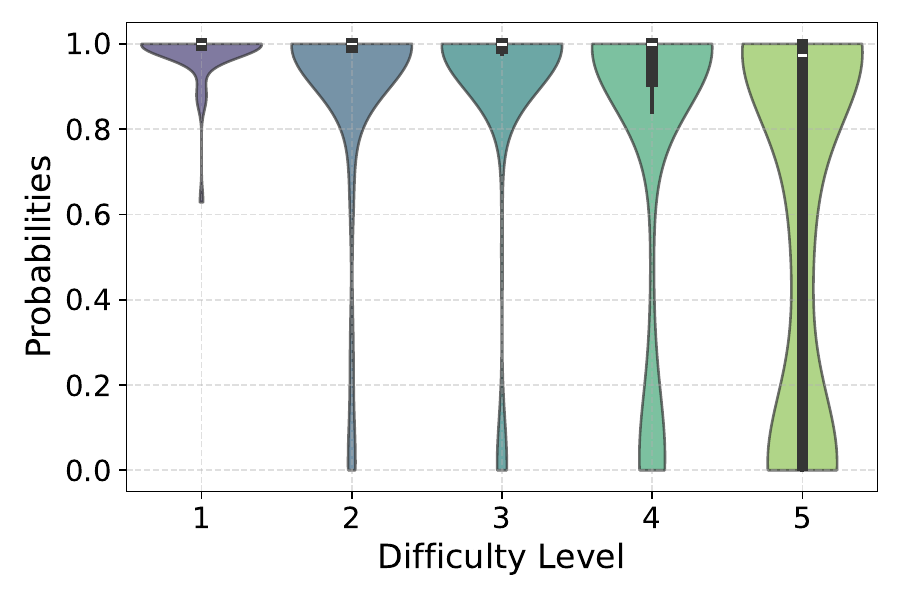}
        \caption{Qwen2.5-Math-1.5B, $N_{\text{budget}}=50$.}
      \label{fig:QWEN-MATH-1.5B_budget_50_NO_01}
  \end{subfigure}
  \hfill
  \begin{subfigure}{0.49\textwidth}
      \centering
      \includegraphics[width=\textwidth]{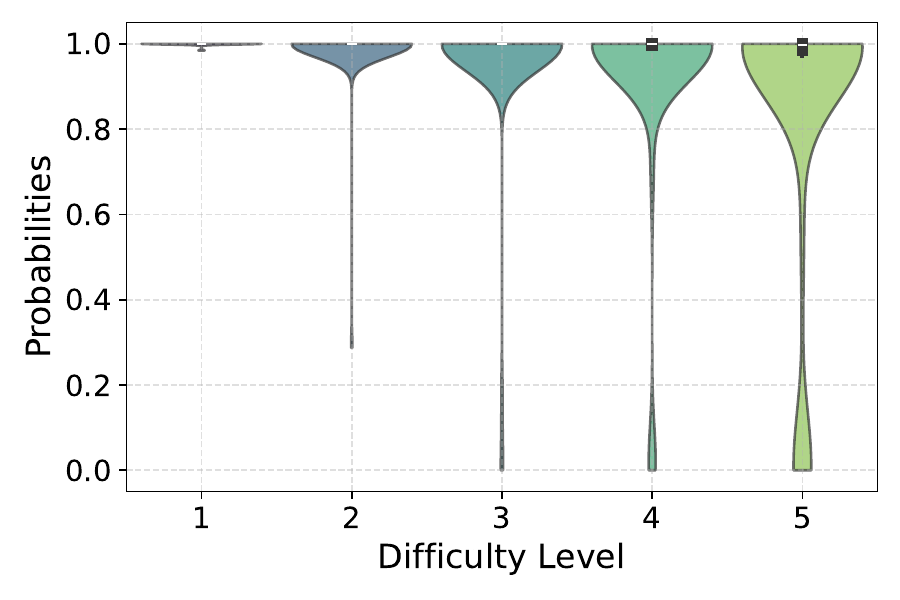}
        \caption{Qwen2.5-Math-7B, $N_{\text{budget}}=50$.}
      \label{fig:QWEN-MATH-7B_budget_50_NO_01}
  \end{subfigure}
  \vfill
  \begin{subfigure}{0.49\textwidth}
      \centering
      \includegraphics[width=\textwidth]{figs/QWEN-MATH-1.5B_violin_maj100_probability_adaptive_01_NO_ground_truth.pdf}
        \caption{Qwen2.5-Math-1.5B, $N_{\text{budget}}=100$.}
      \label{fig:QWEN-MATH-1.5B_budget_100_NO_01}
  \end{subfigure}
  \hfill
  \begin{subfigure}{0.49\textwidth}
      \centering
      \includegraphics[width=\textwidth]{figs/QWEN-MATH-7B_violin_maj100_probability_adaptive_01_NO_ground_truth.pdf}
        \caption{Qwen2.5-Math-7B, $N_{\text{budget}}=100$.}
      \label{fig:QWEN-MATH-7B_budget_100_NO_01}
  \end{subfigure}
  \caption{Violin plots illustrating the distribution of the estimated lower bound on the probability $\mathbb{P}[\widehat{c}_n = c^\star]$ when applying Martingale Majority Certificate stopping rule with $\varepsilon = 0.1$ across different budget values $N_{\text{budget}}$. 
  Results are obtained after test-time training with SNR-based rewards on the MATH-500 dataset.}
  \label{fig:violin_plots_NO_ground_truth_01}
\end{figure}

\begin{figure}[h!]
  \centering
  \begin{subfigure}{0.49\textwidth}
      \centering
      \includegraphics[width=\textwidth]{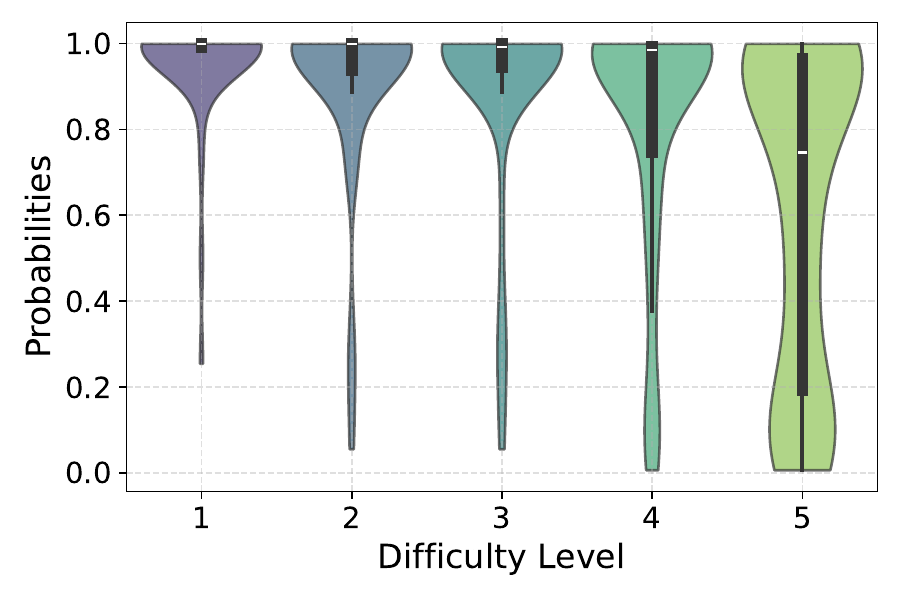}
      \caption{Qwen2.5-Math-1.5B, $N_{\text{budget}}=10$.}
      \label{fig:QWEN-MATH-1.5B_budget_10_NO_04}
  \end{subfigure}
  \hfill
  \begin{subfigure}{0.49\textwidth}
      \centering
      \includegraphics[width=\textwidth]{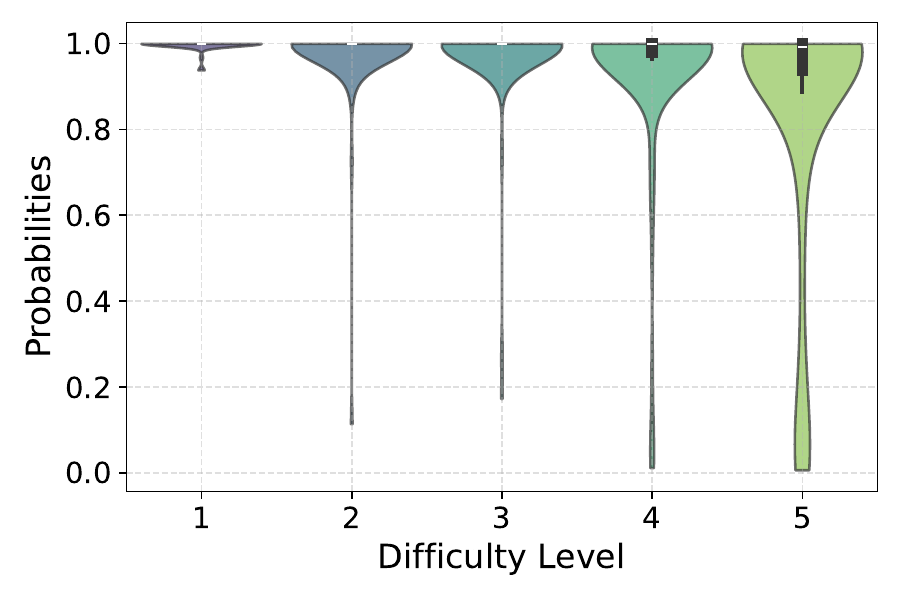}
        \caption{Qwen2.5-Math-7B, $N_{\text{budget}}=10$.}
      \label{fig:QWEN-MATH-7B_budget_10_NO_04}
  \end{subfigure}
  \vfill
  \begin{subfigure}{0.49\textwidth}
      \centering
      \includegraphics[width=\textwidth]{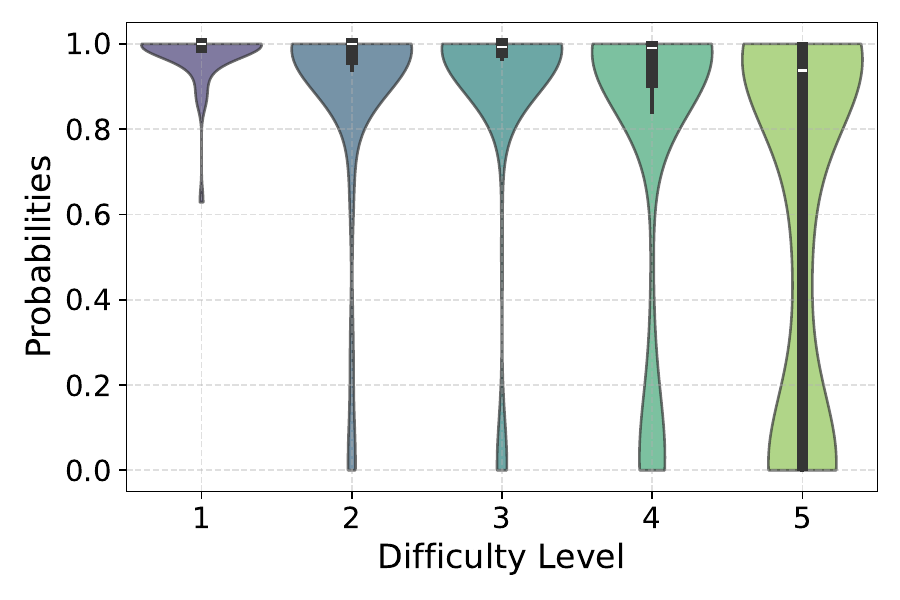}
        \caption{Qwen2.5-Math-1.5B, $N_{\text{budget}}=50$.}
      \label{fig:QWEN-MATH-1.5B_budget_50_NO_04}
  \end{subfigure}
  \hfill
  \begin{subfigure}{0.49\textwidth}
      \centering
      \includegraphics[width=\textwidth]{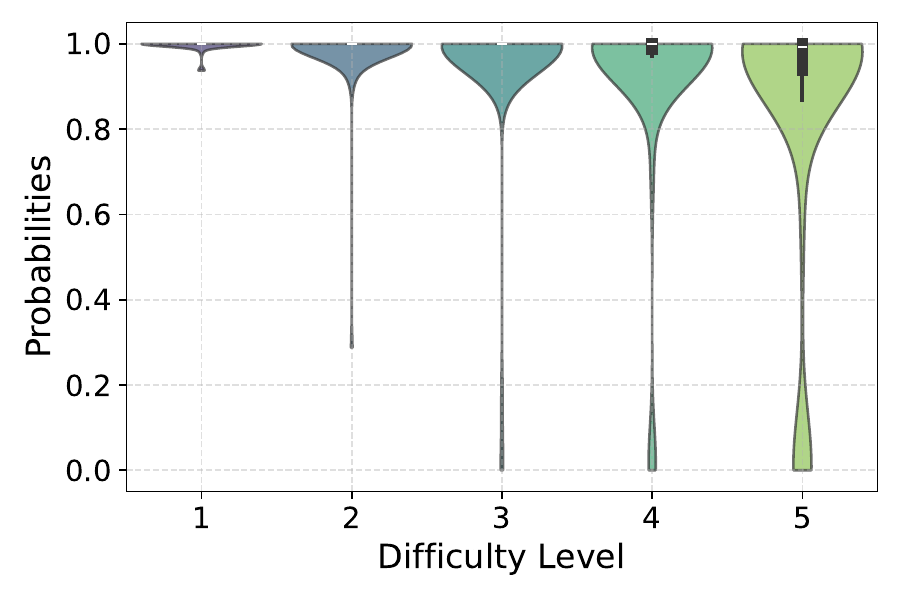}
        \caption{Qwen2.5-Math-7B, $N_{\text{budget}}=50$.}
      \label{fig:QWEN-MATH-7B_budget_50_NO_04}
  \end{subfigure}
  \vfill
  \begin{subfigure}{0.49\textwidth}
      \centering
      \includegraphics[width=\textwidth]{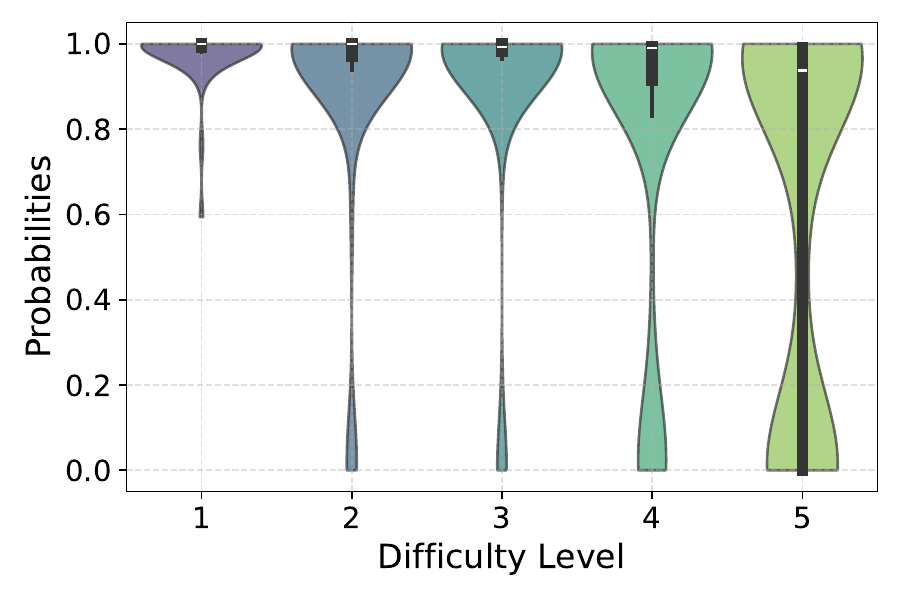}
        \caption{Qwen2.5-Math-1.5B, $N_{\text{budget}}=100$.}
      \label{fig:QWEN-MATH-1.5B_budget_100_NO_04}
  \end{subfigure}
  \hfill
  \begin{subfigure}{0.49\textwidth}
      \centering
      \includegraphics[width=\textwidth]{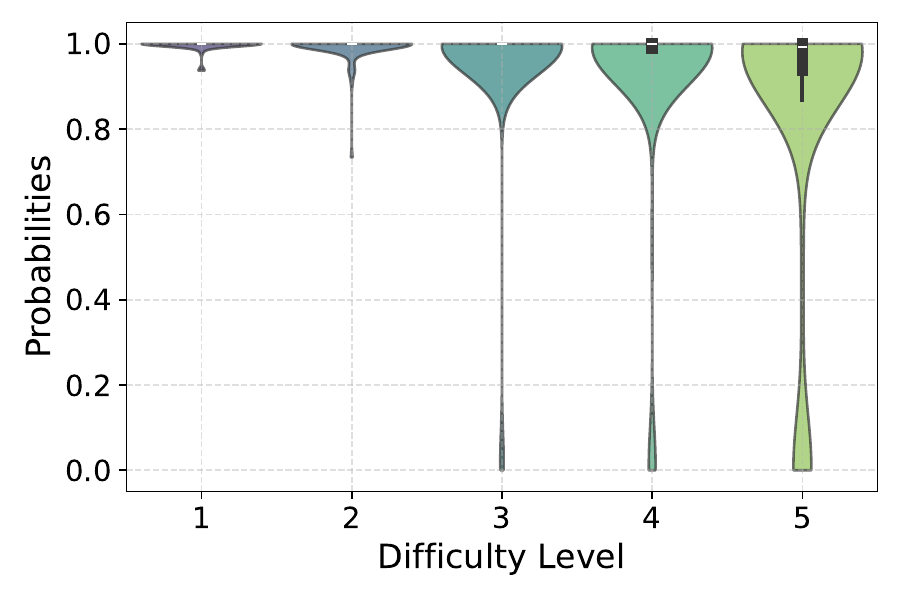}
        \caption{Qwen2.5-Math-7B, $N_{\text{budget}}=100$.}
      \label{fig:QWEN-MATH-7B_budget_100_NO_04}
  \end{subfigure}
  \caption{Violin plots illustrating the distribution of the estimated lower bound on the probability $\mathbb{P}[\widehat{c}_n = c^\star]$ when applying Martingale Majority Certificate stopping rule with $\varepsilon = 0.4$ across different budget values $N_{\text{budget}}$. 
   Results are obtained after test-time training with SNR-based rewards on the MATH-500 dataset.}
  \label{fig:violin_plots_NO_ground_truth_04}
\end{figure}

\begin{figure}[h!]
  \centering
  \begin{subfigure}{0.49\textwidth}
      \centering
      \includegraphics[width=\textwidth]{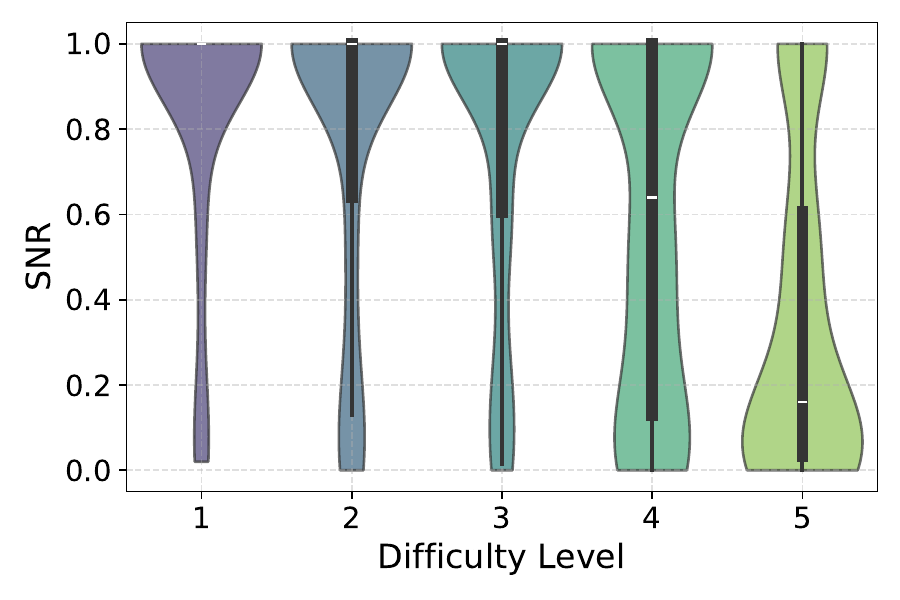}
      \caption{Qwen2.5-Math-1.5B, $N_{\text{budget}}=10$.}
      \label{fig:QWEN-MATH-1.5B_budget_10_SNR_01}
  \end{subfigure}
  \hfill
  \begin{subfigure}{0.49\textwidth}
      \centering
      \includegraphics[width=\textwidth]{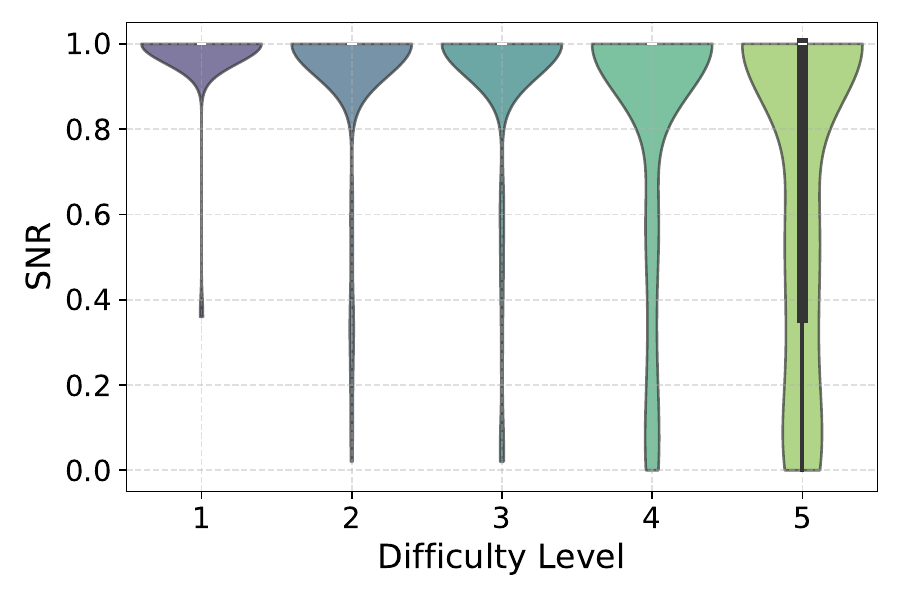}
        \caption{Qwen2.5-Math-7B, $N_{\text{budget}}=10$.}
      \label{fig:QWEN-MATH-7B_budget_10_SNR_01}
  \end{subfigure}
  \vfill
  \begin{subfigure}{0.49\textwidth}
      \centering
      \includegraphics[width=\textwidth]{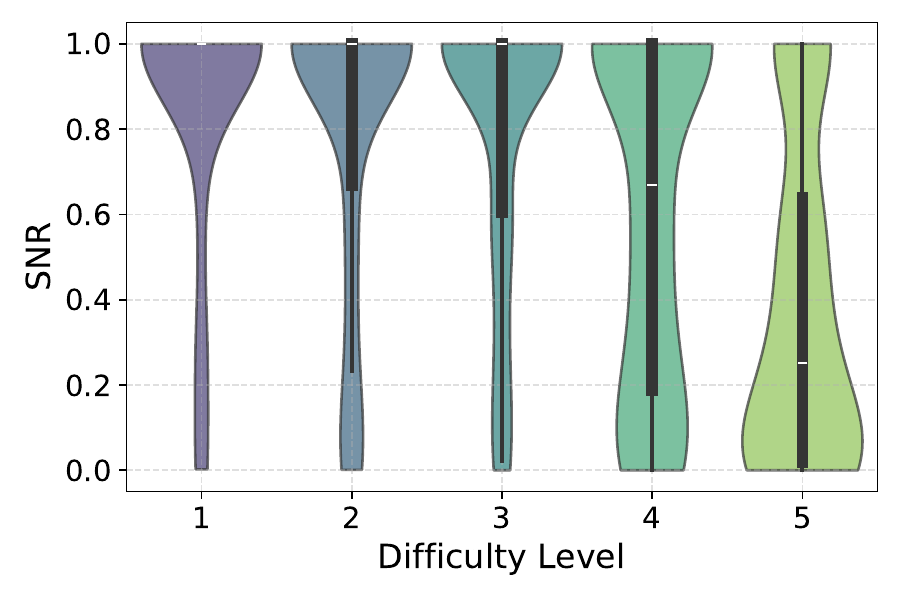}
        \caption{Qwen2.5-Math-1.5B, $N_{\text{budget}}=50$.}
      \label{fig:QWEN-MATH-1.5B_budget_50_SNR_01}
  \end{subfigure}
  \hfill
  \begin{subfigure}{0.49\textwidth}
      \centering
      \includegraphics[width=\textwidth]{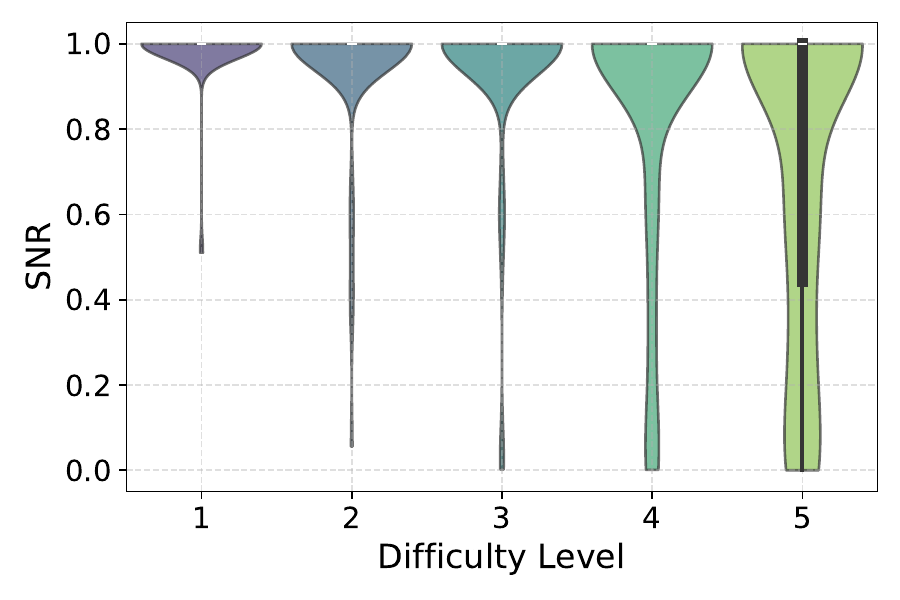}
        \caption{Qwen2.5-Math-7B, $N_{\text{budget}}=50$.}
      \label{fig:QWEN-MATH-7B_budget_50_SNR_01}
  \end{subfigure}
  \vfill
  \begin{subfigure}{0.49\textwidth}
      \centering
      \includegraphics[width=\textwidth]{figs/QWEN-MATH-1.5B_violin_maj100_SNR_01.pdf}
        \caption{Qwen2.5-Math-1.5B, $N_{\text{budget}}=100$.}
      \label{fig:QWEN-MATH-1.5B_budget_100_SNR_01}
  \end{subfigure}
  \hfill
  \begin{subfigure}{0.49\textwidth}
      \centering
      \includegraphics[width=\textwidth]{figs/QWEN-MATH-7B_violin_maj100_SNR_01.pdf}
        \caption{Qwen2.5-Math-7B, $N_{\text{budget}}=100$.}
      \label{fig:QWEN-MATH-7B_budget_100_SNR_01}
  \end{subfigure}
  \caption{Violin plots showing the distribution of the estimated signal-to-noise ratio between the leader and runner-up, $\text{SNR}(\Delta_{j^\star_n})$, when using Martingale Majority Certificate stopping rule with $\varepsilon = 0.1$ across different budget values $N_{\text{budget}}$. Results are obtained after applying test-time training with SNR-based rewards on the MATH-500 dataset.}
  \label{fig:violin_plots_SNR_ground_truth_01}
\end{figure}

\begin{figure}[h!]
  \centering
  \begin{subfigure}{0.49\textwidth}
      \centering
      \includegraphics[width=\textwidth]{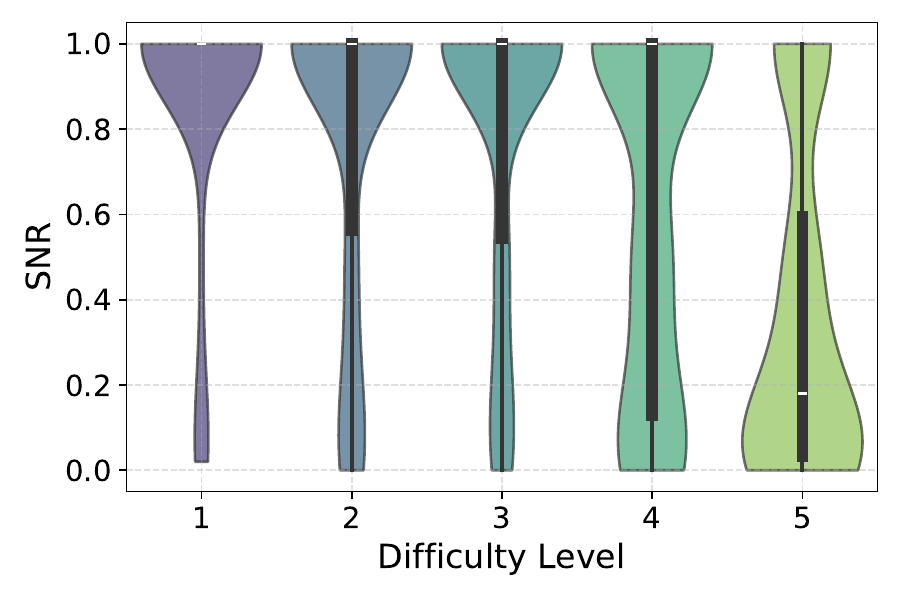}
      \caption{Qwen2.5-Math-1.5B, $N_{\text{budget}}=10$.}
      \label{fig:QWEN-MATH-1.5B_budget_10_SNR_04}
  \end{subfigure}
  \hfill
  \begin{subfigure}{0.49\textwidth}
      \centering
      \includegraphics[width=\textwidth]{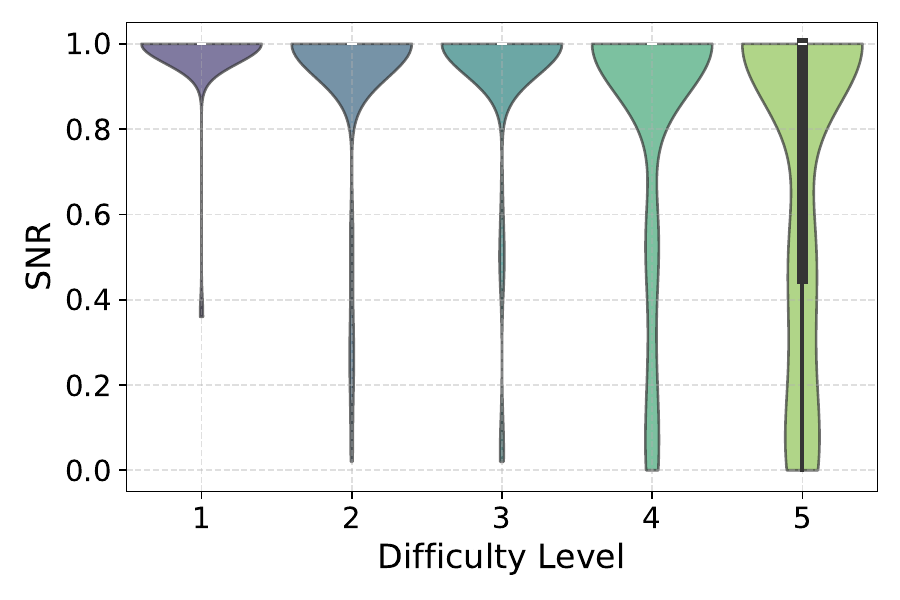}
        \caption{Qwen2.5-Math-7B, $N_{\text{budget}}=10$.}
      \label{fig:QWEN-MATH-7B_budget_10_SNR_04}
  \end{subfigure}
  \vfill
  \begin{subfigure}{0.49\textwidth}
      \centering
      \includegraphics[width=\textwidth]{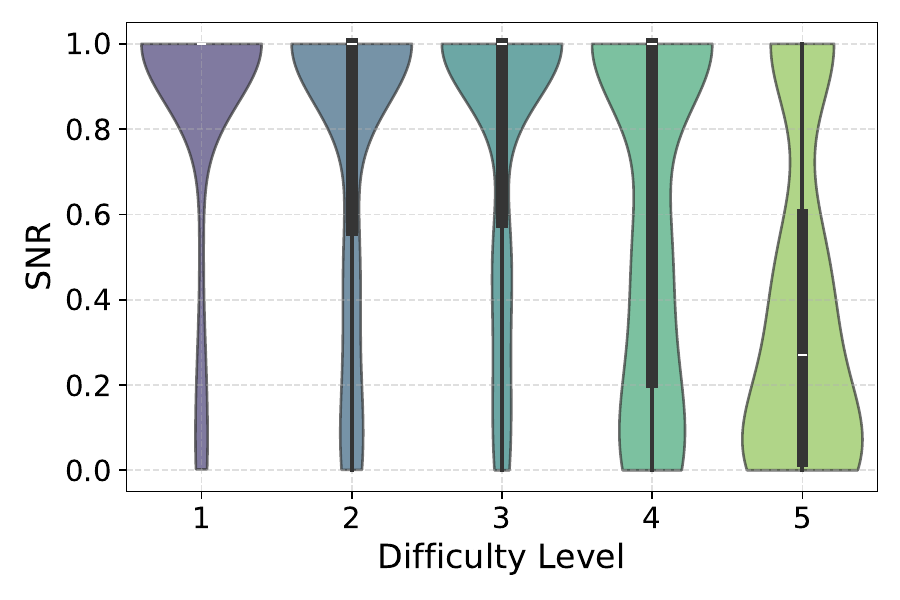}
        \caption{Qwen2.5-Math-1.5B, $N_{\text{budget}}=50$.}
      \label{fig:QWEN-MATH-1.5B_budget_50_SNR_04}
  \end{subfigure}
  \hfill
  \begin{subfigure}{0.49\textwidth}
      \centering
      \includegraphics[width=\textwidth]{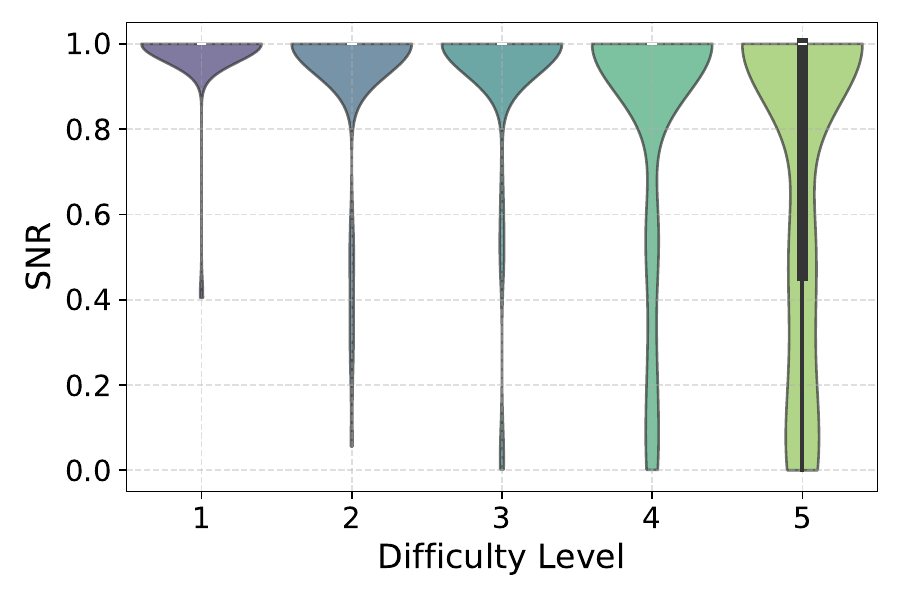}
        \caption{Qwen2.5-Math-7B, $N_{\text{budget}}=50$.}
      \label{fig:QWEN-MATH-7B_budget_50_SNR_04}
  \end{subfigure}
  \vfill
  \begin{subfigure}{0.49\textwidth}
      \centering
      \includegraphics[width=\textwidth]{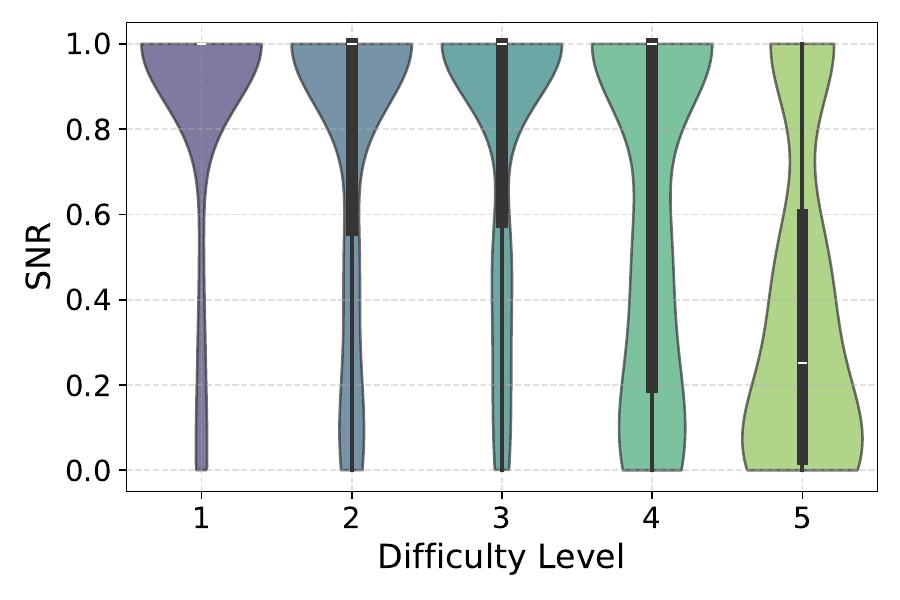}
        \caption{Qwen2.5-Math-1.5B, $N_{\text{budget}}=100$.}
      \label{fig:QWEN-MATH-1.5B_budget_100_SNR_04}
  \end{subfigure}
  \hfill
  \begin{subfigure}{0.49\textwidth}
      \centering
      \includegraphics[width=\textwidth]{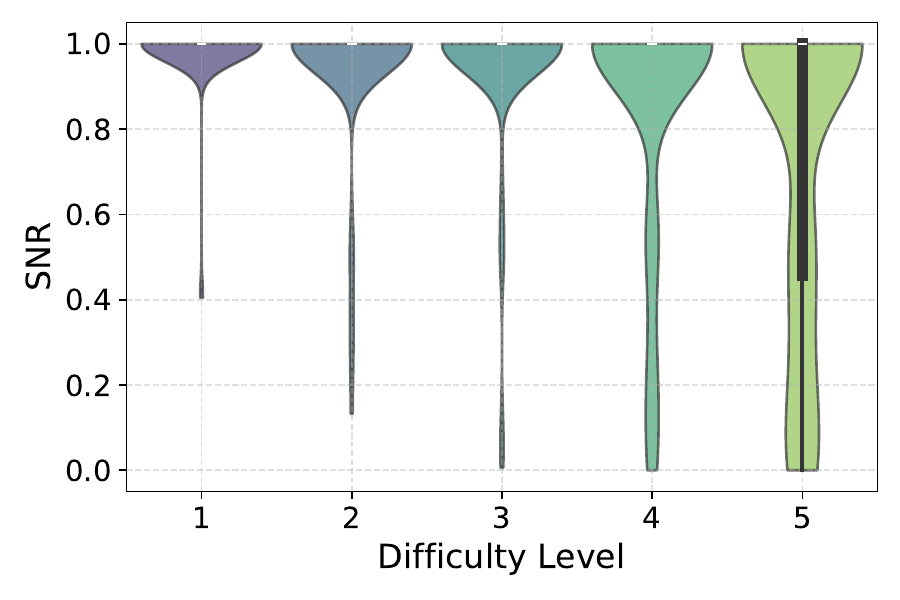}
        \caption{Qwen2.5-Math-7B, $N_{\text{budget}}=100$.}
      \label{fig:QWEN-MATH-7B_budget_100_SNR_04}
  \end{subfigure}
  \caption{Violin plots showing the distribution of the estimated signal-to-noise ratio between the leader and runner-up, ${\text{SNR}}(\Delta_{j^\star_n})$, when using Martingale Majority Certificate stopping rule with $\varepsilon = 0.4$ across different budget values $N_{\text{budget}}$. Results are obtained after applying test-time training with SNR-based rewards on the MATH-500 dataset.}
  \label{fig:violin_plots_SNR_ground_truth_04}
\end{figure}

Finally, Figures \ref{fig:violin_plots_NO_ground_truth_01}-\ref{fig:violin_plots_SNR_ground_truth_04} provide a detailed analysis, for each difficulty level in the MATH-500 dataset, of the distributions of the estimated lower bound on the probability $\mathbb{P}[\widehat{c}_n= c^\star]$, as well as the estimated ${\text{SNR}}(\Delta_{j^\star_n})$ when applying the MMC adaptive sampling scheme under two confidence levels, $\varepsilon = 0.1$ and $0.4$. 
The lower bound estimates of $\mathbb{P}[\widehat{c}_n = c^\star]$ (Figures \ref{fig:violin_plots_NO_ground_truth_01}, \ref{fig:violin_plots_NO_ground_truth_04}) are computed using a Beta approximation (see Appendix \ref{app:subsec_estimator_proability} for details).
Results are reported after test-time training with SNR-based rewards.
The SNR plots (Figures~\ref{fig:violin_plots_SNR_ground_truth_01}, \ref{fig:violin_plots_SNR_ground_truth_04}) further illustrate how SNR can serve as a label-free estimator of problem difficulty.

\end{document}